\def\year{2022}\relax
\documentclass[letterpaper]{article} %
\usepackage{aaai22}  %
\usepackage{times}  %
\usepackage{helvet}  %
\usepackage{courier}  %
\usepackage[hyphens]{url}  %
\usepackage{graphicx} %
\usepackage{natbib}  %
\usepackage{caption} %
\DeclareCaptionStyle{ruled}{labelfont=normalfont,labelsep=colon,strut=off} %
\usepackage[colorlinks=true,citecolor=blue,urlcolor=red]{hyperref}

\usepackage{amsmath}
\usepackage{amssymb}
\usepackage{nicefrac}

\usepackage{booktabs}
\usepackage{multirow}
\usepackage{bbm}
\usepackage{array}
\usepackage{graphbox}
\usepackage{makecell}
\usepackage{arydshln}

\usepackage{color}
\usepackage{colortbl}

\usepackage[linesnumbered,algoruled,boxed,lined,noend]{algorithm2e}
\SetKwProg{Fn}{Function}{}{end}

\newtheorem{proposition}{Proposition}[section]
\newtheorem{innercustomprop}{Proposition}
\newenvironment{customprop}[1]{\renewcommand\theinnercustomprop{#1}\innercustomprop}{\endinnercustomprop}

\newtheorem{remark}{Remark}[section]

\newenvironment{proof}{\textit{Proof.}}{\hfill$\square$}

\newcommand{\norm}[1]{\left\|#1\right\|}

\newcommand{\inner}[1]{\left\langle#1\right\rangle}

\def\Exp{\mathbb{E}}

\def\argmax{\mathop{\rm arg\,max}\limits}%
\def\argmin{\mathop{\rm arg\,min}\limits}%

\def\R{\mathbb{R}}

\def\R{\mathbb{R}}

\def\D{\mathcal{D}}

\def\U{\mathcal{U}}
\def\P{\mathcal{P}}

\newcolumntype{C}[1]{>{\centering\arraybackslash}p{#1}}
\newcolumntype{L}[1]{>{\raggedright\arraybackslash}p{#1}}
\newcolumntype{R}[1]{>{\raggedleft\arraybackslash}p{#1}}
\newcommand\newcomment[1]{}

\makeatletter
\def\blfootnote{\xdef\@thefnmark{}\@footnotetext}
\makeatother

\newcommand{\iter}[2]{#1^{(#2)}}
\def\xorig{x_\textrm{orig}}

\renewcommand{\S}{\mathcal{S}}
\def\T{\mathcal{T}}
\def\rsa{{\texttt{Sparse-RS}}}

\definecolor{lightgrey}{rgb}{0.9, 0.9, 0.9}

\newlength\newl

\newif\ifpaper
\papertrue

\graphicspath{{figures/}}

\title{%
Sparse-RS: a Versatile Framework for Query-Efficient\\ Sparse Black-Box Adversarial Attacks}
\author{
    Francesco Croce,\textsuperscript{\rm 1} Maksym Andriushchenko,\textsuperscript{\rm 2} Naman D. Singh,\textsuperscript{\rm 1}\\ Nicolas Flammarion,\textsuperscript{\rm 2} Matthias Hein\textsuperscript{\rm 1}
}
\affiliations{
    \textsuperscript{\rm 1} University of T{\"u}bingen,
    \textsuperscript{\rm 2} EPFL

}

\begin{document}

\maketitle

\begin{abstract}
We propose a versatile framework based on random search, \texttt{Sparse-RS}, for score-based sparse targeted and untargeted attacks in the black-box setting. \texttt{Sparse-RS} does not rely on substitute models and achieves state-of-the-art success rate and query efficiency for multiple sparse attack models: $l_0$-bounded perturbations, adversarial patches, and adversarial frames. The $l_0$-version of untargeted \texttt{Sparse-RS} outperforms all black-box and even all white-box attacks for different models on MNIST, CIFAR-10, and ImageNet.
Moreover, our untargeted \texttt{Sparse-RS} achieves very high success rates even for the challenging settings of $20\times20$ adversarial patches and $2$-pixel wide adversarial frames for $224\times224$ images. Finally, we show that \texttt{Sparse-RS} can be applied to generate targeted universal adversarial patches where it significantly outperforms %
the existing approaches.
The code of our framework is available at \url{https://github.com/fra31/sparse-rs}.
\end{abstract}

\section{Introduction}
The discovery of the vulnerability of neural networks to adversarial examples \cite{BigEtAl13,SzeEtAl2014} revealed that the decision of a classifier or a detector can be changed by small, carefully chosen perturbations of the input. %
Many efforts have been put into developing increasingly more sophisticated attacks to craft small, semantics-preserving modifications which are able to fool classifiers and bypass many defense mechanisms \cite{CarWag2016,AthEtAl2018}. This is typically achieved by constraining or minimizing the $l_p$-norm of the perturbations, usually either $l_\infty$ \cite{SzeEtAl2014,KurGooBen2016a,CarWag2016,MadEtAl2018,CroHei2019}, $l_2$ \cite{MooFawFro2016,CarWag2016,rony2019decoupling,CroHei2019} or $l_1$ \cite{CheEtAl2018,ModEtAl19,CroHei2019}. Metrics other than $l_p$-norms which are more aligned to human perception have been also recently used, e.g. Wasserstein distance \cite{WonEtAl2019,hu2020improved} or neural-network based ones such as LPIPS \cite{zhang2018unreasonable,laidlaw2020perceptual}.
All these attacks have in common the tendency to modify all the elements of the input.

\begin{figure}[t] \centering
    \includegraphics[width=0.95\columnwidth]{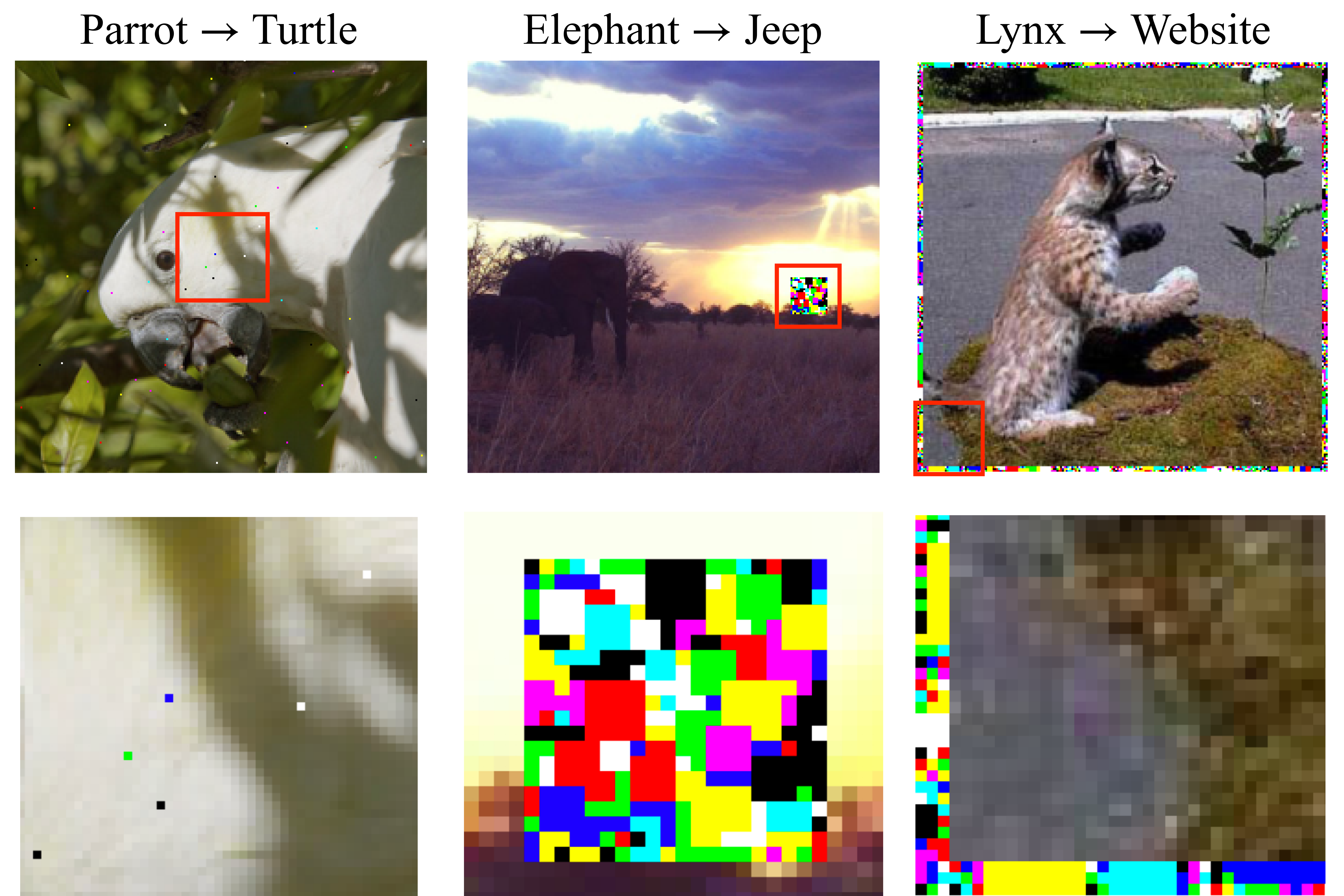}
    \caption{%
    Adversarial examples for sparse threat models ($\l_0$-bounded, patches, frames) generated with our black-box \rsa{} framework which %
    does not require surrogate models and is %
    more query efficient. 
    }
    \label{fig:teaser_picture}%
\end{figure}

Conversely, \emph{sparse attacks} pursue an opposite strategy: they perturb only a small portion of the original input but possibly with large modifications. Thus, the perturbations are indeed visible but do not alter the semantic content, and can even be applied in the physical world \cite{lee2019physical,thys2019fooling,li19stickers}. Sparse attacks include $l_0$-attacks \cite{narodytska2017simple,CarWag2016,papernot2016limitations,SchEtAl19,croce2019sparse}, %
adversarial patches \cite{BroEtAl2017,karmon2018lavan,lee2019physical} and frames \cite{zajac2019adversarial}, where the perturbations have some predetermined structure. Moreover, sparse attacks generalize to tasks outside computer vision, such as malware detection or natural language processing, where the nature of the domain imposes to modify only a limited number of input features \cite{grosse2016adversarial, jin2019bert}.

We focus on the black-box score-based scenario, where the attacker can only access the predicted scores of a classifier $f$, but does not know the network weights %
and in particularly cannot use gradients of $f$ wrt the input (as in the white-box setup). We do not consider more restrictive (e.g., decision-based attacks \cite{BreRauBet18,brunner2018guessing} where the adversary only knows the label assigned to each input) or more permissive (e.g., a surrogate model similar to the victim one is available \cite{cheng2019improving, Huang2020Black-Box}) cases. %
For the $l_0$-threat model only a few black-box attacks exist \cite{narodytska2017simple, SchEtAl19, croce2019sparse, zhao2019design}, which however do not focus on query efficiency or scale to datasets like ImageNet without suffering from prohibitive computational cost.
For adversarial patches and frames, black-box methods are mostly limited to transfer attacks, that is a white-box attack is performed on a surrogate model, with the exception of \cite{yang2020patchattack} who use a predefined %
dictionary of patches.

\textbf{Contributions.} 
Random search is particularly suitable for zeroth-order optimization in presence of complicated combinatorial constraints, as those of sparse threat models. Then, we design specific sampling distributions for the random search algorithm to efficiently generate sparse black-box attacks. The resulting \rsa{} is a simple and flexible framework which handles
\begin{itemize}
    \item %
    \textbf{$l_0$-perturbations}: \rsa{} significantly outperforms the existing black-box attacks in terms of the query efficiency and success rate, and %
    leads to a better success rate even when compared to the state-of-the-art \textit{white-box} attacks on standard and robust models. %
    \item %
    \textbf{Adversarial patches}: \rsa{} %
    achieves better results than both %
    TPA \cite{yang2020patchattack} and a black-box adaptations of projected gradient descent (PGD) attacks via gradient estimation.
    \item %
    \textbf{Adversarial frames}: \rsa{} %
    outperforms the existing adversarial framing method \cite{zajac2019adversarial} with gradient estimation and achieves a very high success rate even with 2-pixel wide frames. 
\end{itemize}
Due to space reasons the results for adversarial frames had to be moved to the appendix.

\section{Black-box adversarial attacks} \label{sec:adversarial_setting}
Let $f:\S \subseteq \R^d \rightarrow \R^K$ be a classifier which assigns input $x\in\S$ to class $y=\text{arg\,max}_{r=1, \ldots, K}f_r(x)$. The goal of an \textit{untargeted} attack is to craft a perturbation $\delta \in \R^d$ s.t.
\begin{align*} 
    \argmax_{r=1, \ldots, K}f_r(x + \delta)\neq y, \quad x+\delta \in \S \quad \text{and} \quad \delta \in \T, 
\end{align*} where $\S$ is the input domain
and $\T$ are the constraints the adversarial perturbation has to fulfill (e.g. bounded $l_p$-norm), %
while a \textit{targeted} attack aims at finding $\delta$ such that 
\begin{align*} 
    \argmax_{r=1, \ldots, K}f_r(x + \delta)= t, \quad x+\delta \in \S \quad \text{and} \quad \delta \in \T, 
\end{align*} with $t$ as target class.
Generating such $\delta$ can be translated into an optimization problem as 
\begin{align}
    \label{eq:adv_opt} 
    \min\limits_{\delta\in\R^d} \; L(f(x+\delta), t) \quad \text{s.t.} \quad x+\delta \in \S \quad \text{and} \quad \delta \in \T 
\end{align} 
by choosing a label $t$ and loss function $L$ whose minimization leads to the desired classification. 
By \textit{threat model} we mean the overall attack setting determined by the goal of the attacker (targeted vs untargeted attack), the level of knowledge (white- vs black-box), and the perturbation set $\T$.

Many algorithms have been proposed to solve Problem~\eqref{eq:adv_opt} in the black-box setting where one cannot use gradient-based methods.
One of the first approaches is by \cite{fawzi2016measuring} %
who propose to sample candidate adversarial occlusions %
via the Metropolis MCMC method, which can be seen as a way to generate adversarial patches whose content is not optimized.
\cite{ilyas2018black,uesato2018adversarial} propose to approximate the gradient through finite difference methods, later improved to reduce their high computational cost in terms of queries of the victim models \cite{bhagoji2018practical,tu2019autozoom,ilyas2019prior}.
Alternatively, \cite{alzantot2018genattack,liu2019adversarial} use genetic algorithms in the context of image classification and malware detection respectively. A line of research has focused on rephrasing $l_\infty$-attacks as discrete optimization problems \cite{MooEtAl2019, AlDujaili2019ThereAN, MeuEtAl2019}, where specific techniques lead to significantly better query efficiency.
\cite{guo2019simple} adopt a variant of random search to produce perturbations with a small $l_2$-norm.

Closest in spirit %
is the Square Attack of \cite{ACFH2019square}, which is state-of-the-art for $l_\infty$- and $l_2$-bounded black-box attacks. It uses random search to iteratively generate samples on the surface of the $l_\infty$- or $l_2$-ball. Together with a particular sampling distribution based on square-shaped updates and a specific initialization, this leads to a simple algorithm which outperforms more sophisticated attacks in %
success rate and query efficiency. In this paper we show that the random search idea is ideally suited for sparse attacks, where the non-convex, combinatorial constraints are not easily handled even by gradient-based \textit{white-box} attacks.

\section{\rsa{} framework}  %
Random search (RS) is a well known scheme for derivative free optimization \cite{rastrigin1963convergence}. Given an objective function $L$ to minimize, a starting point $\iter{x}{0}$ and a sampling distribution $\D$, an iteration of RS at step $i$ is given by
\begin{align}\label{eq:rs_iter}
\delta \sim \D(\iter{x}{i}), \quad \iter{x}{i+1}=\argmin_{y \in \{\iter{x}{i}, \,\iter{x}{i} + \delta\}} L(y). \end{align}
At every step an update of the current iterate $\iter{x}{i}$ is sampled according to $\D$ and accepted only if it decreases the objective value, otherwise the procedure is repeated. Although not explicitly mentioned in Eq. \eqref{eq:rs_iter}, constraints on %
the iterates $\iter{x}{i}$ can be integrated by %
ensuring that $\delta$ is sampled so that $\iter{x}{i} + \delta$ is a feasible solution. Thus even complex, e.g. combinatorial, constraints can easily be integrated as RS just needs to be able to produce feasible points in contrast to gradient-based methods which depend on a continuous set to optimize over. %
While simple and flexible, RS is an effective tool in many tasks \cite{zabinsky2010random, ACFH2019square}, with the key ingredient for its success being a task-specific sampling distribution $\D$ to guide the exploration of the space of possible solutions.

\begin{algorithm}[t]
	\caption{\rsa{}}
	\label{alg:rs_general}
	\SetKwData{Left}{left}
  \SetKwData{Up}{up}
  \SetKwFunction{FindCompress}{FindCompress}
  \SetKwInOut{Input}{input}
  \SetKwInOut{Output}{output}
	\Input{loss $L$, input $\xorig{}$, max query $N$, sparsity $k$, input space constraints $\S$}
	\Output{approximate minimizer of $L$}
    $M \gets$ $k$ indices of elements to be perturbed\\ %
    $\Delta \gets $ values of the perturbation to be applied\\ %
    $z \gets \xorig{}$, \quad $z_M \gets \Delta$ \ \ \tcp*[h]{set elements in $M$ to values in $\Delta$}\\
    $L^* \gets L(z)$, \quad $i\gets 0$ \ \ \tcp*[h]{initialize loss} \\
    \While{$i < N$ {\rm\bf and} success not achieved}{
    $M' \gets $ sampled modification of $M$ \\ \tcp*[h]{new set of indices} \\
    $\Delta' \gets$ sampled modification of $\Delta$ \\ \tcp*[h]{new perturbation} \\
    $z \gets \xorig{}$, \quad $z_{M'} \gets \Delta'$ \\ \tcp*[h]{create new candidate in $\S$} \\
    \If{$L(z) < L^*$}{
    $L^* \gets L(z)$, \ \ $M \gets M'$, \ \ $\Delta \gets \Delta'$ \ \ \tcp*[h]{if loss improves, update sets}
    }%
    $i \gets i + 1$
    }%
    $z \gets \xorig{}$, \quad $z_M \gets \Delta$ \ \ \tcp*[h]{return best $\Delta$} \\
    {\rm \bf return} $z$
\end{algorithm}%

We summarize our general framework based on random search to generate sparse adversarial attacks, \rsa{}, in Alg.~\ref{alg:rs_general}, where the sparsity $k$ indicates the maximum number of features that can be perturbed. 
A sparse attack is characterized by two variables: the set of components to be perturbed $M$ and the values $\Delta$ to be inserted at $M$ to form the adversarial input.
To optimize over both of them we first sample a random update of the locations $M$ of the current perturbation (step 6) and then a random update of its values $\Delta$ (step~7). In some threat models (e.g. adversarial frames) the set $M$ cannot be changed, so $M'\equiv M$ at every step. How $\Delta'$ is generated depends on the specific threat model, so we present the individual procedures in the next sections. We note that for all threat models, the runtime is dominated by the cost of a forward pass through the network, and all other operations are computationally inexpensive.

Common to all variants of \rsa{} is that the whole budget for the perturbations is fully exploited 
both in terms of number of modified components and magnitude of the elementwise changes (constrained only by the limits of the input domain $\S$). This follows the intuition that larger perturbations should lead faster to an adversarial example.
Moreover, the difference of the candidates $M'$ and $\Delta'$ with $M$ and $\Delta$ shrinks gradually with the iterations which mimics the reduction of the step size in gradient-based optimization: initial large steps allow to quickly decrease the objective loss, but smaller steps are necessary to refine a close-to-optimal solution at the end of the algorithm. Finally, we impose a limit $N$ on the maximum number of queries of the classifier, i.e. evaluations of the objective function.

As objective function $L$ to be minimized, we use in the case of untargeted attacks the margin loss $L_\text{margin}(f(\cdot),y)=f_y(\cdot) - \max_{r\neq y}f_r(\cdot)$, where $y$ is the correct class, so that $L <0$ is equivalent to misclassification, whereas for targeted attacks we use the cross-entropy loss $L_\text{CE}$ of the target class $t$, namely $L_\text{CE}(f(\cdot), t) = -f_t(\cdot) + \log\left(\sum_{r=1}^Ke^{f_r(\cdot)}\right)$.

The code of the \texttt{Sparse-RS} framework is available at \url{https://github.com/fra31/sparse-rs}.

\section{\texttt{Sparse-RS} for $l_0$-bounded attacks} \label{sec:l0_bounded}
The first threat model we consider are $l_0$-bounded adversarial examples where only up to $k$ pixels or $k$ features/color channels of an input $\xorig{} \in[0, 1]^{h \times w \times c}$ (width $w$, height $h$, color $c$) can be modified, but there are no constraints on the magnitude of the perturbations except for those of the input domain. Note that constraining the number of perturbed pixels or features leads to two different threat models which are not directly comparable. Due to the combinatorial nature of the $l_0$-threat model, this turns out to be quite difficult for continuous optimization techniques which are more prone to get stuck in suboptimal maxima.

\textbf{\texttt{$l_0$-RS} algorithm.}
We first consider the threat model where up to $k$ pixels can be modified.
Let $U$ be the set of the $h\cdot w$ pixels.
In this case the set $M\subset U$ from Alg.~\ref{alg:rs_general} is initialized sampling uniformly $k$ elements of $U$, while $\Delta \sim \U(\{0, 1\}^{k\times c})$, that is random values in $\{0, 1\}$ (every perturbed pixel gets one of the corners of the color cube $[0,1]^c$).
Then, at the $i$-th iteration, we randomly select $A\subset M$ and $B\subset U\setminus M$, with $|A| =|B|
=\iter{\alpha}{i} \cdot k$, and create $M' = (M \setminus A) \cup B$. 
$\Delta'$ is formed by sampling random values from $\{0, 1\}^c$ for the elements in $B$, i.e. those which were not perturbed at the previous iteration.
The quantity $\iter{\alpha}{i}$ controls 
how much $M'$ differs from $M$ and decays following a predetermined piecewise constant schedule rescaled according to the maximum number of queries $N$.
The schedule is completely determined by the single value $\alpha_\text{init}$, used to calculate $\iter{\alpha}{i}$ for every iteration $i$, which is also the only free hyperparameter of our scheme. We provide details about the algorithm, schedule, and values of $\alpha_\text{init}$ in App.~\ref{sec:app_image_classification} and \ref{sec:app_malware_detection}, and ablation studies for them in App.~\ref{sec:app_ablation_st}. For the feature based threat model each color channel is treated as a pixel and one applies the scheme above to the ``gray-scale'' image ($c=1$) with three times as many ``pixels''.

\subsection{Comparison of query efficiency of $l_0$-RS}
\label{sec:l0_imagenet}
We compare pixel-based \texttt{$l_0$-RS} to other black-box untargeted attacks in terms of success rate versus query efficiency. The results of targeted attacks are in App.~\ref{sec:app_image_classification}. %
Here we focus on attacking normally trained VGG-16-BN and ResNet-50 models on ImageNet, which contains RGB images resized to shape $224 \times 224$, that is 50,176 pixels, belonging to 1,000 classes. %
We consider perturbations of size $k\in\{50, 150\}$ pixels to assess the effectiveness of the untargeted attacks at different thresholds with a limit of 10,000 queries. We evaluate the success rate on the initially correctly classified images out of 500 images from the validation set.

\textbf{Competitors.} Many existing black-box pixel-based $l_0$-attacks \cite{narodytska2017simple,SchEtAl19,croce2019sparse} do not aim at query efficiency %
and rather try to minimize the size of the perturbations. Among them, only CornerSearch \cite{croce2019sparse} and ADMM attack \cite{zhao2019design} scale to ImageNet. However, CornerSearch requires $8\times \# pixels$ queries only for the initial phase, exceeding the query limit we fix by more than $40$ times. The ADMM attack %
tries to achieve a successful perturbation and then reduces its $l_0$-norm. %
Moreover, we introduce black-box versions of PGD\textsubscript{$0$} \cite{croce2019sparse} with the gradient estimated by finite difference approximation as done in prior work, e.g., see \cite{ilyas2018black}. As a strong baseline, we introduce JSMA-CE which is a version of the JSMA algorithm \cite{papernot2016limitations} that we adapt to the black-box setting: (1) %
for query efficiency, we estimate the gradient of the cross-entropy loss instead of gradients of \textit{each} class logit, (2) on each iteration, we modify the \textit{pixels} 
with the highest gradient contribution. More details about the attacks  can be found in App.~\ref{sec:app_image_classification}.

\begin{figure}[t] \centering \small
    \begin{tabular}{C{2mm} c}
    \begin{tabular}{c} 
    \rotatebox[origin=c]{90}{\textbf{VGG}} \\[20mm]
    \rotatebox[origin=c]{90}{\textbf{ResNet}}
    \end{tabular}
    &
    \includegraphics[align=c, width=0.9\columnwidth]{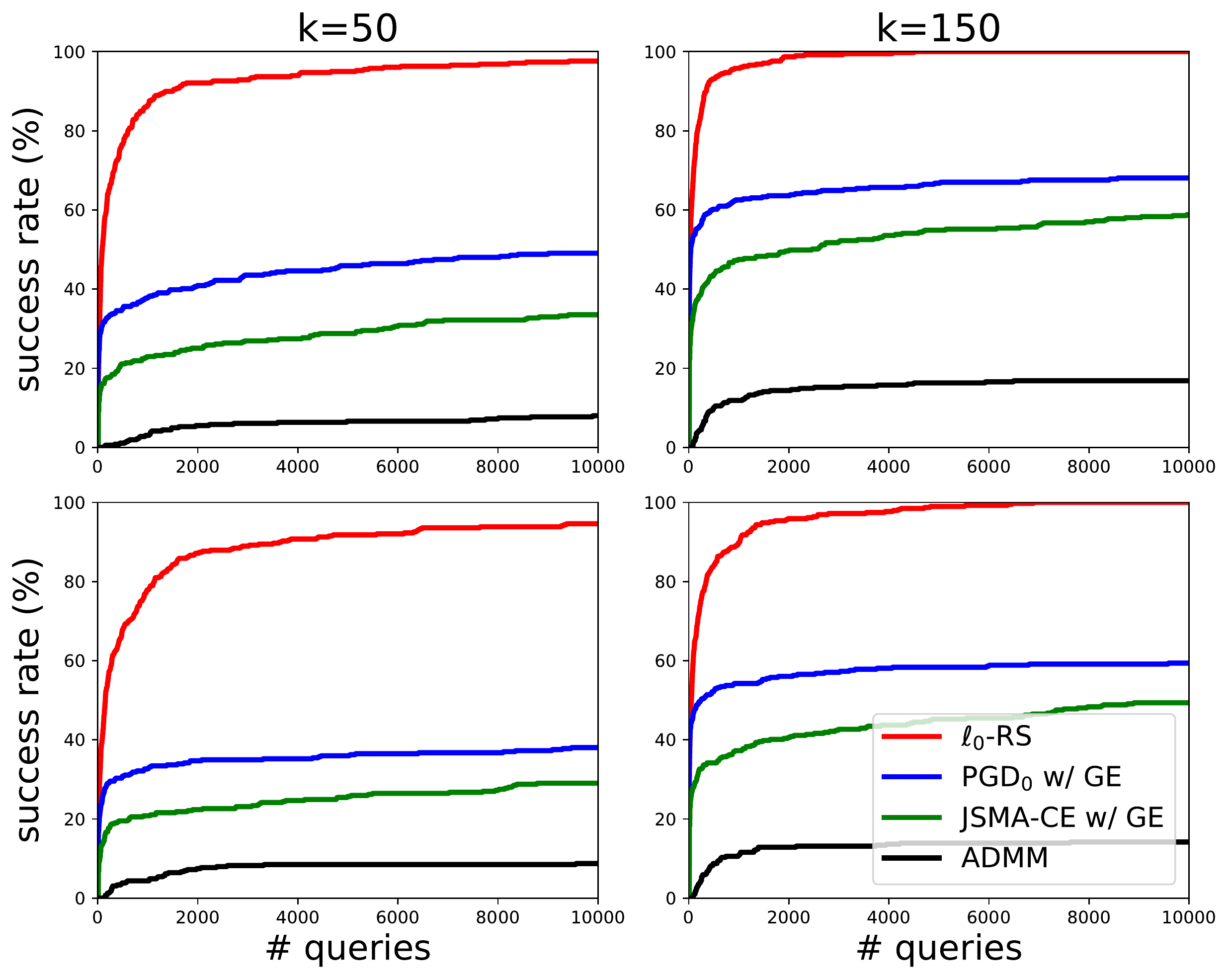}\\
    \end{tabular}
    \caption{Progression of the 
    success rate vs number of queries for black-box pixel-based $l_0$-attacks on ImageNet in the untargeted setting. %
    At all sparsity levels \texttt{$l_0$-RS} (red) outperforms PGD\textsubscript{$0$} (blue) and %
    JSMA-CE (green) with gradient estimation and
    ADMM attack %
    (black).
    }
    \label{fig:succ_rate_imagenet}
\end{figure}
\begin{figure}[t] \centering 
    \setlength{\tabcolsep}{1pt}
    \includegraphics[clip, trim=0mm 280mm 0mm 0mm, width=0.99\columnwidth]{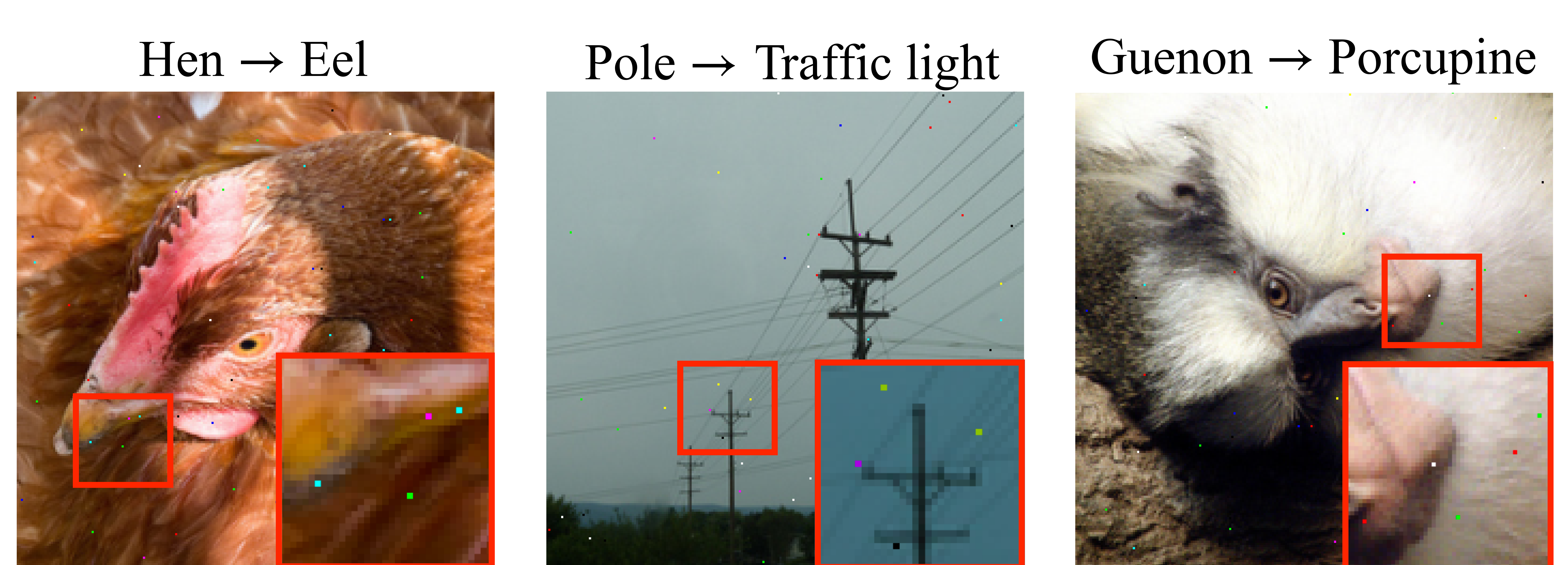}
    \caption{%
    Untargeted $l_0$-adversarial examples generated by our pixel-based \texttt{$l_0$-RS} algorithm for $k=50$ pixels.} \label{fig:l0_visualizations}%
\end{figure}

\textbf{Results.}
We show in Fig.~\ref{fig:succ_rate_imagenet} the success rate vs the number of queries for all black-box attacks. In all cases, \texttt{$l_0$-RS} outperforms its competitors 
in terms of the final success rate by a large margin---the second best method (PGD\textsubscript{$0$} w/ GE) is at least 30\% worse. %
Moreover, \texttt{$l_0$-RS} is query efficient as it achieves results close to the final ones already with a low number of queries. For example, on VGG with $k=150$, \texttt{$l_0$-RS} achieves $100\%$ of success rate using on average only 171 queries, with a median of 25. %
Unlike other methods, \texttt{$l_0$-RS} can achieve almost 100\% success rate by perturbing 50 pixels which is \textit{only} 0.1\% of the total number of pixels. We visualize the adversarial examples of \texttt{$l_0$-RS} in Fig.~\ref{fig:l0_visualizations}.

\subsection{Using $l_0$-RS for accurate robustness evaluation}
In this section, our focus is the accurate evaluation of robustness in the $l_0$-threat model. For this, we evaluate existing white-box methods and black-box methods together. %
Instead of the success rate taken only over correctly classified examples, here we rather consider \textit{robust error} (similarly to \cite{MadEtAl2018}), which is defined as the classification error on the adversarial examples crafted by an attacker. %
\begin{table}
    \centering \small
    \begin{tabular}{c c  c c }
        \setlength\extrarowheight{5pt} \tabcolsep=6pt
        \textit{attack} & \textit{type} & VGG & ResNet \\
\toprule \addlinespace[2mm] %
\rowcolor{lightgrey}
\multicolumn{4}{c}{$l_0$-bound in \textbf{pixel space} $k=50$}\\ \midrule
JSMA-CE %
        & white-box & 42.6\% & 39.6\% \\
PGD\textsubscript{$0$} %
        & white-box & 87.0\% & 81.2\% \\
\hdashline
ADMM %
        & black-box & 30.3\% & 29.0\% \\
        JSMA-CE with GE & black-box & 49.6\% & 44.8\% \\
        PGD\textsubscript{$0$} %
        with GE & black-box & 61.4\% & 51.8\% \\
        CornerSearch$^*$ %
        & black-box & 82.0\% & 72.0\% \\
        \texttt{$l_0$-RS} & black-box & \textbf{98.2\%} & \textbf{95.8\%}\\[2mm]
\multicolumn{4}{c}{$l_0$-bound in \textbf{feature space} $k=50$}\\ \midrule
SAPF$^*$ %
 & white-box & 21.0\% & 18.0\% \\
ProxLogBarrier & white-box & 33.0\% &28.4\%\\
EAD & white-box & 39.8\% & 35.6\% \\
SparseFool & white-box & 43.6\% & 42.0\% \\
VFGA & white-box & 58.8\% & 55.2\% \\
FMN & white-box & 83.8\% & 77.6\% \\
PDPGD & white-box &  89.6\% & 87.2\% \\
\hdashline
ADMM & black-box & 32.6\% & 29.0\% \\
CornerSearch$^*$ & black-box & 76.0\% & 62.0\% \\
\texttt{$l_0$-RS} & black-box & \textbf{92.8\%} & \textbf{88.8\%} \\
\bottomrule
\end{tabular}\caption{\textbf{ImageNet:} Robust test error of %
$l_0$-attacks. 
    The entries with $^*$ are evaluated on 100 points instead of 500 because of their high computational cost. All black-box attacks use 10k queries except CornerSearch which uses 600k.  \texttt{$l_0$-RS} outperforms all black- \emph{and} white-box attacks.
    }\label{tab:l0_white_box} \end{table}

\textbf{White-box attacks on ImageNet.}
We test the robustness of the ImageNet models introduced in the previous section to $l_0$-bounded perturbations. %
As competitors we consider multiple white-box attacks which minimize the $l_0$-norm in \textit{feature space}: SAPF~\cite{fan2020sparse}, ProxLogBarrier~\cite{pooladian2019proxlogbarrier}, EAD \cite{CheEtAl2018}, SparseFool \cite{ModEtAl19}, VFGA \cite{cesaire2020stochastic}, FMN \cite{pintor2021fast} and PDPGD \cite{matyasko2021pdpgd}. %
For the $l_0$-threat model in \textit{pixel space} %
we use two white-box baselines:  PGD\textsubscript{0} %
\cite{croce2019sparse}, and JSMA-CE %
\cite{papernot2016limitations} (where we use the gradient of the cross-entropy loss to generate the saliency map).  %
Moreover, we show the results of the black-box attacks from the previous section (all with a query limit of 10,000), and additionally use the black-box CornerSearch %
for which we use a query limit of 600k and which is thus only partially comparable. Details of the attacks are available in App.~\ref{sec:app_image_classification}. Table~\ref{tab:l0_white_box} shows the robust error given by all competitors: \texttt{$l_0$-RS} achieves the best results for pixel and feature based $l_0$-threat model on both VGG and ResNet, outperforming black- \emph{and} white-box attacks. 
We note that while the PGD attack has been observed to give accurate robustness estimates for $l_\infty$- and $l_2$-norms \cite{MadEtAl2018}, this is not the case for the $l_0$ constraint set. This is due to the discrete structure of the $l_0$-ball which is not amenable for continuous optimization.

\begin{table}
    \centering \small
    \begin{tabular}{c c  c c }
        \setlength\extrarowheight{5pt} \tabcolsep=6pt
        \textit{attack} & \textit{type} & $l_2$-AT ResNet & $l_1$-AT ResNet \\
\toprule \addlinespace[2mm] %
\rowcolor{lightgrey}
\multicolumn{4}{c}{$l_0$-bound in \textbf{pixel space} $k=24$}\\ \midrule PGD\textsubscript{0} & white-box & 68.7\% &72.7\% \\
\hdashline
CornerSearch & black-box & 59.3\%& 64.9\% \\
\texttt{$l_0$-RS} & black-box & \textbf{85.7}\% & \textbf{81.0}\% \\[2mm]

\multicolumn{4}{c}{$l_0$-bound in \textbf{feature space} $k=24$}\\ \midrule
VFGA & white-box & 40.5\% & 27.5\%\\
FMN & white-box & 52.9\% & 28.2\% \\
PDPGD & white-box & 46.4\% & 26.9\%  \\
\hdashline
CornerSearch & black-box & 43.2\% &29.4\% \\
\texttt{$l_0$-RS} & black-box & \textbf{63.4\%} & \textbf{38.3\%} \\
\bottomrule
\end{tabular} \caption{\textbf{CIFAR10:} Robust test error of untargeted $l_0$-attacks in pixel and feature space on a $l_2$- resp. $l_1$-AT model.} \label{tab:l0_cifar10}
\end{table}
\textbf{Comparison on CIFAR-10.} In Table \ref{tab:l0_cifar10} we compare the strongest white- and black-box attacks on $l_1$- resp. $l_2$-adversarially trained %
PreAct ResNet-18 on CIFAR-10 from \cite{croce2021mind} and \cite{rebuffi2021fixing} (details in App.~\ref{app:cifar10_models}). We keep the same computational budget used on ImageNet. As before, we consider perturbations with $l_0$-norm $k=24$ in pixel or feature space: in both cases \texttt{$l_0$-RS} achieves the highest robust test error outperforming even all white-box attacks. Note that, as expected, the model robust wrt $l_1$ is less vulnerable to $l_0$-attacks especially in the feature space, whose $l_1$-norm is close to that used during training. 

\indent\textbf{Robust generative models on MNIST.} \label{sec:mnist_l0}
\cite{SchEtAl19} propose two robust generative models on MNIST, ABS and Binary ABS, which
showed high robustness against multiple types of $l_p$-bounded adversarial examples. These %
classifiers rely on optimization-based inference using a variational auto-encoder (VAE) with 50 steps of gradient descent for each prediction (times 1,000 repetitions). It is too expensive to get gradients with respect to the input through the optimization process, thus \cite{SchEtAl19} evaluate only black-box attacks,
and test $l_0$-robustness with sparsity $k=12$ using their proposed Pointwise Attack with 10 restarts.
We evaluate on both models CornerSearch %
with a budget of 50,000 queries %
and \texttt{$l_0$-RS} with an equivalent budget of 10,000 queries and $5$ random restarts.
Table~\ref{tab:l0_mnist} summarizes the robust test error (on 100 points) achieved by the attacks (the results of Pointwise Attack are taken from \cite{SchEtAl19}). For both classifiers, $l_0$-RS yields the strongest evaluation of robustness suggesting that the ABS models are less robust than previously believed. This illustrates that despite we have \textit{full access} to the attacked VAE model, a strong \textit{black-box} $l_0$-attack can still be useful for an accurate robustness evaluation.
\begin{table}[t]
    \centering \small
    \begin{tabular}{c c | c c }
        \setlength\extrarowheight{5pt}
        \multirow{2}{*}{\textit{attack}} & \multirow{2}{*}{\textit{type}} & \multicolumn{2}{c}{$k=12$ (pixels)}\\
        & & ABS & Binary ABS\\
        \toprule
        Pointwise Attack %
        & black-box & 31\%& 23\% \\
        CornerSearch %
        & black-box & 29\% & 28\% \\
        \texttt{$l_0$-RS} & black-box & \textbf{55\%} & \textbf{51\%} \\
\bottomrule
\end{tabular}%
\caption{Robust test error on robust models \cite{SchEtAl19} on MNIST by different attacks on 100 test points.} \label{tab:l0_mnist}%
\end{table}

\textbf{\texttt{$l_0$-RS} on malware detection.} We apply our method on a malware detection task and show its effectiveness in App.~\ref{sec:app_malware_detection}.

\subsection{Theoretical analysis of $l_0$-RS} \label{sec:theory}
Given the empirical success of \texttt{$l_0$-RS}, here we analyze it theoretically for a binary classifier. While the analysis does not directly transfer to neural networks,
most modern neural network architectures result in piecewise linear classifiers \cite{AroEtAl2018}, so that the result approximately holds in a sufficiently small neighborhood of the target point $x$. 

As in the malware detection task in App.~\ref{sec:app_malware_detection}, we assume that the input $x$ has binary features, $x \in \{0, 1\}^d$, and we denote the label by  $y \in \{-1, 1\}$ and the gradient of the linear model by $w_x \in \R^d$. Then the Problem \eqref{eq:adv_opt} of finding the optimal $l_0$ adversarial example is equivalent to:
\begin{align*}
    \argmin_{\substack{\norm{\delta}_0 \leq k \\ 
                      x_i + \delta_i \in \{0, 1\}}} 
        \hspace{-4pt} y\inner{w_x, x+\delta} 
        &= 
    \hspace{-4pt} \argmin_{\substack{\norm{\delta}_0 \leq k \\ 
                            \delta_i \in \{0, 1 - 2 x_i\}}}  
        \hspace{-7pt} \inner{yw_x, \delta} \\
        &= 
    \argmin_{\substack{\norm{\delta}_0 \leq k \\
                            \delta_i \in \{0, 1\}}} 
        \hspace{3pt} \langle \underbrace{yw_x\odot(1 - 2 x)}_{\hat{w}_x}, \delta \rangle ,
\end{align*}
where $\odot$ denotes the elementwise product.
In the white-box case, i.e. when $w_x$ is known, the solution is to simply set $\delta_i = 1$ for the $k$ smallest weights of $\hat{w}_x$. 
The black-box case, where $w_x$ is unknown and we are only allowed to query the model predictions $\inner{\hat{w}_x, z}$ for any $z \in \R^d$, is more complicated since the naive weight estimation algorithm requires $O(d)$ queries to first estimate $\hat{w}_x$ and then to perform the attack by selecting the $k$ minimal weights. This naive approach is prohibitively expensive for high-dimensional datasets (e.g., $d = 150{,}528$ on ImageNet assuming $224\times224\times3$ images). 
However, the problem of generating adversarial examples does not have to be always solved exactly, and often it is enough to find an approximate solution. 
Therefore we can be satisfied with only identifying  $k$ among the $m$ smallest weights. Indeed, the focus is not on exactly identifying the solution but rather on having an algorithm that in expectation requires a \textit{sublinear} number of queries. 
With this goal, we show that \texttt{$l_0$-RS} satisfies this requirement for large enough $m$.
\begin{proposition}
\label{prop:l0_rs}
    The expected number $t_k$ of queries needed for \texttt{$l_0$-RS} with $\iter{\alpha}{i} = \nicefrac{1}{k}$ to find a set of $k$ weights out of the smallest $m$ weights of a linear model is:  
    \begin{align*}
        \Exp \left[ t_k \right] = (d-k)k\sum_{i=0}^{k-1} \frac{1}{(k-i)(m-i)}  <  (d-k)k \frac{\ln(k) + 2}{m-k}.
    \end{align*}
\end{proposition}

\begin{table*}[ht!]
    \centering \small
    \setlength{\tabcolsep}{4.5pt}
    \begin{tabular}{p{1.45cm} l c  | c c  c |c  c c } %
    &&\multirow{2}{*}{\textit{attack}} &  %
    \multicolumn{3}{c}{VGG} & \multicolumn{3}{c}{ResNet}\\
    & & &\textit{success rate} & \textit{mean queries} & \textit{med. queries} & \textit{success rate} & \textit{mean queries} & \textit{med. queries}\\
    \toprule
    \multirow{6}{*}{\begin{tabular}{c} \textbf{untarget.}\\ 
    $20\times 20$
    \end{tabular}} & \multirow{5}{*}{
    \rotatebox[origin=c]{90}{black-box}}&
    LOAP %
    w/ GE & 55.1\% $\pm$ 0.6 & 5879 $\pm$ 51 & 7230 $\pm$ 377 &40.6\% $\pm$ 0.1 & 6870 $\pm$ 10 & 10000 $\pm$ 0\\
    & &TPA %
    & %
      46.1\% $\pm$ 1.1 & 6085\textsuperscript{*} $\pm$ 83 & 8080\textsuperscript{*} $\pm$ 1246 & 49.0\% $\pm$ 1.2 & 5722\textsuperscript{*} $\pm$ 64 & 5280\textsuperscript{*} $\pm$ 593\\
    & &\texttt{Sparse-RS} + SH %
    & %
     82.6\% & 2479 & 514 & 75.3\% & 3290 & 549 \\
    &&\texttt{Sparse-RS} + SA %
    & %
     85.6\% $\pm$ 1.1 & 2367 $\pm$ 83 & 533 $\pm$ 40  & 78.5\% $\pm$ 1.0 & 2877 $\pm$ 64 & 458 $\pm$ 43\\ %
    &&\texttt{Patch-RS} & %
     \textbf{87.8\% $\pm$ 0.7} & \textbf{2160 $\pm$ 44} & \textbf{429 $\pm$ 22} & \textbf{79.5\% $\pm$ 1.4} & \textbf{2808 $\pm$ 89} & \textbf{438 $\pm$ 68}\\
    \cdashline{2-9}
    && White-box LOAP %
    & 98.3\% & - &- &82.2\%&-&-\\
    \midrule
    \multirow{6}{*}{\begin{tabular}{c} \textbf{targeted}\\ $40\times40$
    \end{tabular}} & \multirow{5}{*}{
    \rotatebox[origin=c]{90}{black-box}}& LOAP %
    w/ GE&23.9\% $\pm$ 0.9 & 44134 $\pm$ 71 & 50000 $\pm$ 0 & 18.4\% $\pm$ 0.9 & 45370 $\pm$ 88 & 50000 $\pm$ 0\\
    & &TPA %
    & 5.1\% $\pm$ 1.2 & 29934\textsuperscript{*} $\pm$ 462 & 34000\textsuperscript{*} $\pm$ 0 & 6.0\% $\pm$ 0.5 & 31690\textsuperscript{*} $\pm$ 494 & 34667\textsuperscript{*} $\pm$ 577\\
    && \texttt{Sparse-RS} + SH %
     & 63.6\% & 25634 & 19026 & 48.6\% & 31250 & 50000 \\
    &&\texttt{Sparse-RS} + SA %
    & %
     70.9\% $\pm$ 1.2 & 23749 $\pm$ 346 & 15569 $\pm$ 568 & 53.7\% $\pm$ 0.9 & 32290 $\pm$ 239 & 40122 $\pm$ 2038\\
    & &\texttt{Patch-RS} & %
     \textbf{72.7\% $\pm$ 0.9} & \textbf{22912 $\pm$ 207} & \textbf{14407 $\pm$ 866} & \textbf{55.6\% $\pm$ 1.5} & \textbf{30290 $\pm$ 317} & \textbf{34775 $\pm$ 2660}\\
    \cdashline{2-9}
    & &White-box LOAP %
    & 99.4\% & - & - & 94.8\%&-& - \\
    \bottomrule
    \end{tabular}
    \caption{%
    Success rate and query statistics of image-specific patches.
    Black-box attacks are given 10k/50k queries for untargeted/targeted case. %
    SH %
    is a deterministic method. The query statistics are computed on \textit{all} images with 5 random seeds. $^*$ TPA uses an early stopping mechanism to save queries, thus %
    might not use all queries. \texttt{Patch-RS} outperforms all other methods in success rate and query efficiency.}
    \label{tab:image_spec_large}
\end{table*}
\begin{figure*}[ht!] \centering 
    \setlength{\tabcolsep}{1pt}
    \begin{tabular}{ccc|ccc}
        \multicolumn{3}{c|}{\textbf{Untargeted patches}} & \multicolumn{3}{c}{\textbf{Targeted patches}} \\
        \scriptsize Parrot $\rightarrow$ Torch & 
        \scriptsize Castle $\rightarrow$ Traffic light & 
        \scriptsize Speedboat $\rightarrow$ Amphibian & 
        \scriptsize Bee eater $\rightarrow$ Tiger cat & 
        \scriptsize Newfoundland $\rightarrow$ Bucket & 
        \scriptsize Geyser $\rightarrow$ Racer \\
        \includegraphics[width=0.34\columnwidth]{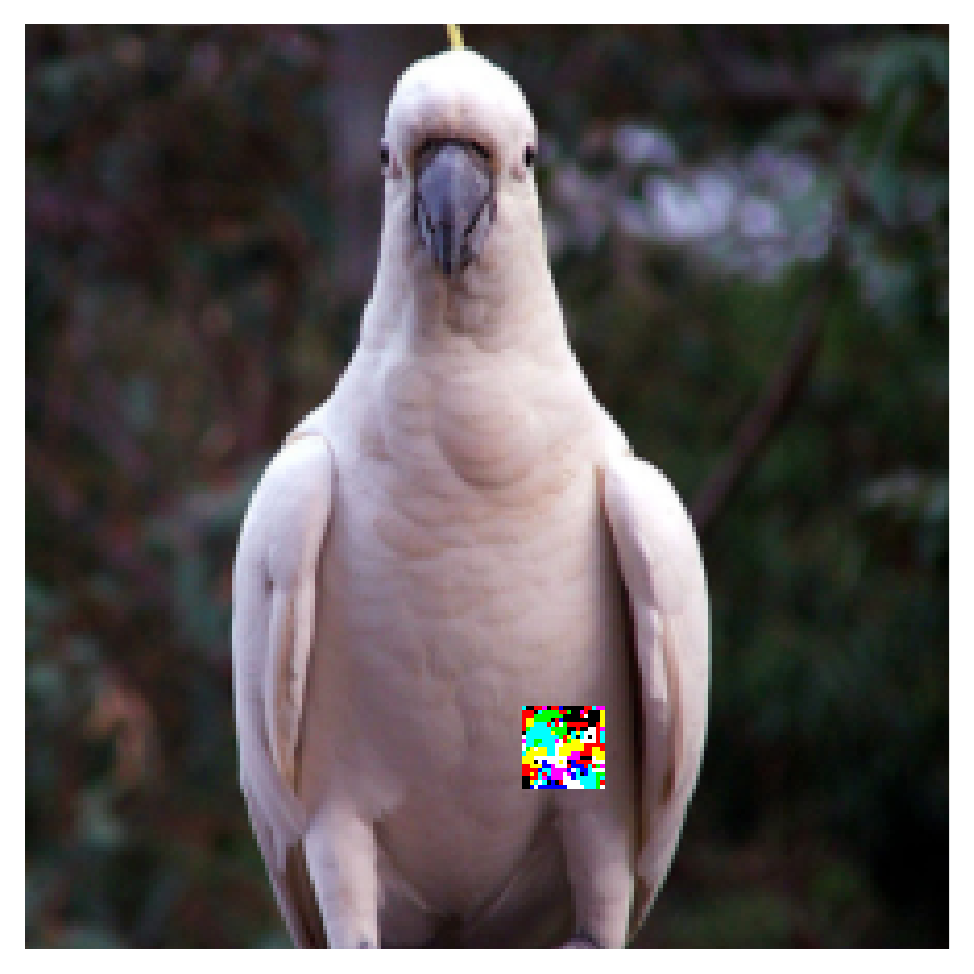} &
        \includegraphics[width=0.34\columnwidth]{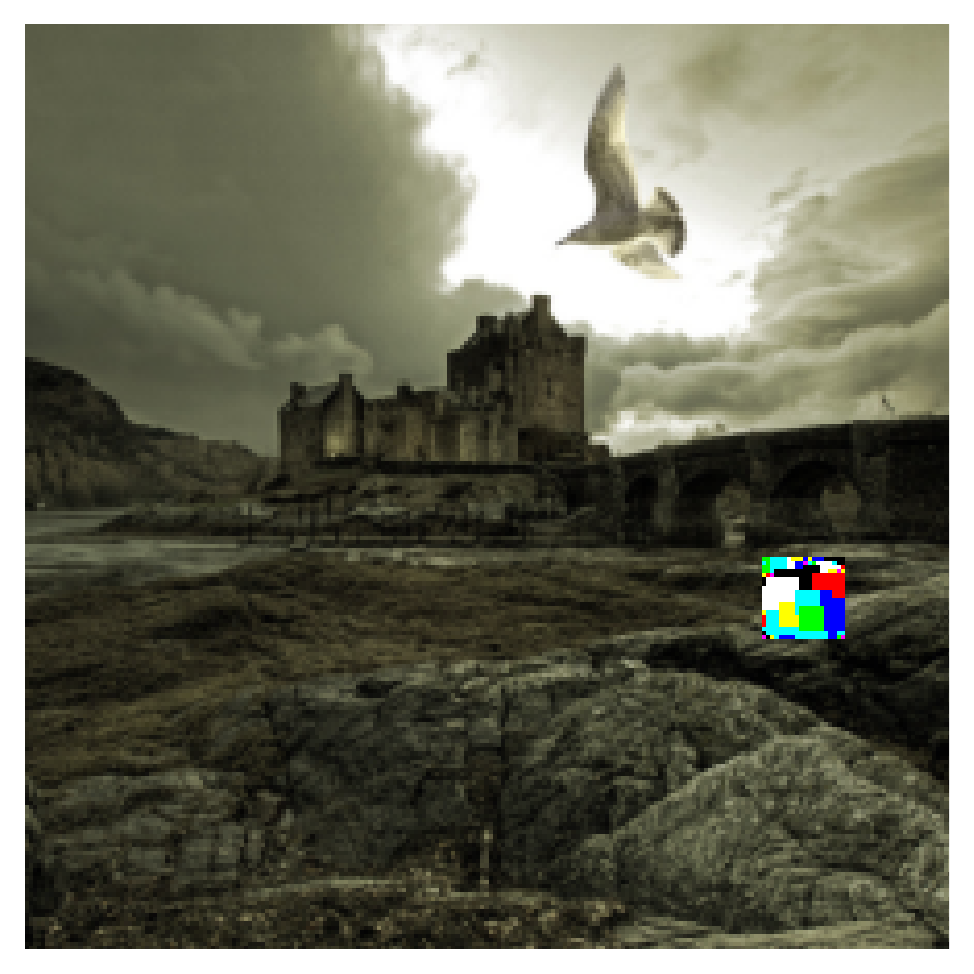} &
        \includegraphics[width=0.34\columnwidth]{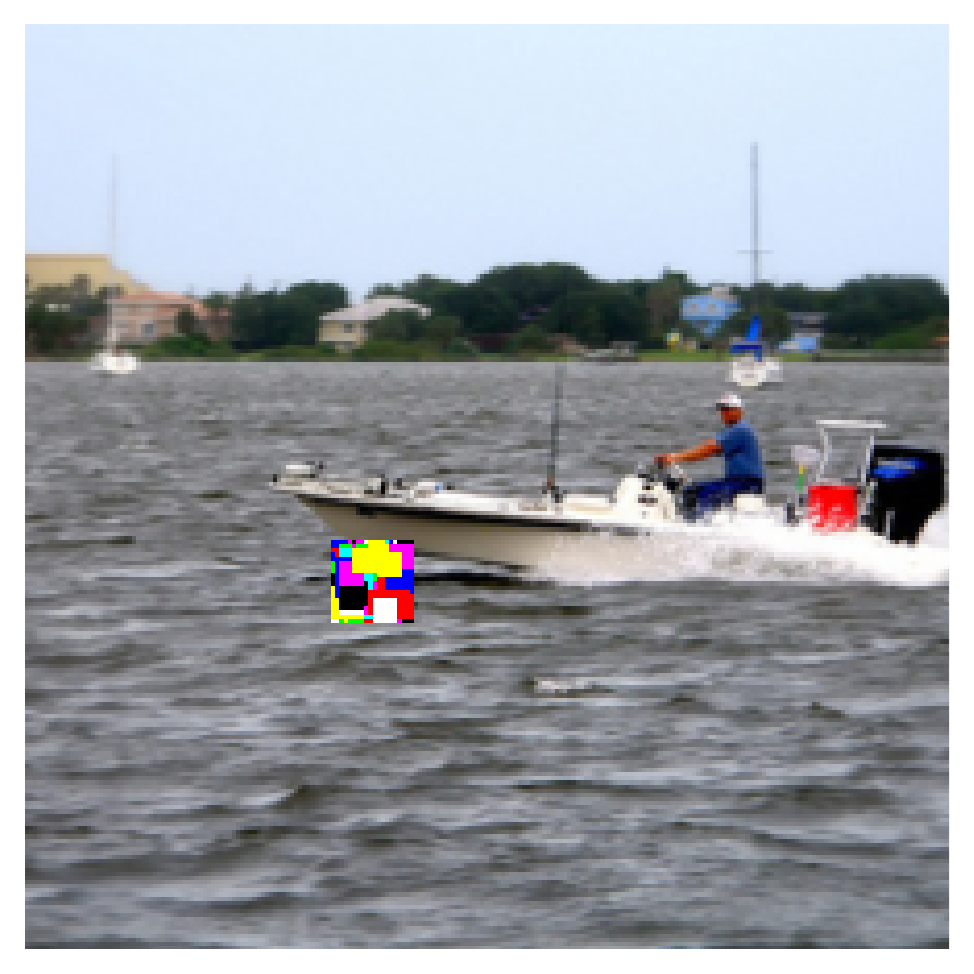} &
        \includegraphics[width=0.34\columnwidth]{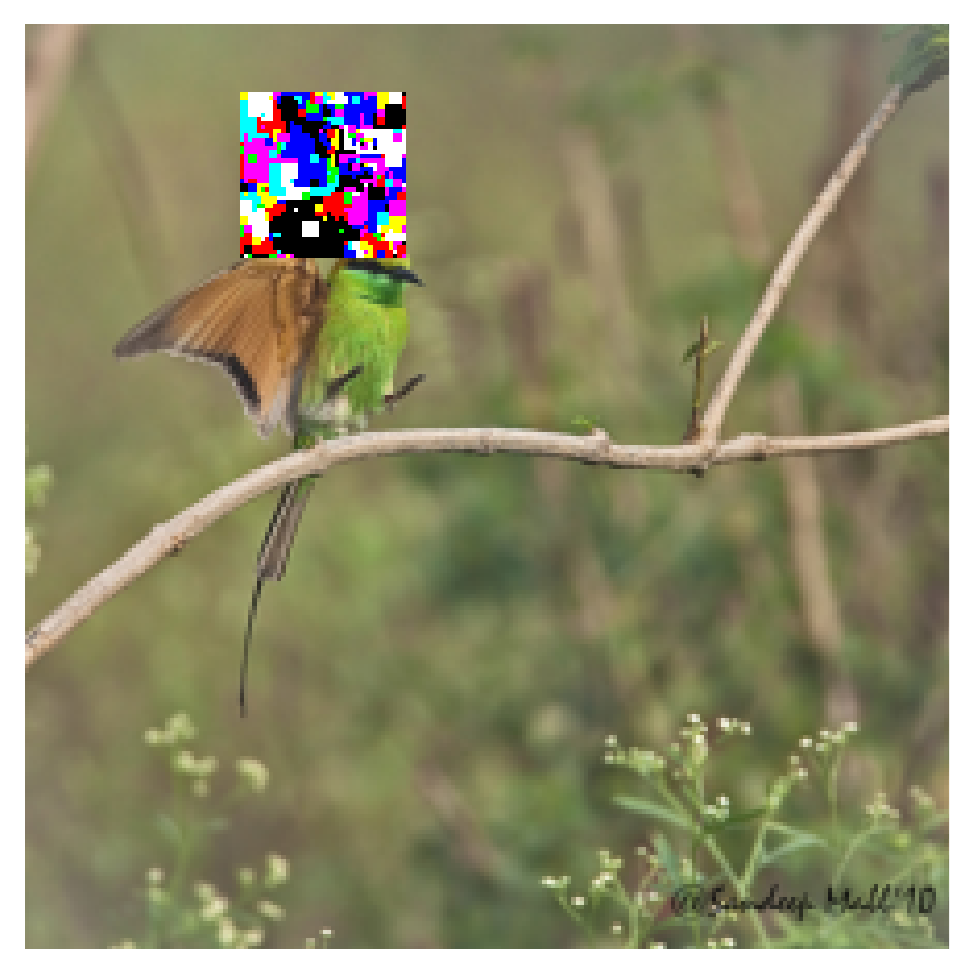} &
        \includegraphics[width=0.34\columnwidth]{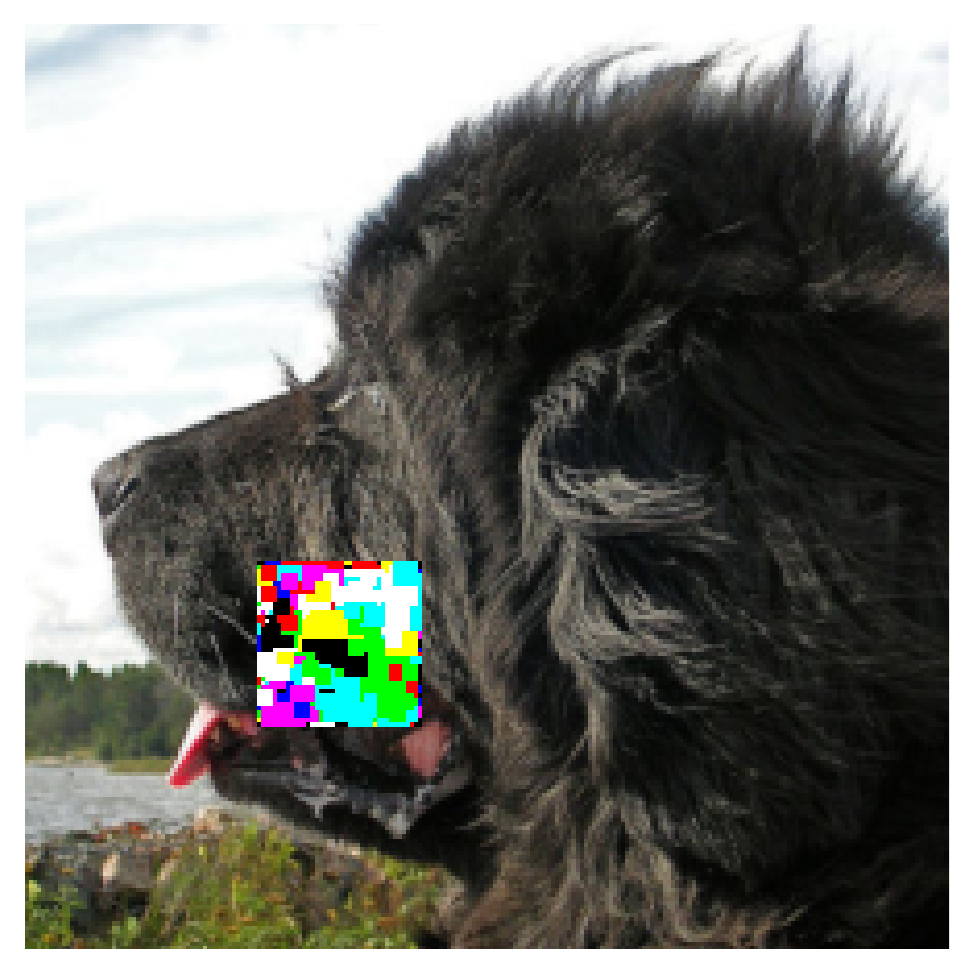} &
        \includegraphics[width=0.34\columnwidth]{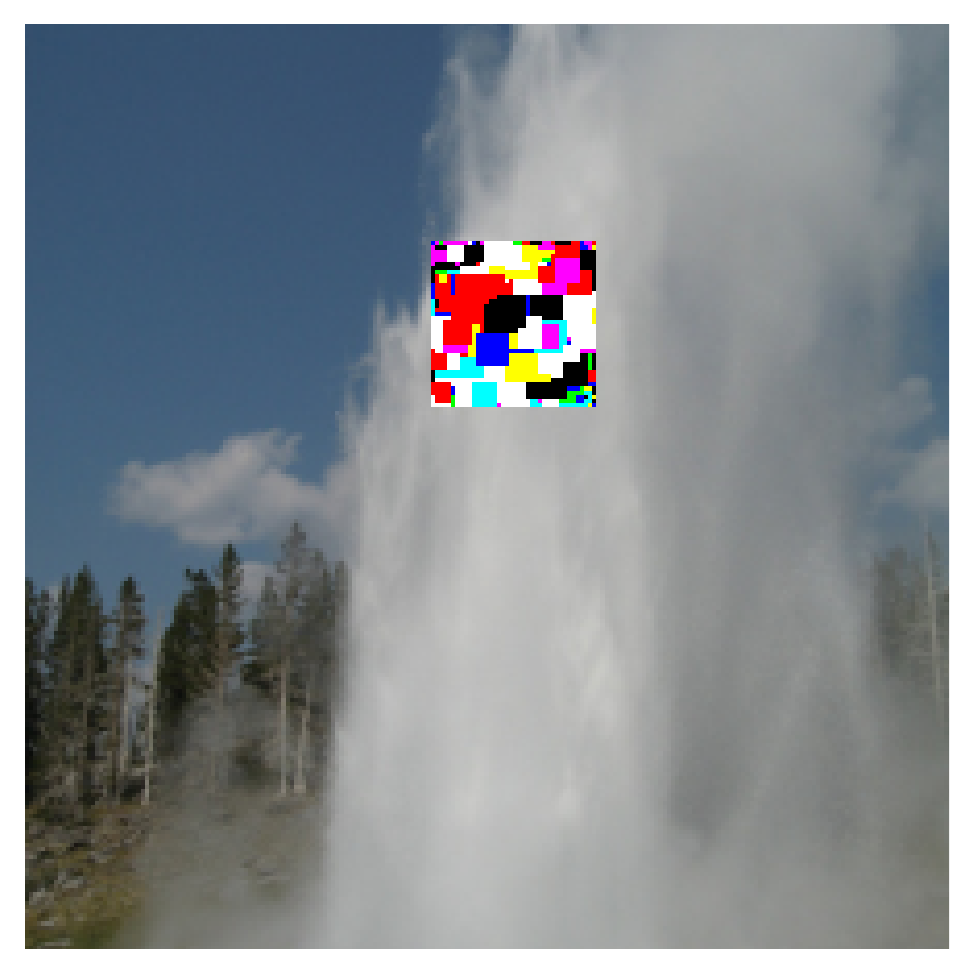} 
    \end{tabular}
    \caption{
    Image-specific untargeted ($20\times20$ pixels) and targeted ($40\times40$) patches generated by our \texttt{Patch-RS} algorithm.} \label{fig:image_spec_patches_visualizations}%
\end{figure*}
The proof is deferred to \ifpaper App.~\ref{sec:app_theory} \else the supplement\fi and resembles that of the coupon collector problem. %
For non-linear models, \texttt{$l_0$-RS} uses $\iter{\alpha}{i} > \nicefrac{1}{k}$ for better exploration initially, but then progressively reduces it.
The main conclusion from
Proposition~\ref{prop:l0_rs} is that $\Exp \left[ t_k \right]$ becomes %
sublinear for large enough gap $m-k$, as we %
illustrate in Fig.~\ref{fig:theory_exps} in App.~\ref{sec:app_theory}. %

\label{sec:image_specific_patches}

\section{\texttt{Sparse-RS} for adversarial patches} \label{sec:patches}

Another type of sparse attacks which recently received attention are adversarial patches introduced by \cite{BroEtAl2017}. There the perturbed pixels are localized, often square-shaped, and limited to a small portion of the image but can be changed in an arbitrary way.
While some works \cite{BroEtAl2017,karmon2018lavan} aim at universal patches, which fool the classifier regardless of the image and the position where they are applied, which we consider in Sec. \ref{sec:universal}, we focus first on image-specific patches as in \cite{yang2020patchattack, rao2020adversarial} where one optimizes both the content and the location of the patch for each image independently.

\textbf{General algorithm for patches.}
Note that both location (step 6 in Alg.~\ref{alg:rs_general}) and content (step~7 in Alg.~\ref{alg:rs_general})
have to be optimized in \texttt{Sparse-RS}, and on each iteration we check only one of these updates.
We test the effect of different frequencies of location/patch updates in an ablation study in App.~\ref{sec:app_abl_patches}.
Since the location of the patch is a discrete variable, random search is particularly well suited for its optimization. For the location updates in step 6 in Alg.~\ref{alg:rs_general}, we randomly sample a new location in a 2D $l_\infty$-ball around the
current patch position (using clipping so that the patch is fully contained in the image). %
The radius of this $l_\infty$-ball %
shrinks with increasing iterations in order to perform progressively more local optimization (see App.~\ref{sec:app_image_specific_patches} for details).

For the update of the patch itself in step~7 in Alg.~\ref{alg:rs_general}, the only constraints are given by the input domain $[0, 1]^d$. Thus in principle any black-box method for an $l_\infty$-threat model can be plugged in there. We use Square Attack (SA) \cite{ACFH2019square} and SignHunter (SH) \cite{AlDujaili2019ThereAN} as they represent the state-of-the-art in terms of success rate and query efficiency. We integrate %
both in our framework and refer to them as \texttt{Sparse-RS}~+~SH and \texttt{Sparse-RS}~+~SA. Next we propose a novel random search based attack motivated by SA which together with our location update yields our novel \texttt{Patch-RS} attack. %

\textbf{\texttt{Patch-RS}.}
While SA and SH are state-of-the-art for $l_\infty$-attacks, they have been optimized for rather small perturbations whereas for patches all pixels can be manipulated arbitrarily in $[0,1]$. 
Here, we design an initialization scheme and a sampling distribution specific for adversarial patches. As initialization (step~2 of Alg.~\ref{alg:rs_general}), \texttt{Patch-RS} uses randomly placed squares with colors in $\{0,1\}^3$, %
then it samples updates of the patch (step~7) with shape of squares, of size decreasing according to a piecewise constant schedule, until a refinement phase in the last iterations, 
when it performs single-channel updates (exact schedule in App.~\ref{sec:app_image_specific_patches}).
This is in contrast to SA where random vertical stripes are used as initialization and always
updates for all three channels of a pixel are sampled. %
The ablation study in App.~\ref{sec:app_abl_patches} shows how both modifications contribute to the improved performance of \texttt{Patch-RS}.

\textbf{Experiments.}
In addition to \texttt{Sparse-RS}~+~SH, \texttt{Sparse-RS}~+~SA, and \texttt{Patch-RS}, %
we consider two existing methods. i) TPA \cite{yang2020patchattack} which is a black-box attack aiming to produce image-specific adv. patches based on reinforcement learning. While \cite{yang2020patchattack} allows multiple patches for an image, we use TPA in the standard setting of a single patch. ii) Location-Optimized Adversarial Patches (LOAP) \cite{rao2020adversarial}, a white-box attack that uses PGD for the patch updates, which we combine with gradient estimation in order to use it in the black-box scenario %
(see App.~\ref{sec:app_image_specific_patches} for details). 
In Table~\ref{tab:image_spec_large} we report success rate, mean and median number of queries used %
for untargeted attacks with patch size $20 \times 20$ and query limit of 10,000 and for targeted attacks (random target class for each image) with patch size $40\times 40$ and maximally 50,000 queries. We attack 500 images of ImageNet with VGG and ResNet as target models. The query statistics are computed on \textit{all} 500 images, i.e. without restricting to only successful adversarial examples, as this
makes the query efficiency comparable for different success rates.
Our \texttt{Sparse-RS}~+~SH, \texttt{Sparse-RS}~+~SA and \texttt{Patch-RS} outperform existing methods by a large margin, showing the effectiveness of our scheme to optimize both location and patch. Among them, our specifically designed \texttt{Patch-RS} achieves the best results in all metrics.
We visualize its resulting adversarial examples in Fig.~\ref{fig:image_spec_patches_visualizations}.

\section{Universal adversarial patches}\label{sec:universal}
\begin{figure}[ht!] \centering 
    \setlength{\tabcolsep}{1pt}
    \begin{tabular}{ccc}
        \multicolumn{3}{c|}{\textbf{Targeted universal patches}} \\
        \scriptsize Butterfly $\rightarrow$ Rottweiler & 
        \scriptsize Persian cat $\rightarrow$ Slug & 
        \scriptsize Starfish $\rightarrow$ Polecat \\ 
        \includegraphics[clip, trim=1mm 1mm 1mm 1mm, width=0.31\columnwidth]{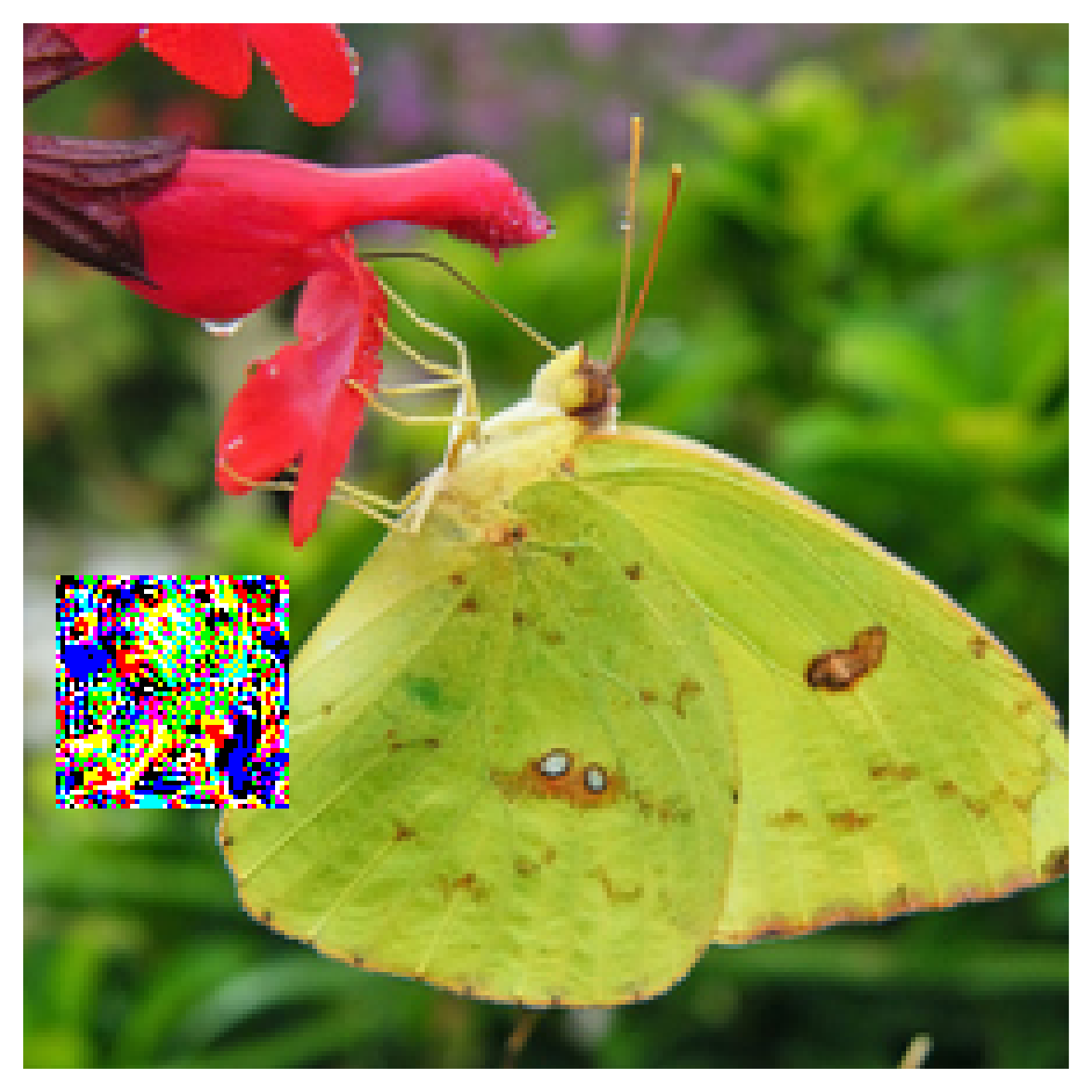} &
        \includegraphics[clip, trim=1mm 1mm 1mm 1mm, width=0.31\columnwidth]{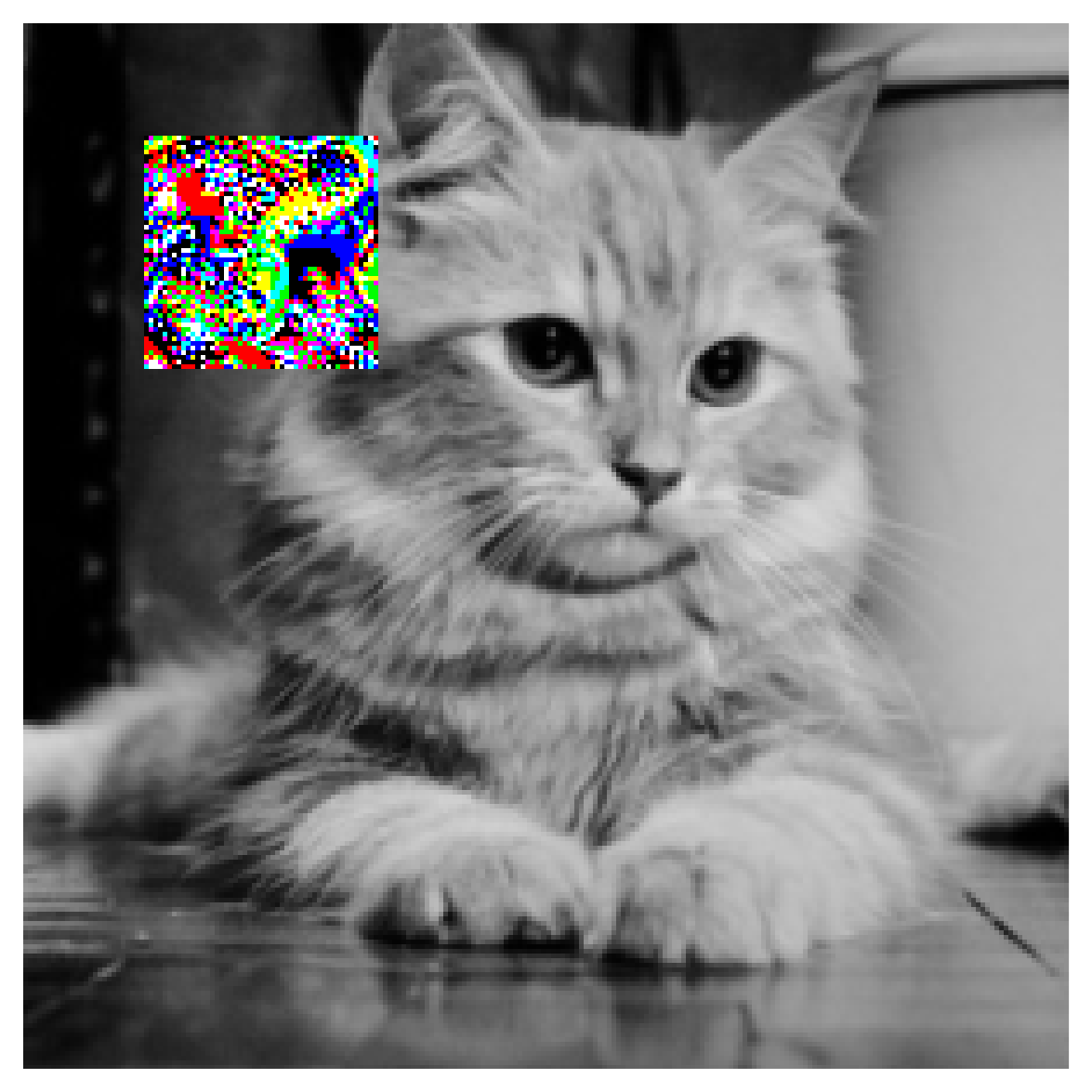} &
        \includegraphics[clip, trim=1mm 1mm 1mm 1mm, width=0.31\columnwidth]{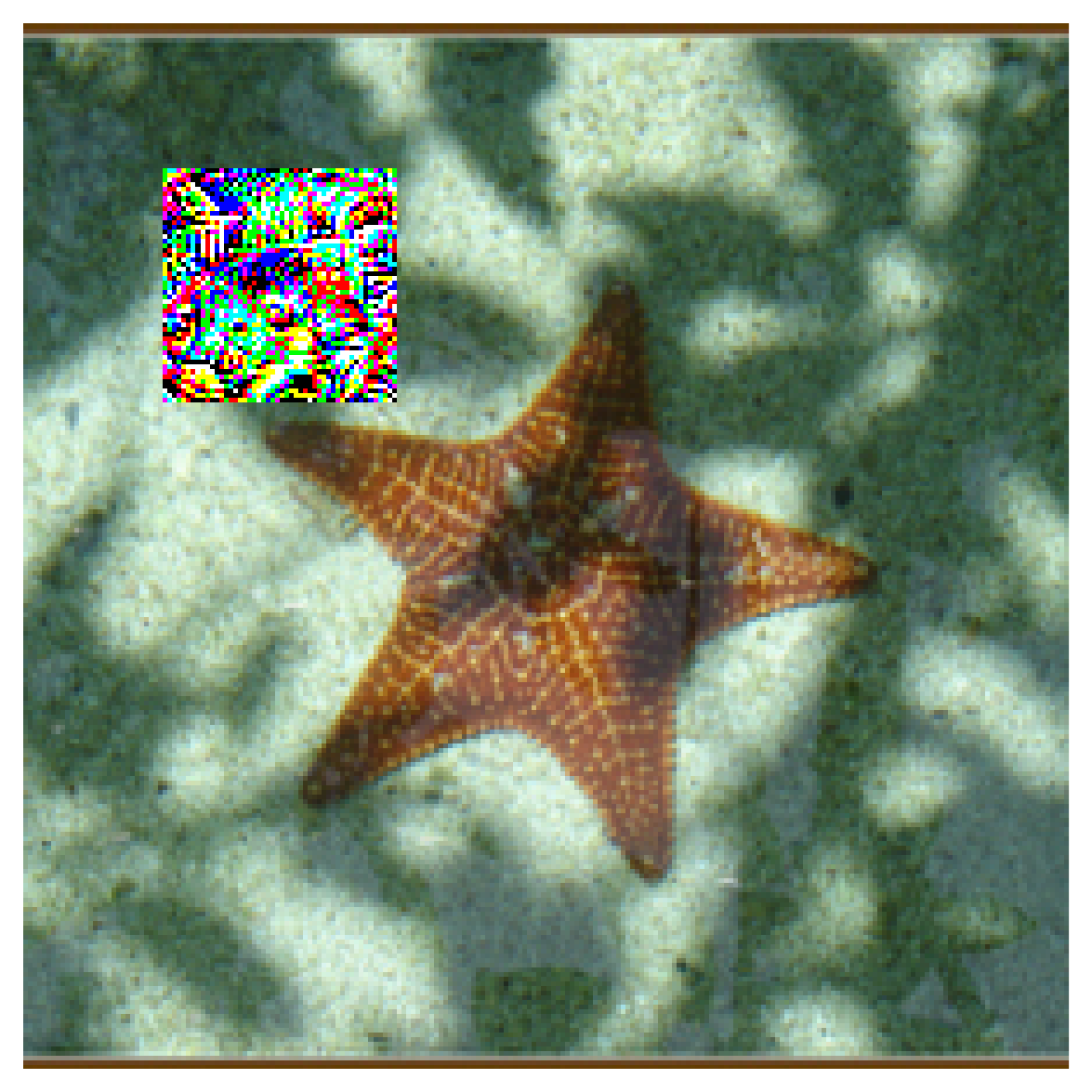} \\
        \scriptsize Echidna $\rightarrow$ Rottweiler & 
        \scriptsize Geyser $\rightarrow$ Slug & 
        \scriptsize Electric guitar $\rightarrow$ Polecat \\
        \includegraphics[clip, trim=1mm 1mm 1mm 1mm, width=0.31\columnwidth]{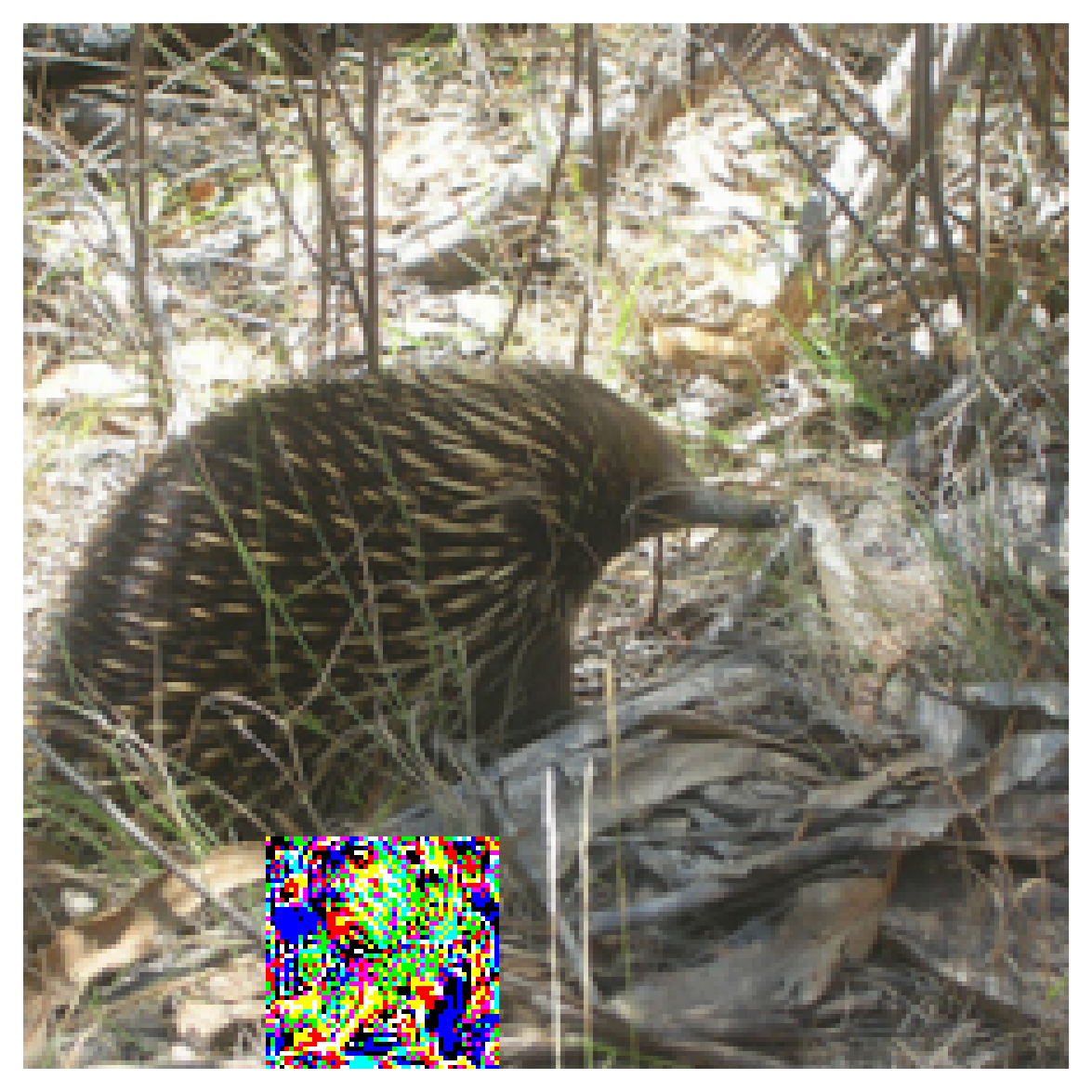} &
        \includegraphics[clip, trim=1mm 1mm 1mm 1mm, width=0.31\columnwidth]{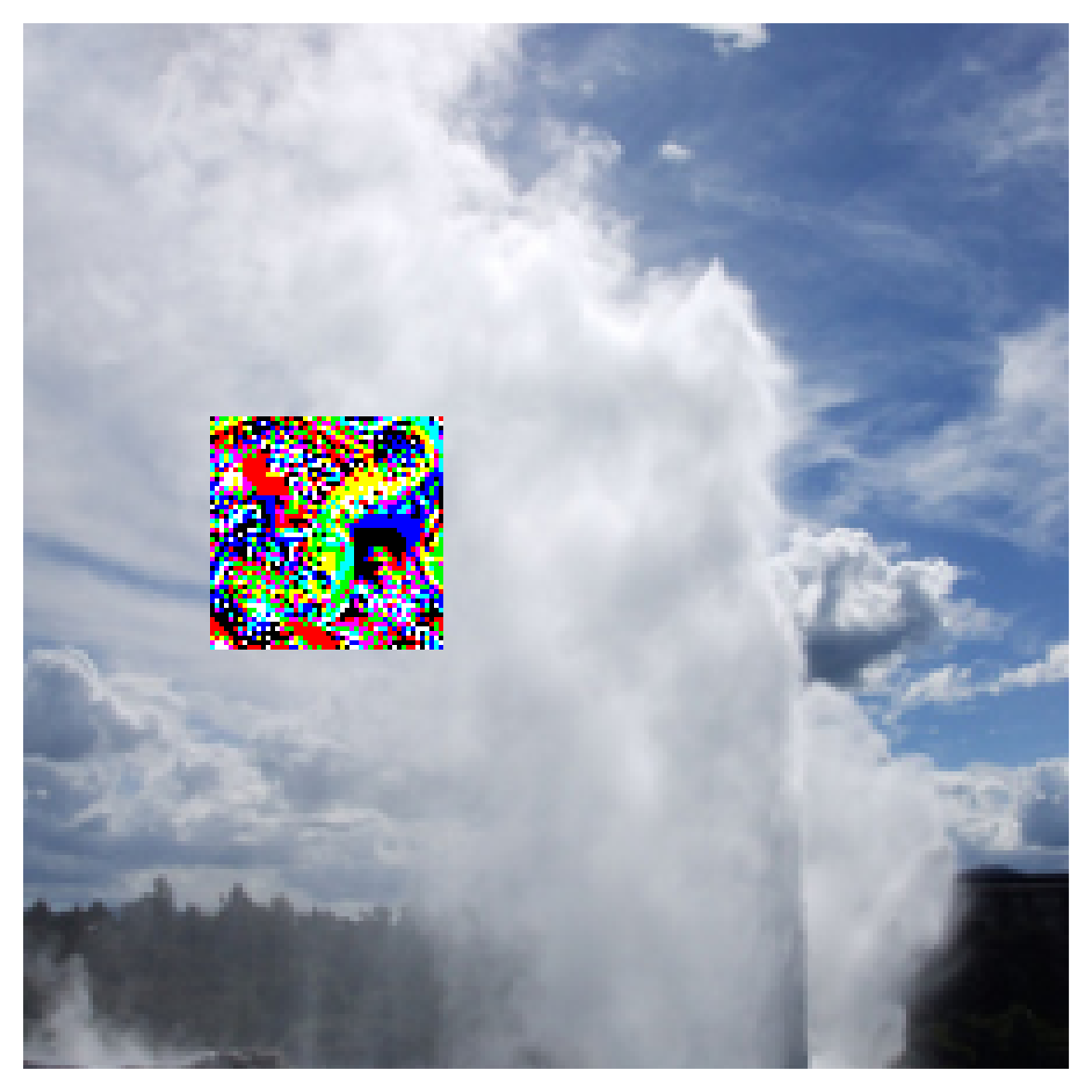} &
        \includegraphics[clip, trim=1mm 1mm 1mm 1mm, width=0.31\columnwidth]{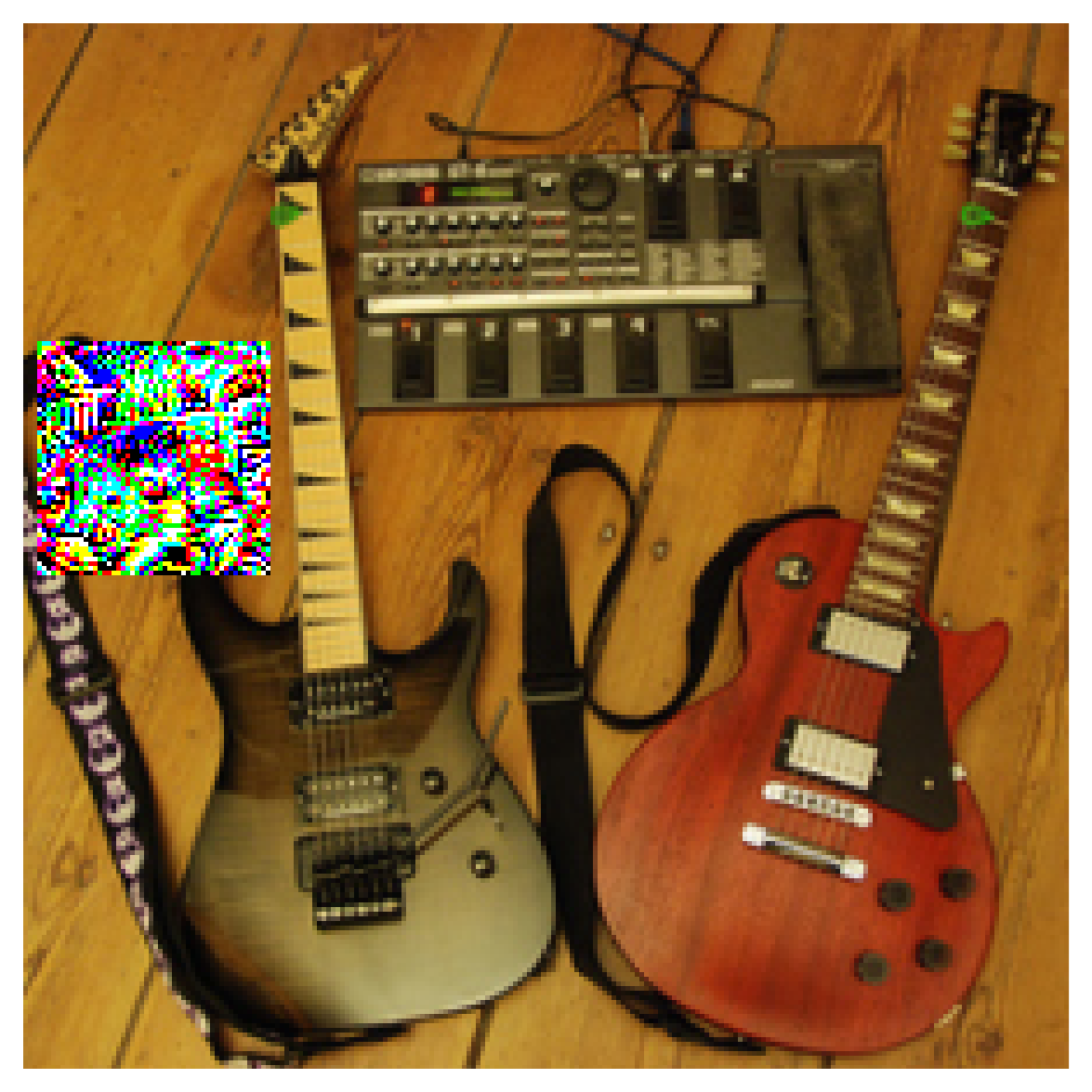}
    \end{tabular}
    \caption{
    We visualize two images in each column with the \textit{same} targeted universal patch generated by \texttt{Patch-RS} that changes the predictions to the desired target class. %
    } \label{fig:univ_patches_main}
\end{figure}

A challenging threat model is that of a black-box, targeted universal adversarial patch attack %
where %
the classifier should be fooled into a chosen target class when the patch is applied inside any image of some other class. 
Previous works rely on transfer attacks: in \cite{BroEtAl2017} %
the universal patch is created %
using a white-box attack on surrogate models, %
while the white-box attack of \cite{karmon2018lavan} directly optimizes the patch for the target model on a set of training images and then only tests generalization to unseen images. %
Our goal is a targeted black-box attack which crafts universal patches that generalize to unseen images when applied at random locations (see examples in Fig.~\ref{fig:univ_patches_main}). To %
our knowledge, this is the first method for this threat model %
which does not rely on a surrogate model.\\
\indent We employ Alg.~\ref{alg:rs_general} where for the creation of the patches in step~7 we use either SH, SA or our novel sampling distribution introduced in \texttt{Patch-RS} in Sec.~\ref{sec:patches}. The loss in Alg.~\ref{alg:rs_general} is computed on a small batch of 30 training images and the initial locations $M$ of the patch in each of the training images are sampled randomly. In order not to overfit on the training batch, %
we resample training images and locations of the patches (step 6 in Alg.~\ref{alg:rs_general}) every $10k$ queries %
(total query budget $100k$). %
As stochastic gradient descent this is a form of stochastic optimization of the population loss (expectation over images and locations) via random search.

\textbf{Experiments.} We apply the above scheme to \rsa{}~+~SH/SA and \texttt{Patch-RS} 
to create universal patches of size $50\times50$ for 10 random target classes on VGG (we repeat it for 3 seeds for RS-based methods). We compare to (1) the transfer-based attacks obtained via PGD \cite{MadEtAl2018} and MI-FGSM \cite{Dong_2018_CVPR} using ResNet as surrogate model, and to (2) ZO-AdaMM \cite{chen2019zo} based on gradient estimation. The results in Table~\ref{tab:universal} show that our \rsa{}~+~SH/SA and \texttt{Patch-RS} outperform other methods by large margin. We provide extra details and results for frames in App.~\ref{sec:app_univ}.

\begin{table}[t] 
\centering \small
\begin{tabular}{c c}
    \textit{attack} & VGG \\
    \toprule
    Transfer PGD %
    & 3.3\%\\
    Transfer MI-FGSM %
    & 1.3\%\\
    PGD w/ GE & 35.1\% \\
    ZO-AdaMM %
    & 45.8\% \\
    \texttt{Sparse-RS} + SH %
    &63.9\%\\
    \texttt{Sparse-RS} + SA %
    & \textbf{72.9\% $\pm$ 3.6}\\
    \texttt{Patch-RS} & 70.8\% $\pm$ 1.3\\
\bottomrule
\end{tabular} 
\caption{\label{tab:universal}Success rate of targeted universal $50\times50$ patches.}
\end{table}

\section{Conclusion}
We propose a versatile framework, \rsa{}, which achieves state-of-the-art success rate and query efficiency in multiple sparse threat models: $l_0$-perturbations, %
adversarial patches and adversarial frames (see App. \ref{sec:app_frames}). Moreover, it is effective in the challenging task of crafting universal adversarial patches without relying on surrogate models, unlike the existing methods.
We think that strong black-box adversarial attacks are a very important component
to assess the robustness against such localized and structured attacks, 
which go beyond the standard $l_p$-threat models.

\section*{Acknowledgements}
We thank %
\citet{yang2020patchattack} for quickly releasing their code and answering our questions. F.C., N.S. and M.H. acknowledge support from the German Federal Ministry of
Education and Research (BMBF) through the Tübingen AI Center (FKZ: 01IS18039A), the DFG Cluster of Excellence “Machine Learning – New
Perspectives for Science”, EXC 2064/1, project number 390727645, and by DFG grant 389792660 as part of TRR 248.

\bibliography{Literatur}

\ifpaper
\clearpage
\appendix

\begin{center}
	\Large\textbf{Appendix}
\end{center}

\section*{Organization of the appendix}
The appendix contains several additional results that we omitted from the main part of the paper due to the space constraints, as well as additional implementation details for each method.
The organization of the appendix is as follows:
\begin{itemize}
    \item Sec.~\ref{sec:app_image_classification}: additional results and implementation details of $l_0$-bounded attacks where we also present results of \textit{targeted} attacks on ImageNet.
    \item Sec.~\ref{sec:app_malware_detection}: %
    $l_0$-bounded attacks on malware detection.
    \item Sec.~\ref{sec:app_image_specific_patches}: implementation details for generating image- and location-specific adversarial patches.
    \item Sec.~\ref{sec:app_frames}: %
    image-specific adversarial frames with \rsa{}.
    \item Sec.~\ref{sec:app_univ}: results of universal attacks, i.e. image- and location-independent attacks, which include targeted universal patches and targeted universal frames.
    \item Sec.~\ref{sec:app_theory}: theoretical analysis of the \texttt{$l_0$-RS} algorithm.
    \item Sec.~\ref{sec:app_ablation_st}: ablation studies for \texttt{$l_0$-RS} and \texttt{Patch-RS}.
\end{itemize}

\section{$l_0$-bounded attacks: image classification} \label{sec:app_image_classification}
In this section, we first describe the results of targeted attacks, then we show additional statistics for untargeted attacks, and describe the hyperparameters used for \texttt{$l_0$-RS} and competing methods.

\subsection{Targeted attacks}
Here we test %
$l_0$-bounded attacks in the targeted scenario using the sparsity level $k=150$ (maximum number of perturbed pixels or features) on ImageNet. %
For black-box attacks including \texttt{$l_0$-RS}, we set the query budget to $100,000$, and show the success rate on VGG and ResNet in Table~\ref{tab:l0_targeted} computed on $500$ points. We also give additional budget to white-box attacks compared to the untargeted case (see details in App.~\ref{subsec:app_l0_competitors}).
The target class for each point is randomly sampled from the set of labels which excludes the true label.
We report the success rate in Table~\ref{tab:l0_targeted} instead of robust error since we are considering targeted attacks where the overall goal is to cause a misclassification towards a particular class instead of an arbitrary misclassification.
We can see that targeted attacks are more challenging than untargeted attacks for many methods, particularly for methods like CornerSearch that were designed primarily for untargeted attacks. 
When considering the pixel space, $l_0$-RS outperforms both black- and white-box attacks, while in the feature space it is second only to PDPGD although achieving similar results on the VGG model.

Additionally, we report the query efficiency curves in Fig.~\ref{fig:l0_targeted_curves} for \texttt{$l_0$-RS}, PGD$_0$ and JSMA-CE with gradient estimation. We omit the ADMM attack %
since it has shown the lowest success rate in the untargeted setting. We can see that \texttt{$l_0$-RS} achieves a high success rate (90\%/80\% for VGG/ResNet) even after 20,000 queries. At the same time, the targeted setting is very challenging for other methods that have nearly 0\% success rate even after 100,000 queries.
Finally, we visualize targeted $l_0$ adversarial examples generated by pixel-based \texttt{$l_0$-RS} in Fig.~\ref{fig:l0_visualizations_targeted}.

\begin{table}[t] \centering \small
    \begin{tabular}{c c | c c }
    \setlength\extrarowheight{5pt}
    \textit{attack}& \textit{type}& VGG & ResNet\\
    \toprule\addlinespace[2mm]
    \multicolumn{4}{c}{$l_0$-bound in \textbf{pixel space} $k=150$}\\ \midrule
    JSMA-CE %
    & white-box & 0.4\% & 1.4\% \\
    PGD\textsubscript{$0$} %
    & white-box & 62.8\% & 67.8\% \\ %
    \hdashline
    JSMA-CE %
    w/ GE & black-box &0.0\% & 0.0\%\\
    PGD\textsubscript{$0$} %
    w/ GE & black-box & 0.4\%& 1.4\% \\
    CornerSearch$^*$ %
    & black-box & 3.0\% & 2.0\% \\
    \texttt{$l_0$-RS} & black-box & \textbf{98.2\%} & \textbf{95.6\%} \\ \addlinespace[2mm]
    \multicolumn{4}{c}{$l_0$-bound in \textbf{feature space} $k=150$}\\ \midrule
    FMN & white-box & 73.0\% & 79.8\% \\
    PDPGD & white-box & \textbf{92.8\%} & \textbf{94.2\%} \\
    \hdashline
    \texttt{$l_0$-RS} & black-box & 90.8\% & 80.0\% \\
    \bottomrule
    \end{tabular}%
    \caption{Success rate of targeted $l_0$-attacks on ImageNet. The entry with $^*$ is evaluated on 100 points instead of 500 because of their high computational cost. All black-box attacks use 100k queries except CornerSearch which uses 450k. \texttt{$l_0$-RS} outperforms all black- \emph{and} white-box attacks in the pixel space scenario, and is competitive with white-box attacks in the feature space one.}
    \label{tab:l0_targeted} 
\end{table}
\begin{figure}[t]\centering
    \begin{tabular}{c c}
        \hspace{0.68cm} \textbf{VGG} \hspace{2.5cm} & \textbf{ResNet} \\
    \end{tabular} 
    \includegraphics[width=\columnwidth]{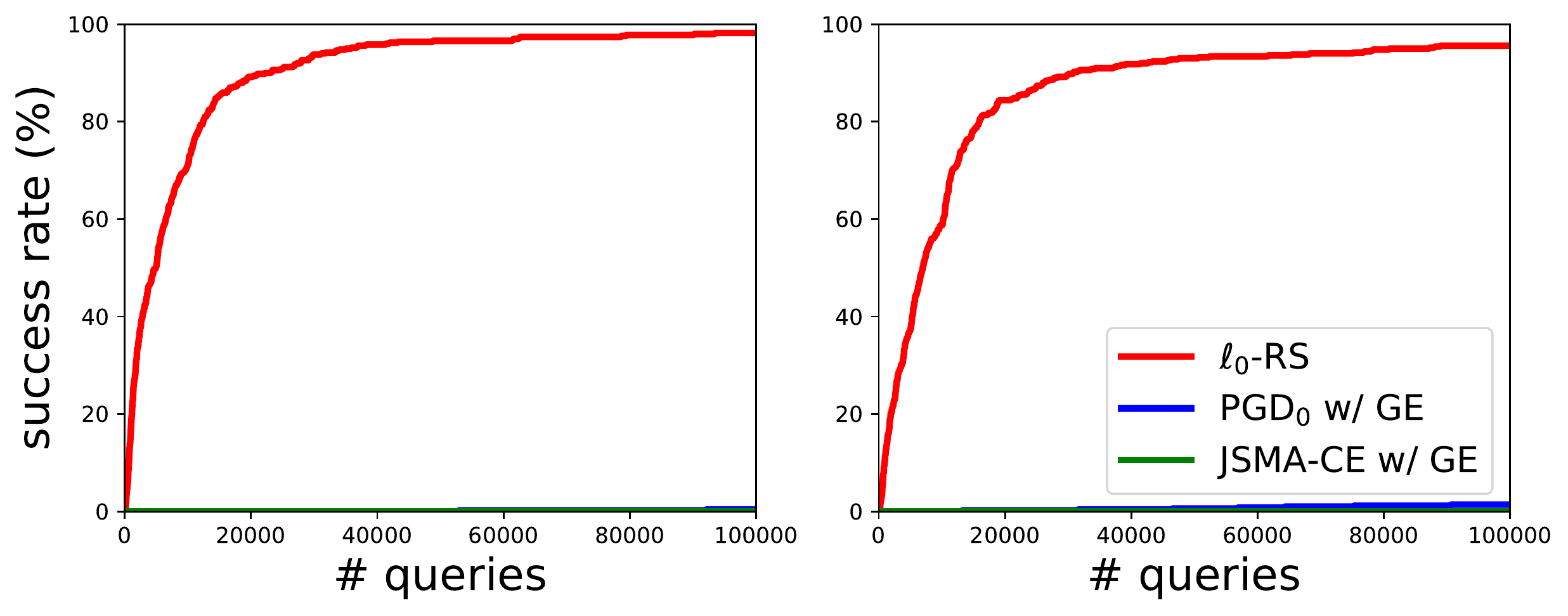}
    \caption{Success rate over queries for targeted black-box attacks in the pixel-based $l_0$-threat model with $k=150$.}
    \label{fig:l0_targeted_curves}
\end{figure}
\begin{figure}[t] \centering 
    \setlength{\tabcolsep}{1pt}
    \includegraphics[clip, trim=0mm 265mm 0mm 0mm, width=0.99\columnwidth]{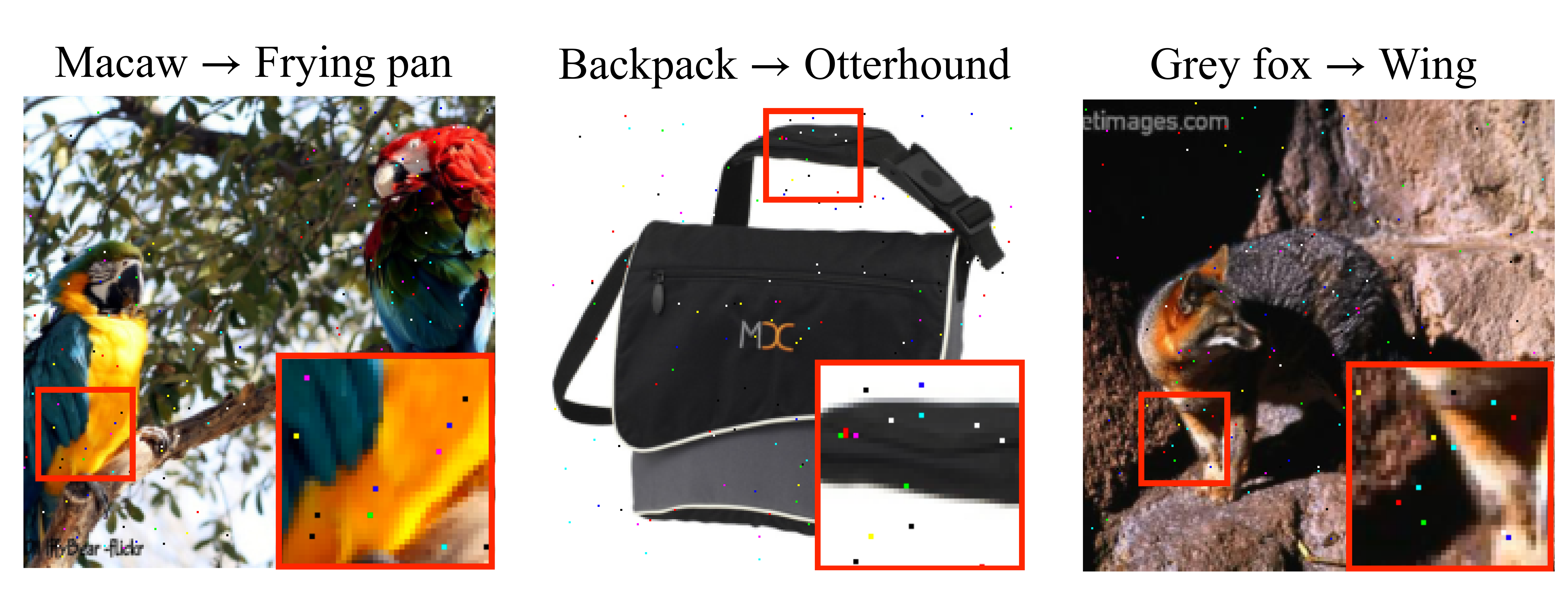}
    \caption{%
    Targeted $l_0$ adversarial examples generated by our pixel-based \texttt{$l_0$-RS} algorithm for $k=150$ pixels.} \label{fig:l0_visualizations_targeted}%
\end{figure}

\subsection{Additional statistics for targeted and untargeted attacks}
In Table~\ref{tab:l0_black_box}, we report the details of success rate and query efficiency of the black-box attacks on ImageNet for sparsity levels $k\in\{50,150\}$ in the pixel space and 10,000/100,000 query limit for untargeted/targeted attacks respectively.
\begin{table*}[t] \centering \small \setlength{\tabcolsep}{5pt}
    \begin{tabular}{c c c | c cc|c cc} & \textit{pixels} & \multirow{2}{*}{\textit{attack}} &  
        \multicolumn{3}{c}{VGG} & \multicolumn{3}{c}{ResNet}\\
        & \textit{modified} & & \textit{succ. rate} & \textit{mean queries} & \textit{med. queries} & \textit{succ. rate} & \textit{mean queries} & \textit{med. queries}\\
    \toprule
    \multirow{8}{*}{\textbf{untargeted}}&\multirow{4}{*}{$k=50$}& ADMM %
     &8.3\% & 9402 & 10000 & 8.8\% & 9232 & 10000\\
    && JSMA-CE %
    w/ GE& 33.5\% & 7157 &10000 & 29.0\% & 7509 & 10000 \\
    && PGD\textsubscript{0} %
    w/ GE & 49.1\% & 5563 &10000 & 38.0\% & 6458 & 10000\\
    &&\texttt{$l_0$-RS} & \textbf{97.6\%} & \textbf{737} & \textbf{88} & \textbf{94.6\%} & \textbf{1176} & \textbf{150}\\
    \cmidrule{2-9}
    
    &\multirow{4}{*}{$k=150$}& ADMM %
    & 16.9\% & 8480 & 10000 &14.2\% & 8699 & 10000\\
    &&JSMA-CE %
    w/ GE& 58.8\% & 4692 &2460 & 49.4\% & 5617 & 10000 \\
    & & PGD\textsubscript{0} %
    w/ GE & 68.1\% & 3447& 42& 59.4\% & 4263 & 240 \\
    &&\texttt{$l_0$-RS} & \textbf{100\%} & \textbf{171} & \textbf{25} & \textbf{100\%} & \textbf{359} & \textbf{49}\\
    \midrule
    \multirow{3}{*}{\textbf{targeted}}&\multirow{3}{*}{$k=150$}& JSMA-CE %
    w/ GE & 0.0\% & 100000 & 100000 & 0.0\% & 100000 & 100000\\
    &&PGD\textsubscript{0} %
    w/ GE & 0.4\% & 99890 & 100000& 1.4\% & 99357 & 100000\\
    &&\texttt{$l_0$-RS} & \textbf{98.2\%} & \textbf{9895} & \textbf{4914} & \textbf{95.6\%} & \textbf{14470} & \textbf{6960}\\
    \bottomrule
    \end{tabular}%
    \caption{We report the success rate and statistics about query consumption of black-box $l_0$-bounded attacks in the pixel space on ImageNet. We consider sparsity levels $k\in\{50,150\}$ and compute the average and median number of queries needed to generate adversarial examples on all correctly classified points.}\label{tab:l0_black_box}
\end{table*}

\subsection{\texttt{$l_0$-RS} algorithm}
As mentioned in Sec.~\ref{sec:l0_bounded}, at iteration $i$ the new set $M'$ is formed modifying $\iter{\alpha}{i}\cdot k$ (rounded to the closest positive integer) elements of $M$ containing the currently modified dimensions (see step 6 in Alg.~\ref{alg:rs_general}). Inspired by the step-size reduction in gradient-based optimization methods, we progressively reduce $\iter{\alpha}{i}$.
Assuming $N=10,000$, the schedule of $\iter{\alpha}{i}$ is piecewise constant where the constant segments start at iterations $j\in\{0, 50, 200, 500, 1000, 2000, 4000, 6000, 8000\}$ with values $\nicefrac{\alpha_\text{init}}{\beta_j}$, $\beta_j \in\{2, 4, 5, 6, 8, 10, 12, 15, 20\}$. For a different maximum number of queries $N$, the schedule is linearly rescaled accordingly. In practice, we use $\alpha_\text{init}=0.3$ and $\alpha_\text{init}=0.1$ for the untargeted and targeted scenario respectively on ImageNet (for both pixel and feature space), $\alpha_\text{init}=0.3$ and $\alpha_\text{init}=0.1$ when considering the pixel and feature space respectively on CIFAR-10.

\subsection{Competitors}
\label{subsec:app_l0_competitors}

\textbf{SparseFool.} We use SparseFool \cite{ModEtAl19} %
in the original implementation and optimize the hyperparameter which controls the sparsity of the solutions (called $\lambda$, setting finally $\lambda=1$) to achieve the best results. %
we use 60 and 100 steps for ImageNet and CIFAR-10 %
(default value is 20), %
after checking that higher values do not lead to improved performance but significantly increase the computational cost. Note that SparseFool has been introduced for the $l_1$-threat model but can generate sparse perturbations which are comparable to $l_0$-attacks.

\textbf{PGD\textsubscript{$0$}.}
For PGD\textsubscript{$0$} \cite{croce2019sparse} we use 2,000 iterations (5,000 for targeted attacks) and step size $0.05\cdot d$ ($0.025\cdot d$ for targeted attacks) for ImageNet ($0.5\cdot d$ for CIFAR-10) where $d=224\times 224 \times 3$ (resp. $d=32\times 32 \times 3$) is the input dimension. PGD\textsubscript{$0$} with gradient estimation uses the same gradient step of PGD\textsubscript{$0$} but with step size $5\cdot d$ ($1\cdot d$ for targeted attacks). To estimate the gradient we use finite differences, similarly to \cite{ilyas2018black}, as shown %
in Alg.~\ref{alg:ge}, with the current iterate as input $x$, %
$\eta=0.01/\sqrt{d}$ (in line with what suggested in \cite{ilyas2019prior} for this algorithm), $m=1$ and a zero vector as $p$. In case of targeted attacks we instead use $m=50$ (the budget of queries is also 10 times larger) and the current estimated gradient as $p$. We optimized the aforementioned parameters to achieve the best results and tested that sampling more points to better approximate the gradient at each iteration leads to similar success rate with worse query consumption.

\begin{algorithm}[t]
	\caption{\texttt{GradEst}: Gradient estimation via finite differences}
	\label{alg:ge}
	\SetKwData{Left}{left}
  \SetKwData{Up}{up}
  \SetKwFunction{FindCompress}{FindCompress}
  \SetKwInOut{Input}{input}
  \SetKwInOut{Output}{output}
	\Input{loss function $L$, input point $x$, number of iterations $m$, step size $\eta$, prior estimation $p$}
	\Output{estimated gradient $g$}
$g \gets p$ \\
\For{$i=1, \ldots, m$}{
    $s\gets \mathcal{N}(0, I)$\\
    $l_1 \gets L(x + \eta \cdot s)$\\
    $l_2 \gets L(x - \eta \cdot s)$\\
    $g \gets g + (l_1 - l_2)/(2\eta) \cdot s$
}
\end{algorithm}

\textbf{CornerSearch.} For CornerSearch on ImageNet, we used the following hyperparameters: \texttt{n\_max}~$=500$, \texttt{n\_iter}~$=200$, \texttt{n\_classes}~$=1,000$ (i.e. all classes, which is the default option) for untargeted attacks and \texttt{n\_max}~$=1,500$, \texttt{n\_iter}~$=50,000$ for targeted attacks. We note that CornerSearch has a limited scalability to ImageNet as it requires $8\cdot224\cdot224=401,408$ queries only for the initial phase of the attack to obtain the orderings of pixels, and then for the second phase \texttt{n\_iter}~$\cdot$~\texttt{n\_classes}~$ = 200,000$ queries for untargeted attacks and \texttt{n\_iter}~$=50,000$ queries for targeted attacks. On CIFAR-10 we set \texttt{n\_max}~$=450$, \texttt{n\_iter}~$=1000$, \texttt{n\_classes}~$=10$, which amounts to more than $17,000$ queries of budget. Thus, CornerSearch requires significantly more queries compared to \texttt{$l_0$-RS} while achieving a lower success rate. %
Finally, we note that CornerSearch was designed to minimize the number of modified pixels, while we use it here for \textit{$l_0$-bounded} attacks.

\textbf{SAPF.} For SAPF~\cite{fan2020sparse}, we use the hyperparameters suggested in their paper and we use the second-largest logit as the target class for untargeted attacks. With these settings, the algorithm is still very expensive computationally, so we could evaluate it only on 100 points. %
Note that \cite{fan2020sparse} report that SAPF achieves the average $l_0$-norm of at least 30,000 which is orders of magnitudes more than what we consider here. But we evaluate it for completeness as their goal is to obtain a sparse adversarial attack measured in terms of $l_0$.

\textbf{ADMM.} For ADMM \cite{zhao2019design} we perform a grid search over the parameters to optimize the success rate and set \texttt{ro=10}, \texttt{gama=4}, \texttt{binary\_search\_steps=500} and \texttt{Query\_iterations=20} (other parameters are used with the default values). Note that \cite{zhao2019design} report results for the $l_0$-norm only on MNIST. We apply it to both pixel and feature space scenarios by minimizing the corresponding metric.

\textbf{JSMA-CE.} As mentioned in Sec.~\ref{sec:l0_imagenet} we adapt the method introduce in \cite{papernot2016limitations} to scale it to attack models on ImageNet. We build the saliency map as the gradient of the cross-entropy loss instead of using the gradient of the logits because ImageNet has 1000 classes and it would be very expensive to get gradients of each logit separately (particularly in the black-box settings where one has to first estimate them). Moreover, at each iteration we perturb the pixel which has the gradient with the largest $l_1$-norm, setting each channel to either 0 or 1 according to the sign of the gradient at it. We perform these iterations until the total sparsity budget is reached. For the black-box version, we modify the algorithm as follows: the gradient is estimated via finite difference approximation as in Alg.~\ref{alg:ge}, the original image $\xorig{}$ as input throughout the whole procedure (this means that the prior $p$ is always the current estimation of the gradient). We found that this gives stronger results compared to an estimation of the gradient at a new iterate. We use the gradient estimation step size $\eta=0.01$. Then every 10 queries we perturb the $k$ pixels with the largest gradient as described above for the white-box attack and check whether this new image is misclassified (we do not count this extra query for the budget of 10k). This is in line with the version of the attack we use for the malware detection task %
described in detail in Sec.~\ref{sec:app_malware_detection}.

\textbf{EAD.} We use EAD \cite{CheEtAl2018} from Foolbox \cite{foolbox} with $l_1$ decision rule, regularization parameter $\beta=0.1$ optimized to achieve highest success rate in the $l_0$-threat model and 1000 iterations (other parameters as default). Note that EAD is a strong white-box attack for $l_1$ and we include it as additional baseline since it can generate sparse attacks.

\textbf{VFGA.} We use VFGA \cite{cesaire2020stochastic} as implemented in Adversarial Library \cite{rony2020adversarial} with the default parameters.

\textbf{FMN.} For the Fast Minimum-Norm attack \cite{pintor2021fast} we use the implementation of Adversarial Library \cite{rony2020adversarial} with $1,000$ iterations for the untargeted case (as in the original paper), $10,000$ for the targeted one. We set the hyperparameter $\alpha_0$ to $10$ which was optimized with a grid search. %
We also note that \cite{pintor2021fast} do not report results for the $l_0$-version of their attack on ImageNet.

\textbf{PDPGD.} We run PDPGD \cite{matyasko2021pdpgd} with the implementation of Adversarial Library \cite{rony2020adversarial} with $1,000$ and $10,000$ iterations for untargeted and targeted scenarios respectively, $l_{2/3}$-proximal operator and other parameters with default values. While \cite{matyasko2021pdpgd} introduce also a version of their attack for the pixel space, it is not available in Adversarial Library, then we only consider the method for the comparison in the feature space.

\subsection{CIFAR-10 models} \label{app:cifar10_models}
We here report the details of the classifiers used for the experiments on CIFAR-10. We consider the $l_2$-robust PreActResNet-18 from \cite{rebuffi2021fixing}, available in RobustBench \cite{croce2020robustbench}, and the $l_1$-robust one from \cite{croce2021mind}. In fact, training for robustness in such $l_p$-threat models is expected to yield some level of robustness also against $l_0$-attacks, as confirmed in the experiments.

\section{$l_0$-bounded attacks: malware detection}
\label{sec:app_malware_detection}
In the following we apply \texttt{$l_0$-RS} in a different domain, i.e. malware detection, showing its versatility.
We consider the Drebin dataset \cite{arp2014drebin} which consists of $129,013$
Android applications, among which $123,453$ are benign and $5,560$
malicious, with $d=545,333$ features divided into 8 families. Data points are represented by a binary vector $x \in \{0,1\}^d$ indicating whether each feature is present or not in $x$ (unlike image classification the input space is in this case discrete).
As \cite{grosse2016adversarial}, we restrict the attacks to only adding features from the first 4 families, that is modifications to the manifest of the applications, to preserve the functionality of the samples (no feature present in the clean data is removed), which leaves a maximum of 233,727 alterable dimensions.

\textbf{Model.}
We trained the classifier, which has 1 fully-connected hidden layer with 200 units and uses ReLU as activation function, with 20 epochs of SGD minimizing the cross-entropy loss, with a learning rate of $0.1$ reduced by a factor of 10 after 10 and 15 epochs.
We use batches of size 2000, consisting of $50\%$ of benign and $50\%$ of malicious examples.
For training, we merged the training and validation sets of one of the splits provided by \cite{arp2014drebin}.
It achieves a test accuracy of $98.85\%$, with a false positive rate of $1.01\%$, and a false negative rate of $4.29\%$.

\textbf{\texttt{$l_0$-RS} details for malware detection tasks.} We apply \texttt{$l_0$-RS} as described at the beginning of Sec.~\ref{sec:l0_bounded} for images as input by setting $h=d$ and $w=c=1$ such that $x \in \{0, 1\}^{d\times 1 \times 1}$. Only adding features means that all values in $\Delta$ equal 1, thus $\Delta' \equiv \Delta$ at every iteration (step~7 in Alg.~\ref{alg:rs_general}) and only the set $M$ is updated. For our attack we use $\alpha_\text{init}=1.6$ and the same schedule of $\iter{\alpha}{i}$ of \texttt{$l_0$-RS} on image classification tasks (see Sec.~\ref{sec:app_image_classification}).

\textbf{Competitors.}
\cite{grosse2016adversarial} successfully fooled similar models with a variant of the white-box JSMA \cite{papernot2016limitations}, and \cite{podschwadt2019effectiveness} confirms that it is the most effective technique on Drebin, compared to the attacks of \cite{stokes2017attack, aldujaili2018adversarial, hu2017generating} including adaptation of FGSM \cite{SzeEtAl2014} and PGD \cite{KurGooBen2016a,MadEtAl2018}. We use JSMA %
and PGD\textsubscript{0} %
in a black-box version with gradient estimation (details below).
\cite{liu2019adversarial} propose a black-box genetic algorithm with prior knowledge of the importance of the features for misclassification (via a pretrained random forest) which is not comparable to our threat model.

\begin{figure}[t]\centering \includegraphics[width=0.99\columnwidth]{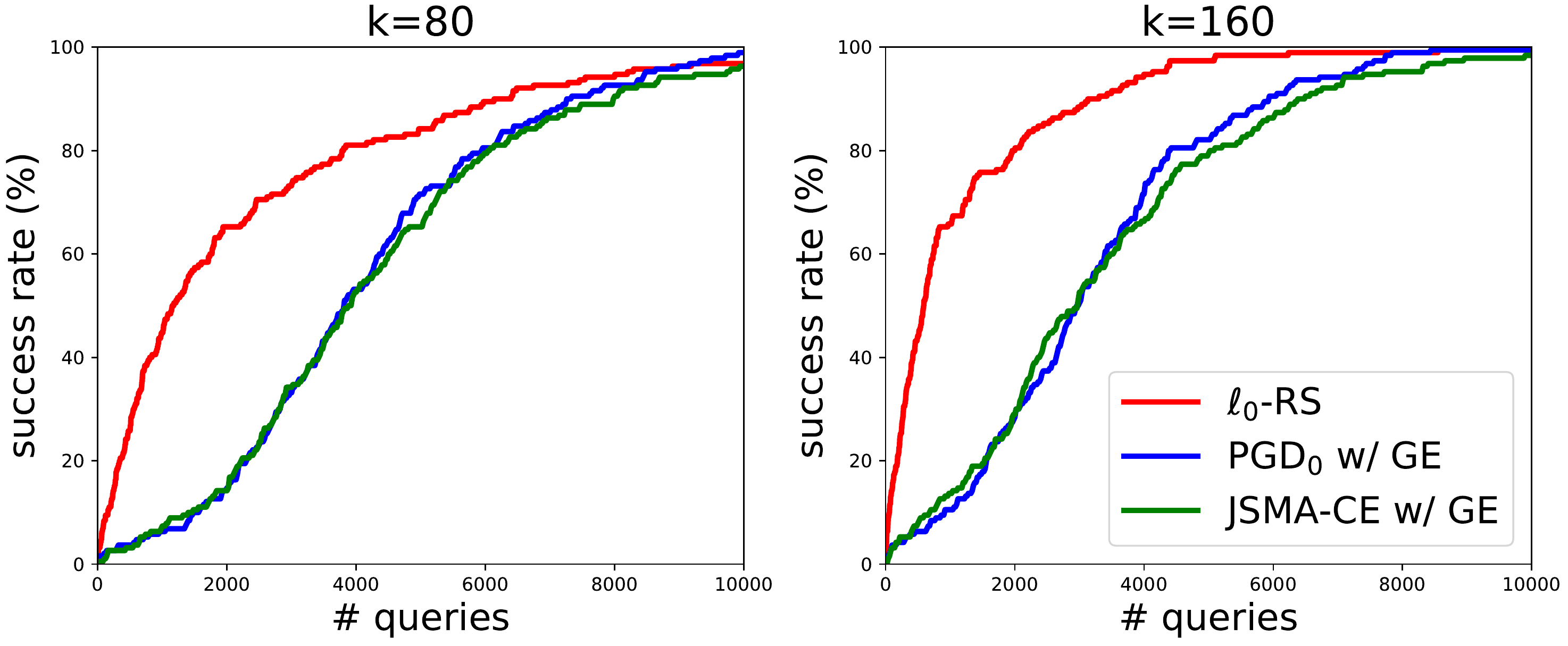}
\caption{%
Success rate vs number of queries %
in fooling a malware detector for sparsity levels $k\in\{80, 160\}$. \texttt{$l_0$-RS} outperforms the competitors, with a large gap in the low query regime.} \label{fig:drebin}%
\end{figure}

\textbf{Results.} We compare the attacks allowed to modify $k \in \{80,160\}$ features and with a maximum of $10000$ queries. Fig.~\ref{fig:drebin} shows the progression of success rate (computed on the initially correctly classified apps) of the attacks over number of queries used. At both sparsity levels, our \texttt{$l_0$-RS} attack (in red) achieves very high success rate and outperforms JSMA (green) and PGD\textsubscript{0} (blue) in the low query regime, where the approximation of the gradient is likely not sufficiently precise to identify the most relevant feature for the attacker to perturb.

\begin{algorithm}[t]
	\caption{JSMA-CE with gradient estimation (malware detection)}
	\label{alg:jsma_ge}
	\SetKwData{Left}{left}
  \SetKwData{Up}{up}
  \SetKwFunction{FindCompress}{FindCompress}
  \SetKwInOut{Input}{input}
  \SetKwInOut{Output}{output}
	\Input{cross-entropy loss $L$, input point $\xorig{}$, number of iterations $N$, pixel budget $k$, number of iterations for gradient estimation $m$, gradient estimation step size $\eta$}
	\Output{$z$}
$g \gets \mathbf{0}$, \quad $q \gets 2\cdot m$\\
\While{$q\leq N$}{
$g \gets \texttt{GradEst}(L, \xorig, m, \eta, g)$\\
$z\gets \xorig{}$\\
$A \gets $ indices of $k$ largest positive components of $g$ (if there are)\\
$z_A \gets 1$\\
$q\gets q + 2 \cdot m$\\
\lIf{$z$ is adversarial}{{\rm \bf return}}
}%
\end{algorithm}

\subsection{Competitors}
\textbf{JSMA-CE with gradient estimation.} The idea of the white-box attack of \cite{grosse2016adversarial} is, given a sparsity level of $k$, to perturb iteratively the feature of the iterate $\iter{x}{i}$ which corresponds to the largest value (if positive) of $\nabla_x L_\text{CE}(f(\iter{x}{i}),y)$, with $f$ the classifier, $y$ the correct label and $L_\text{CE}$ the cross-entropy loss, until either the maximum number of modifications are made or misclassification is achieved. With only approximate gradients this approach is not particularly effective.
However, since $k \ll d$
and only additions can be made, we aim at estimating the gradient of the cross-entropy loss at $\xorig{}$ 
and then set to 1 the $k$ elements (originally 0) of $\xorig{}$ with the largest component in the approximated gradient.
With the goal of query efficiency, every $m$ iterations of gradient estimation through finite differences we check if an adversarial example has been found (we do not count these queries in the total
for the query limit).
Alg.~\ref{alg:jsma_ge} shows the procedure, and %
we set $m=5$ and $\eta=1$ (we tested other values which achieved worse or similar performance).

\textbf{PGD\textsubscript{0} with gradient estimation.} We use PGD\textsubscript{0} %
with the gradient estimation technique presented in Alg.~\ref{alg:ge} with the original point as input $x$, $m=10$, $\eta=100$ and the current estimate of the gradient as prior $p$ (unlike on image classification tasks where the gradient is estimated at the current iterate and no prior used). Moreover, we use step size $4 \cdot \sqrt{d}$ and modify the projection step the binary input case and so that features are only added and not removed (see above).

\section{Image- and location-specific adversarial patches} \label{sec:app_image_specific_patches}
We here report the details of the attacks for image- and location-specific adversarial patches.

\subsection{\textbf{\texttt{Patch-RS}} algorithm}
As mentioned in Sec.~\ref{sec:image_specific_patches}, in this scenario we optimize via random search the attacks for each image independently. In \texttt{Patch-RS} we alternate iterations where a candidate update of the patch is sampled with others where a new location is sampled.

\textbf{Location updates.} We have a location update every $m$ iterations, %
with $m=5$ ($1:4$ scheme Sec.~\ref{sec:app_abl_patches}, four patch updates then one location update) for untargeted attacks and $m=10$ for targeted ones: the latter have more queries available (50k vs 10k) and wider patches ($40\times 40$ vs $20\times 20$), therefore it is natural to dedicate a larger fraction of iterations to optimizing the content rather than the position (which has also a smaller feasible set). We present in Sec.~\ref{sec:app_abl_patches} an ablation study the effect of different frequencies of location updates in the untargeted scenario.
The position of the patch is updated with an uniformly sampled shift in $[-\iter{h}{i}, \iter{h}{i}]$ for each direction (plus clipping to the image size if necessary), where $\iter{h}{i}$ is linearly decreased from $\iter{h}{0}=0.75\cdot s_\text{image}$ to $\iter{h}{N}=0$ ($s_\text{image}$ indicates the side of the squared images). In this way, initially, the patch can be easily moved on the image, while towards the final iterations it is kept (almost) fixed and only its values are optimized.

\textbf{Patch updates.} The patch is initialized by superposing at random positions on a black image 1000 squares of random size and color in $\{0, 1\}^3$. Then, we update the patch following the scheme of Square Attack \cite{ACFH2019square}, that is sampling random squares with color one of the corners of the color cube and accept the candidate if it improves the target loss. The size of the squares is decided by a piecewise constant schedule, which we inherit from \cite{ACFH2019square}. Specifically, at iteration $i$ the square-shaped updates of a patch with size $s\times s$ have side length $s^{(i)}=\sqrt{\alpha^{(i)}} \cdot s$, where the value of $\iter{\alpha}{i}$ starts with $\iter{\alpha}{0}=\alpha_\text{init}$ and then is halved at iteration $j\in\{10, 50, 200, 500, 1000, 2000, 4000, 6000, 8000\}$ if the query limit is $N=10,000$, otherwise the values of $j$ are linearly rescaled according to the new $N$. Hence, $\alpha_\text{init}$ is the only tunable parameters of \texttt{Patch-RS}, and we set $\alpha_\text{init}=0.4$ for untargeted attacks and $\alpha_\text{init}=0.1$ for targeted ones.

In contrast to Square Attack, in the second half of the iterations dedicated to updates of size $1\times 1$, \texttt{Patch-RS} performs a refinement of the patch applying only single-channel updates. We show in Sec.~\ref{sec:app_abl_patches} that both the initialization tailored for the specific threat model and the single-channel updates allow our algorithm to achieve the best success rate and query efficiency.

\subsection{Competitors}
\textbf{\texttt{Sparse-RS} + SH and \texttt{Sparse-RS} + SA.} When integrating SignHunter \cite{AlDujaili2019ThereAN} and Square Attack \cite{ACFH2019square} in our framework for adversarial patches, the updates of the locations are performed as described above, while the patches are modified with the $l_\infty$ attacks with constraints given by the input domain $[0,1]^d$ (in practice one can fix a sufficiently large $l_\infty$ threshold $\epsilon_\infty=1$). SH does not have free parameters, while we tune the only parameter $p$ of SA (which has the same role as $\alpha_\text{init}$ of \texttt{Patch-RS}) and use $p=0.4$ and $p=0.2$ for the untargeted and targeted case respectively.

\textbf{LOAP.} %
Location-Optimized Adversarial Patches (LOAP) \cite{rao2020adversarial} is a white-box PGD-based attack which iteratively first updates the patch with a step in the direction of the sign of the gradient to maximize the cross-entropy function at the current iterate, then it tests if shifting of \texttt{stride} pixels the patch in one of the four possible directions improves the loss and, if so, the location is updated otherwise kept (\cite{rao2020adversarial} have also a version where only one random shift is tested). %
We use LOAP as implemented in \url{https://github.com/sukrutrao/Adversarial-Patch-Training} with 1,000 iterations, learning rate 0.05 and the \texttt{full} optimization of the location with \texttt{stride=5}.
To adapt LOAP to the black-box setup, we use the gradient estimation via finite differences shown in Alg.~\ref{alg:ge} restricted to the patch to optimize (this means that the noise sampled in step 3 is applied only on the components of the image where the patch is). In particular we use the current iterate as input $x$, $m=5$ iterations, $\eta=10/(s\cdot\sqrt{c})$ if the patch has dimension $s\times s \times c$ and the gradient estimation at the previous iteration as prior $p$. Moreover, we perform the update of the location by sampling only 1 out of the 4 possible directions (as more queries are necessary for better gradient estimation) with \texttt{stride=5}, and learning rate 0.02 for the gradient steps (the sign of the gradient is used as a direction). Note that we optimized these hyperparameters to achieve the best final success rate. Moreover, each iteration costs 11 queries (10 for Alg.~\ref{alg:ge} and 1 for location updates) and we iterate the procedure until the budget of queries is exhausted.

\textbf{TPA.} The second method we compare to is TPA \cite{yang2020patchattack}, which is based on reinforcement learning and exploits a dictionary of patches. Note that in \cite{yang2020patchattack} TPA was used primarily putting multiple patches on the same image to achieve misclassification, while our threat model does not include this option. We obtained the optimal values for the hyperparameters for untargeted attacks (\texttt{rl\_batch=400} and \texttt{steps=25}) via personal communication with the authors and set those for the targeted scenario (\texttt{rl\_batch=1000} and \texttt{steps=50}) similarly to what reported in the original code at \url{https://github.com/Chenglin-Yang/PatchAttack}, doubling the value of \texttt{rl\_batch} to match the budget of queries we allow. TPA has a mechanism of early stopping, which means that it might happen that not the whole budget of queries is exploited even for unsuccessful points. Finally, \cite{yang2020patchattack} show that TPA significantly outperforms the methods of \cite{fawzi2016measuring}, which also generates image-specific patches, although optimizing the shape and the location but not the values of the perturbation. Thus we do not compare directly to \cite{fawzi2016measuring} in our experiments.

\section{Image-specific adversarial frames} \label{sec:app_frames}
Adversarial frames introduced in \cite{zajac2019adversarial} are another sparse threat model where the attacker is allowed to perturb only pixels along the borders of the image. In this way, the total number of pixels available for the attack is small (3\%-5\% in our experiments), and in particular the semantic content of the image is not altered by covering features of the correct class. Thus, this threat model shows how even peripheral changes can influence the classification. %

\begin{table*}[t]
    \centering \small
    \setlength{\tabcolsep}{4.5pt}
    \begin{tabular}{p{1.6cm} l c  |  c c c  |  c c c } %
    &&\multirow{2}{*}{\textit{attack}} &  
    \multicolumn{3}{c}{VGG} & \multicolumn{3}{c}{ResNet}\\
    & & &\textit{success rate} & \textit{mean queries} & \textit{med. queries} & \textit{success rate} & \textit{mean queries} & \textit{med. queries}\\
    \toprule
    \multirow{5}{*}{\begin{tabular}{c} \textbf{untarget.}\\ 
    (2 px wide)
    \end{tabular}} & \multirow{4}{*}{
    \rotatebox[origin=c]{90}{black-box}}
    
    & AF %
    w/ GE& 57.6\% $\pm$ 0.4 & 5599 $\pm$ 23 & 6747 $\pm$ 127 & 70.6\% $\pm$ 0.7 & 4568 $\pm$ 13 & 3003 $\pm$ 116\\
    & &\texttt{Sparse-RS} + SH %
     & 79.9\% & 3169 & 882 & 88.7\% & 2810 & 964 \\
    & &\texttt{Sparse-RS} + SA %
     & 77.8\% $\pm$ 2.0 & 3391 $\pm$ 215 & 1292 $\pm$ 220 & 80.7\% $\pm$ 1.4 & 3073 $\pm$ 111 & 1039 $\pm$ 93\\
    &&\texttt{Frame-RS}  & \textbf{90.2\% $\pm$ 0.3} & \textbf{2366 $\pm$ 26} & \textbf{763 $\pm$ 32}  & \textbf{94.0\% $\pm$ 0.3} & \textbf{1992 $\pm$ 34} & \textbf{588 $\pm$ 10} \\ 
    \cdashline{2-9}
    & & White-box AF %
    & 100\% & - & -& 100\% & - & - \\
    \midrule
    \multirow{5}{*}{\begin{tabular}{c} \textbf{targeted}\\ 
    (3 px wide)
    \end{tabular}} & \multirow{4}{*}{
    \rotatebox[origin=c]{90}{black-box}}& AF %
    w/ GE& 16.1\% $\pm$ 1.5 & 45536 $\pm$ 328 & 50000 $\pm$ 0 & 40.6\% $\pm$ 1.4 & 39753 $\pm$ 459 & 50000 $\pm$ 0\\
    &&\texttt{Sparse-RS} + SH %
    & 52.0\% & 32223 & 41699 & 73.2\% & 25929 & 20998 \\
    & &\texttt{Sparse-RS} + SA %
     & 32.9\% $\pm$ 1.9 & 38463 $\pm$ 870 & 50000 $\pm$ 0 & 47.5\% $\pm$2.8 & 33321 $\pm$ 980 & 48439 $\pm$ 3500\\
    & &\texttt{Frame-RS}  & \textbf{65.7\% $\pm$ 0.4} & \textbf{28182 $\pm$ 99} & \textbf{27590 $\pm$ 580}  & \textbf{88.0\% $\pm$ 0.9} & \textbf{19828 $\pm$ 130} & \textbf{15325 $\pm$370}\\ 
    \cdashline{2-9}
    & &White-box AF %
    & 100\% & - & -& 100\% &- &-\\
    \bottomrule
    \end{tabular}
    \caption{Success rate and query statistics of image-specific frames %
    computed for 5 seeds. 
    Black-box attacks are given 10k/50k queries for untargeted/targeted case.
    SH %
    is a deterministic method. \texttt{Frame-RS} achieves the best success rate and efficiency in all settings. %
    }
    \label{tab:image_spec_frame}
\end{table*}

\begin{figure*}[t] \centering 
    \setlength{\tabcolsep}{1pt}
    \begin{tabular}{ccc|ccc}
        \multicolumn{3}{c|}{\textbf{Untargeted frames}} & \multicolumn{3}{c}{\textbf{Targeted frames}} \\
        \scriptsize Lens cap $\rightarrow$ Chain & 
        \scriptsize Rugby ball $\rightarrow$ Bottle cap & 
        \scriptsize Blue heron $\rightarrow$ Pole & 
        \scriptsize Collie $\rightarrow$ Canoe & 
        \scriptsize Lakeside $\rightarrow$ Tiger & 
        \scriptsize Bullterrier $\rightarrow$ Website \\
        \includegraphics[clip, trim=35mm 13mm 32mm 14mm, width=0.34\columnwidth]{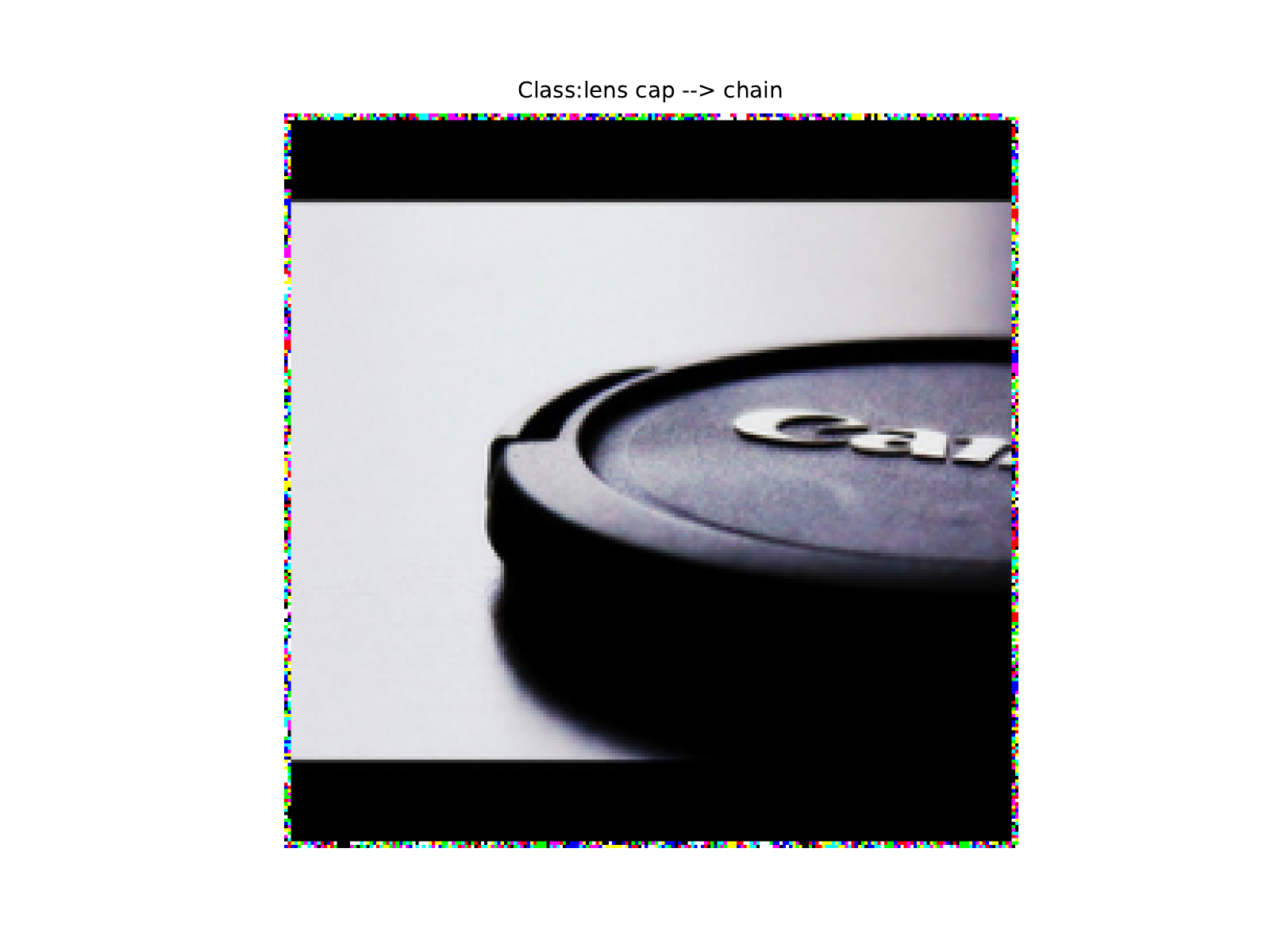} &
        \includegraphics[clip, trim=35mm 13mm 32mm 14mm, width=0.34\columnwidth]{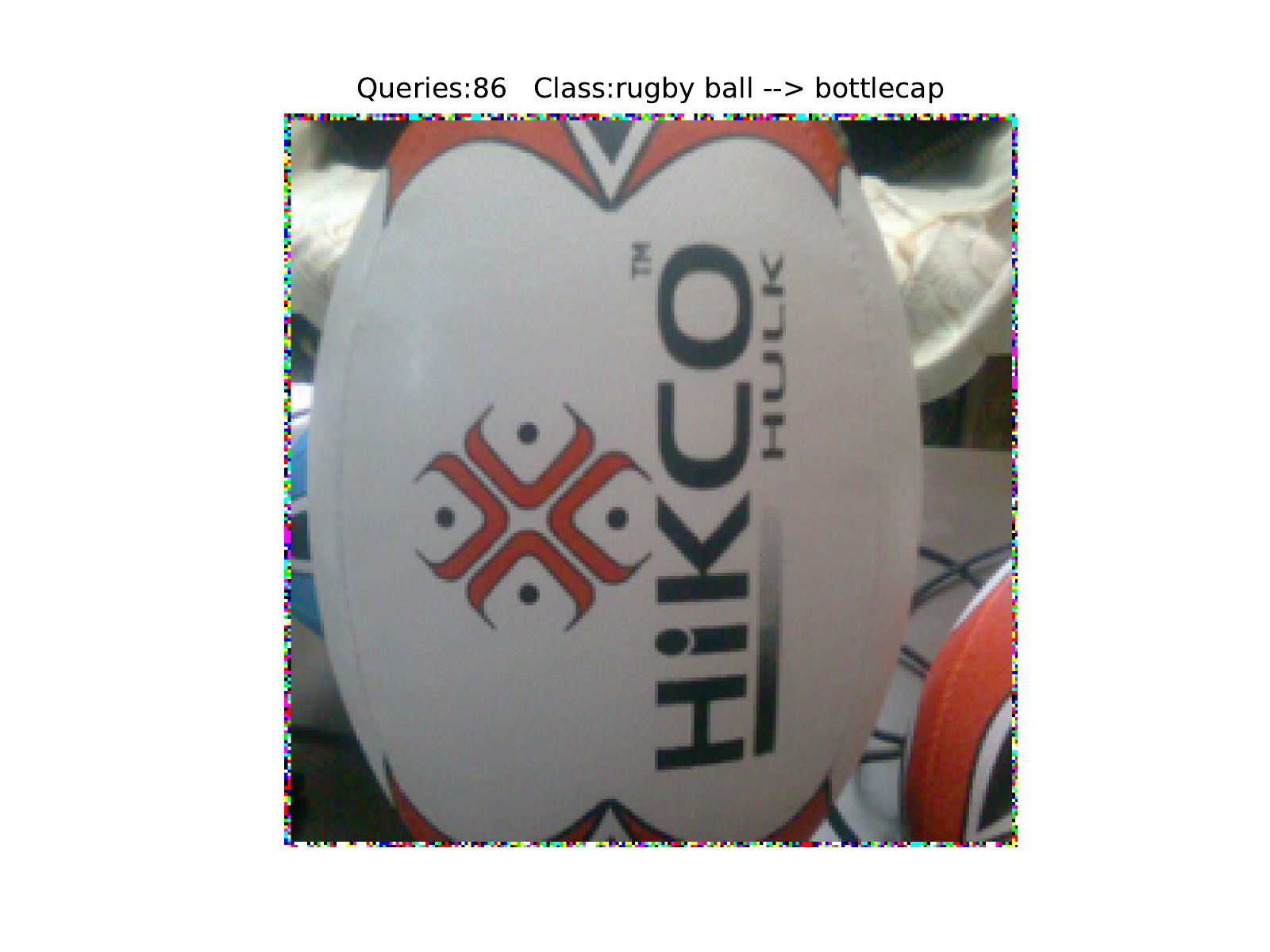} &
        \includegraphics[clip, trim=35mm 13mm 32mm 14mm, width=0.34\columnwidth]{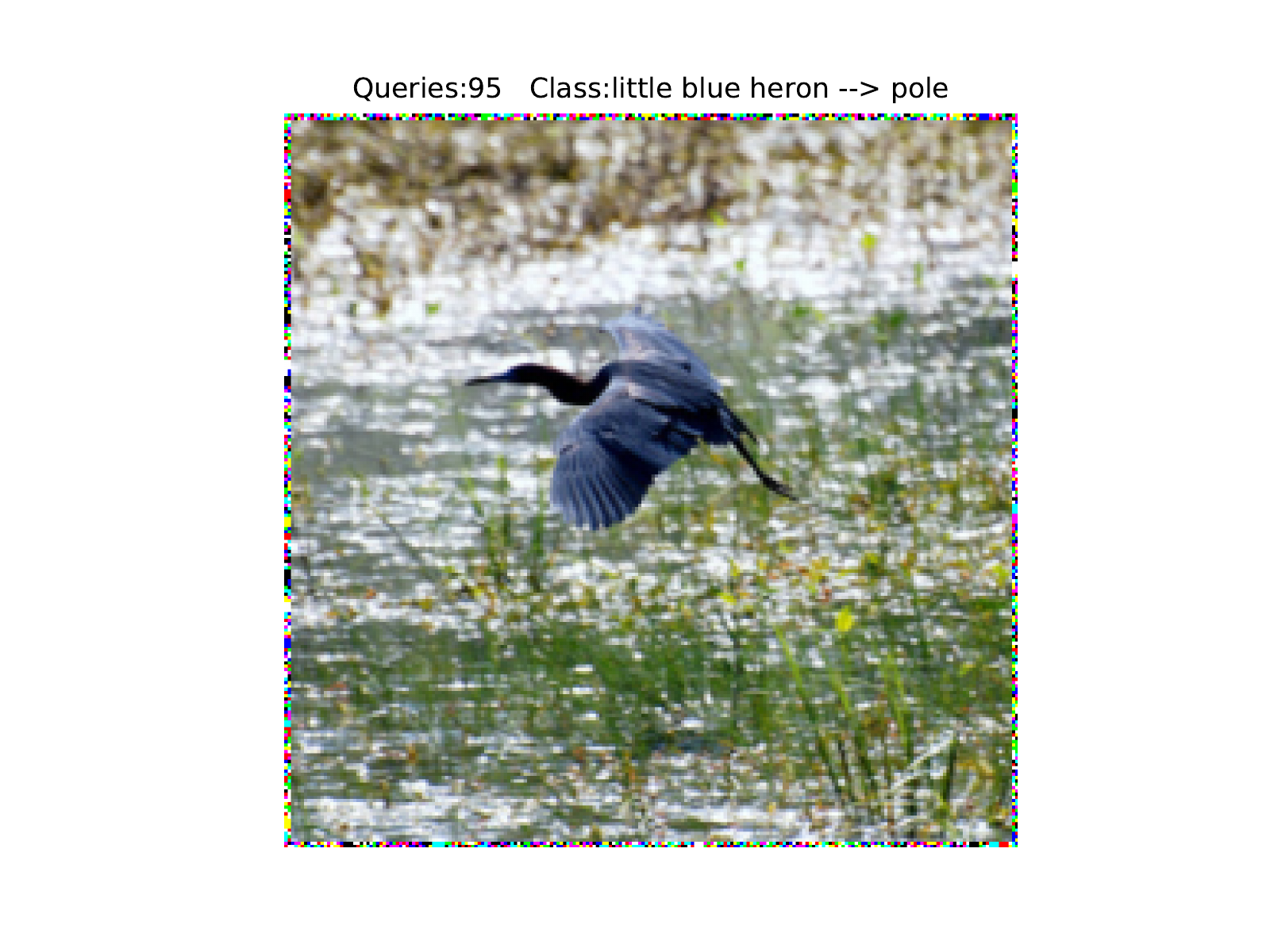} &
        \includegraphics[clip, trim=35mm 13mm 32mm 14mm, width=0.34\columnwidth]{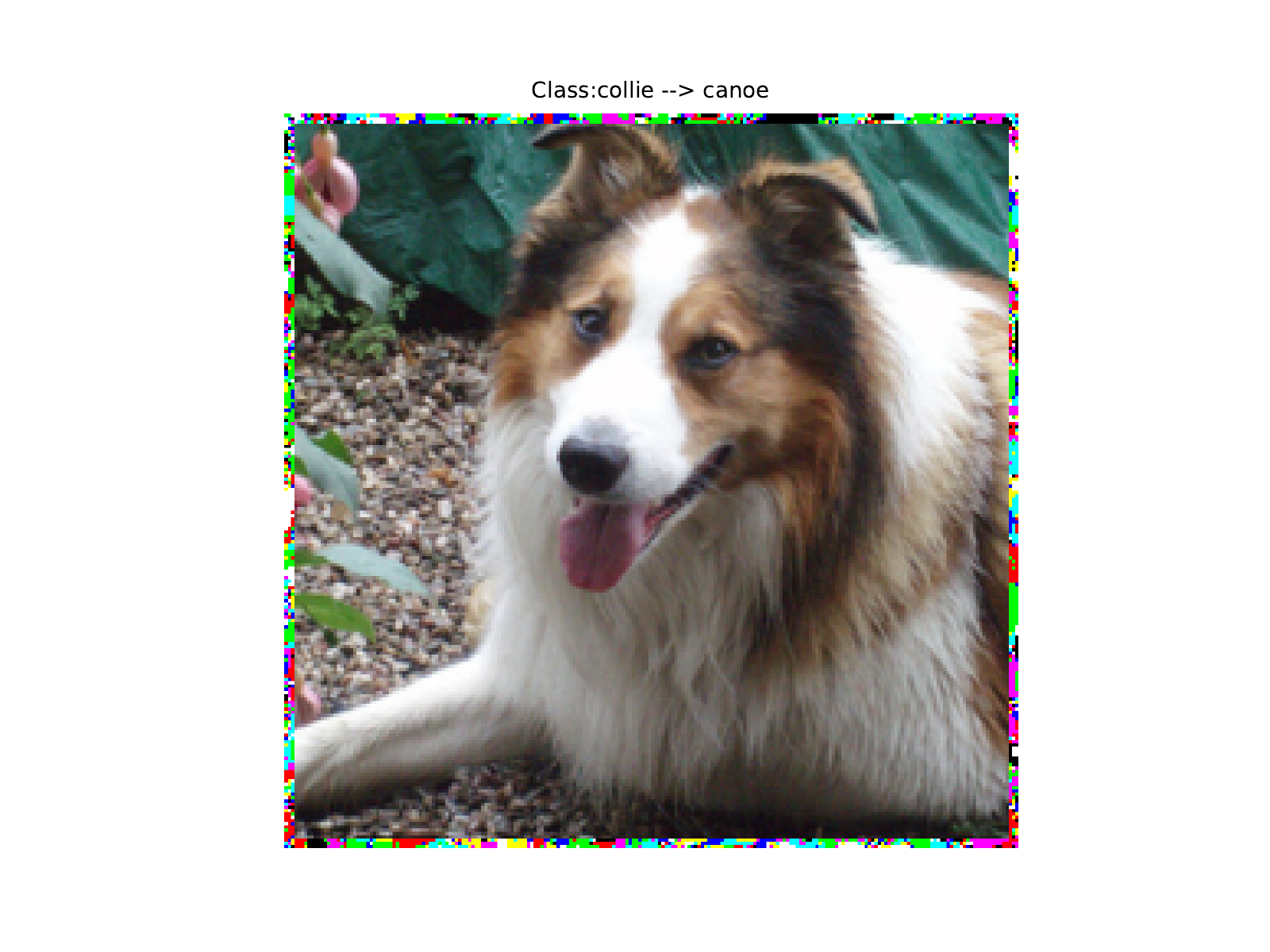} &
        \includegraphics[clip, trim=35mm 13mm 32mm 14mm, width=0.34\columnwidth]{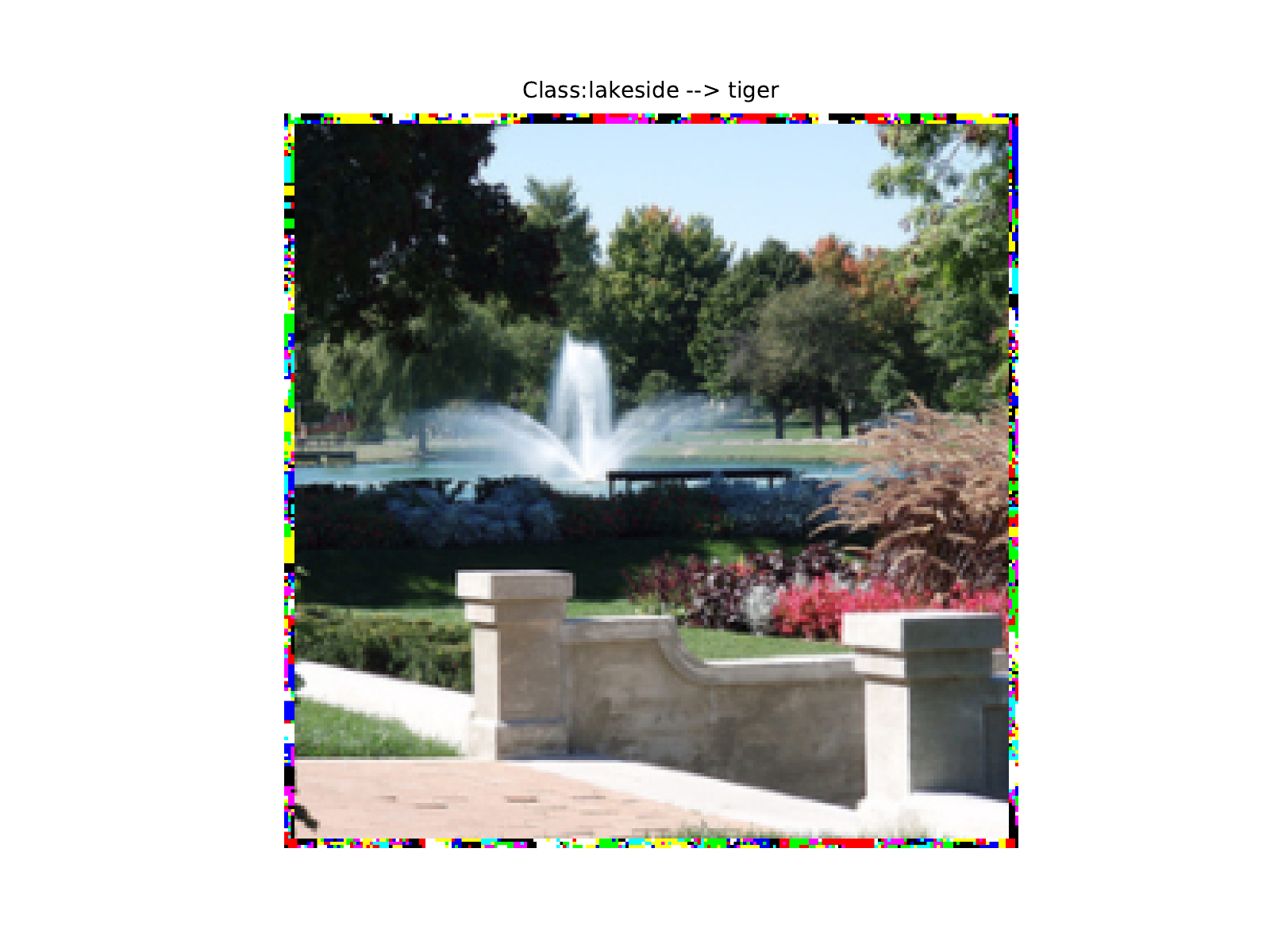} &
        \includegraphics[clip, trim=35mm 13mm 32mm 14mm, width=0.34\columnwidth]{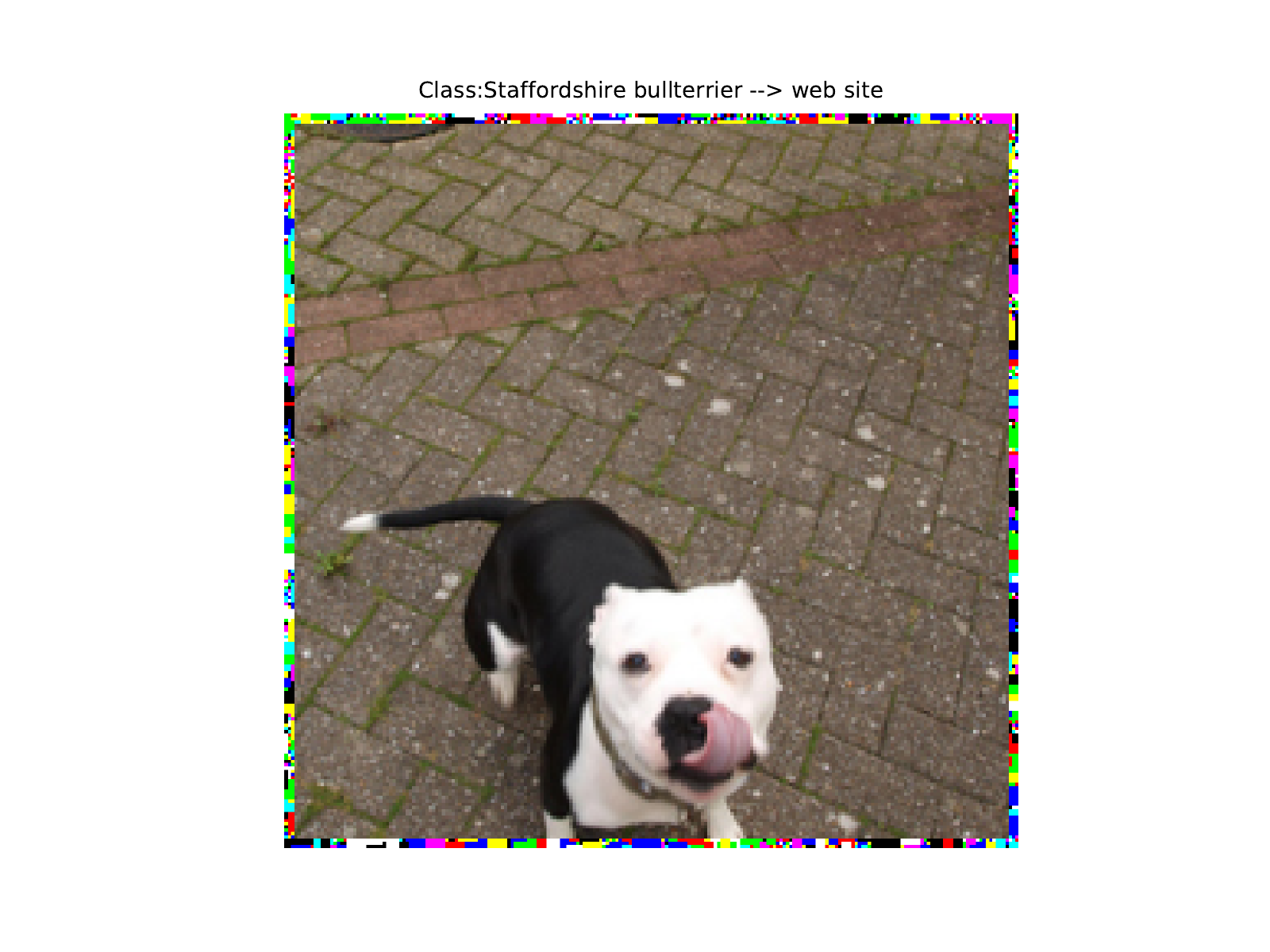}
    \end{tabular}
    \caption{Image-specific untargeted (2 pixels wide) and targeted (3 pixels) frames generated by our \texttt{Frame-RS} algorithm.} \label{fig:image_spec_frames_visualizations}%
\end{figure*}

\subsection{General algorithm for frames}
Unlike patches (Sec.~\ref{sec:patches}), for frames the set of pixels $M$ which can be perturbed (see Alg.~\ref{alg:rs_general}) is fixed. However, similarly to patches, the pixels of the frame can be changed arbitrarily in $[0,1]^3$ and thus we can use $l_\infty$-black-box attacks for step~7 in Alg.~\ref{alg:rs_general} which we denote as \texttt{Sparse-RS}~+~SH and \texttt{Sparse-RS}~+~SA. While SH is agnostic to the shape of the perturbations defined by adversarial frames and can be successfully applied for step~7 without further modifications, \texttt{Sparse-RS}~+~SA leads to suboptimal performance (see Table~\ref{tab:image_spec_frame}). Indeed, in the frame setting, the sampling distribution of SA requires its square updates to be entirely contained in the frame which due to its small width strongly limits the potential updates. Thus, to achieve better performance, we propose a new sampling distribution specific to frames which, combined with \rsa{} framework, we call \texttt{Frame-RS}.

\subsection{\textbf{\texttt{Frames-RS}} algorithm}
Our new sampling distribution (used for step~7 of Alg.~\ref{alg:rs_general}) returns squares that only \textit{partially} overlap with the frame. Effectively, we have more flexible updates, often of non-squared shape. Moreover, similarly to \texttt{Patch-RS}, we introduce single-channel updates in the final phase. We also use a different initialization in step~2: uniformly random sampling instead of vertical stripes. As a result, \texttt{Frame-RS} shows significantly better results than both \texttt{Sparse-RS}~+~SH and \texttt{Sparse-RS}~+~SA.

We initialize the components of the frame with random values in $\{0,1\}$. Then, at each iteration $i$ we sample uniformly a pixel of the frame to be a random corner of a square of size $s^{(i)} \times s^{(i)}$: all the pixels in the intersection of such square with the frame are set to the same randomly chosen color in $\{0, 1\}^c$ to create the candidate update. The length of the side of the square is regulated by a linearly decreasing schedule ($s^{(i)}= 3 \lceil \alpha_\text{init}\cdot w^2 \cdot (0.5 - i/N) \rceil$, with $N$ the total number of queries available and $w$ the width of the frame) for the first 50\% of queries, after which $s^{(i)}=1$. Finally, for the last 75\% of iterations, only single-channel updates are performed. Similarly to the other threat models, the only hyperparameter is $\alpha_\text{init}$.

\subsection{Competitors}
\textbf{\texttt{Sparse-RS} + SH and \rsa{} + SA.} SignHunter %
can be directly used to generate adversarial frames as this method does not require a particular shape of the perturbation set. Square Attack is specifically designed for rectangular images and we adapted it to this setting by constraining the sampled updates to lay inside the frame, as they are contained in the image for the $l_\infty$- and $l_2$- threat models. Since the perturbation set has a different shape than in the other threat models a new schedule for the size of squares specific for this case: at iteration $i$ of $N$ and with frames of width $w$, we have $s^{(i)}= \lceil \alpha_\text{init}\cdot w \cdot (0.5 - i/N) \rceil$ until $i<N/2$, then $s^{(i)}=1$ (we set $\alpha_\text{init}=2$).

\textbf{Adversarial framing (AF).} In \cite{zajac2019adversarial} propose a method based on gradient descent to craft universal adversarial frames. We here use standard PGD on the cross-entropy loss to generate image-specific perturbations by updating only the components corresponding to the pixels belonging to the frame. In details, we use 1000 iterations with step size 0.05 using the sign of the gradient as direction and each element is projected on the interval $[0,1]$. This white-box attack achieves 100\% of success in all settings we tested (consider that 1176 and 2652 pixels be perturbed for the untargeted and targeted scenarios respectively). For the black-box version, we use the gradient estimation technique as for patches (Sec.~\ref{sec:app_image_specific_patches}), that is Alg.~\ref{alg:ge} with the current iterate as input $x$, $m=5$, $\eta=10/\sqrt{d}$ and the gradient estimated at the previous iteration as prior $p$, where $d$ represents the number of features in the frame. Moreover, we use step size 0.1 for the untargeted case, 0.01 for the targeted one. Note that we found the mentioned values via grid search to optimize the success rate. We iterate the procedure, which costs 10 queries of the classifier, until reaching the maximum number of allowed queries or finding a successful adversarial perturbation.

\subsection{Experimental evaluation}
For untargeted attacks we use frames of width 2 and a maximum of 10,000 queries, and for targeted attacks we use frames of width 3 and maximally 50,000 queries. In Table~\ref{tab:image_spec_frame} we show that our \texttt{Frame-RS} achieves the best success rate (at least 10\% higher than the closest competitor) and much better query efficiency in all settings. In particular, \texttt{Frame-RS} significantly outperforms \texttt{Sparse-RS}~+~SA, showing that our new sampling distribution is crucial to obtain an effective attack. For the untargeted case the success rate of our black-box \texttt{Frame-RS} is close to that of the white-box AF method which achieves 100\% success rate. Finally we illustrate some examples of the resulting images in Fig.~\ref{fig:image_spec_frames_visualizations}.

\section{Targeted universal attacks: patches and frames}\label{sec:app_univ}
We here report the additional results and implementation details of the attacks for  targeted universal patches and frames.

\subsection{\textbf{\rsa{}} for targeted universal  attacks}\label{sec:app_scheme_univ_attacks}
Targeted universal  attacks have to generalize to unseen images and, in the case of patches, random locations on the image. We propose to generate them in a black-box scenario with the following scheme, with a budget of 100k iterations. We select a small batch of training images $\{x_i\}_{i=1}^n$ ($n=30$) and apply the patch at random positions on them (for frames the location is determined by its width). This positions are kept fixed for 10k iterations during which the patch is updated (step 7 of Alg.~\ref{alg:rs_general}) with some black-box attack (we use the same as for image-specific attacks in Sec.~\ref{sec:patches} and Sec.~\ref{sec:app_frames}) to minimize the loss
\begin{align} \label{eq:loss_univ_targ}
L_\text{targ}(\{x_i\}_{i=1}^{n})=\sum_{i=1}^{n}L_\text{CE}(f(x_i), t),\end{align}
where $t$ is the target class. Then, to foster generalization, we resample both the batch of training images and the locations of the patch over them (step 6 in Alg.~\ref{alg:rs_general}). In this way it is possible obtain black-box attacks without relying on surrogate models. Note that 30 queries of the classifiers are performed at every iteration of the algorithm.

In this scheme, as mentioned, we integrate either SignHunter %
or Square Attack %
to have \rsa{} + SH and \rsa{} + SA, and our novel \texttt{Patch-RS} introduced for image-specific attacks. In the case of patches, for \rsa{} + SA we set $p=0.05$ and for \texttt{Patch-RS} $\alpha_\text{init}=0.05$ (recall that both parameters control the schedule for sampling the updates in a similar way for the two attacks, details in Sec.~\ref{sec:app_image_specific_patches}). For frames, %
for SA we use the schedule for image-specific attacks (Sec.~\ref{sec:app_frames}) with $\alpha_\text{init}=0.7$, while in \texttt{Frame-RS} we adapt the schedule for image-specific attacks to the specific case, so that the size of the squares decrease to $s^{(i)}=1$ at $i=25,000$ and single-channel updates are sampled from $i=62,500$.

\subsection{Transfer-based attacks}
We evaluate the performance of targeted universal patches and frames generated by PGD \cite{MadEtAl2018} and MI-FGSM \cite{Dong_2018_CVPR} on a surrogate model and then transferred to attack the target model. %
In particular, at every iteration of the attack we apply on the training images (200) the patch at random positions (independently picked for every image), average the gradients of the cross-entropy loss at the patch for all images, and take a step in the direction of the sign of the gradients (we use 1000 iterations, step size of 0.05 and momentum coefficient 1 for MI-FGSM). 
For patches, we report the success rate as the average over 500 images, 10 random target classes, and 100 random positions of the patches for each image. For frames, we average over 5000 images and the same 10 random target classes.
In Table~\ref{tab:transfer_attacks} we show the success rate of the transfer based attacks on the surrogate model but with images unseen when generating the perturbation and on the target model (we use VGG and ResNet alternatively as the source and target models). We see that the attacks achieve a very high success rate when applied to new images on the source model, but are not able to generalize to another architecture. We note that \cite{BroEtAl2017} also report a similarly low success rate for the same size of the patch ($50\times50$ which corresponds to approximately 5\% of the input image, see Fig.~3 in \cite{BroEtAl2017}) on the target class they consider. %
Finally, we note that \cite{Dong_2018_CVPR} show that the perturbations obtained with MI-FGSM have better transferability compared to PGD, while in our experiments they are slightly worse, but we highlight that such observation was done for image-specific $l_\infty$- and $l_2$-attacks, which are significantly different threat models from what we consider here.

\begin{table}[h] \centering \small
\setlength{\tabcolsep}{4pt}
\begin{tabular}{c c | c c |c c} &\multirow{2}{*}{\textit{attack}} &  \multicolumn{4}{c}{\textit{succ. rate}} \\ && V $\rightarrow $ V& V $\rightarrow$ R & R $\rightarrow$ R & R $\rightarrow$ V \\
\toprule
\multirow{2}{*}{\textbf{patches}} & PGD %
& 99.6\% & 0.0\% & 94.9\% &3.3\%\\
&MI-FGSM %
& 99.1\% & 0.0\% & 92.6\% & 1.3\%\\
\midrule
\multirow{2}{*}{\textbf{frames}}&PGD %
& 99.4\% & 0.0\% & 99.8\% & 0.0\%\\
&MI-FGSM %
& 99.3\% & 0.0\% & 99.7\% & 0.0\%\\
\bottomrule
\end{tabular} %
\caption{Success rate of transfer-based targeted universal patches, with VGG (V) and ResNet (R) used as either source or target model averaged over 10 classes. When source and target model are the same network, the attack is only used on unseen images. These targeted universal attacks are very effective when transferred to unseen images on the source model, but do not generalize to the other architecture.} \label{tab:transfer_attacks}
\end{table}

\subsection{Targeted universal attacks via gradient estimation}
\textbf{PGD w/ GE.} Consistently with the other threat models, we test a method based on PGD with gradient estimation. In particular, in the scheme introduced in Sec.~\ref{sec:app_scheme_univ_attacks} for \rsa{}, we optimize the loss with the PGD attack \cite{MadEtAl2018} and the gradient estimation technique via finite differences as in Alg.~\ref{alg:ge}, similarly to what is done for image-specific perturbations in Sec.~\ref{sec:app_image_specific_patches} and Sec.~\ref{sec:app_frames} (we also use the same hyperparameters, step size 0.1 for PGD).

\textbf{ZO-AdaMM.} \cite{chen2019zo} propose a black-box method for universal attacks of minimal $l_2$-norm able to fool a batch of images, without testing them on unseen images. We adapt it to the case of patches and frames removing the penalization term in the loss which minimizes the $l_2$-norm of the perturbations since it is not necessary in our threat models, and integrate it in our scheme introduced in Sec.~\ref{sec:app_scheme_univ_attacks}. We used the standard parameters and tuned learning rate, setting 0.1 for patches, 0.5 for frames. In particular, 10 random perturbations, each equivalent to one query, of the current universal perturbation are sampled to estimate the gradient at each iterations, recalling that the total budget is 100,000 queries.

\begin{table*}[t] \centering \small
\begin{tabular}{c| c c |c c} \multirow{2}{*}{\textit{attack}} & \multicolumn{2}{c}{\textbf{patches}} & \multicolumn{2}{c}{\textbf{frames}}\\ & VGG & ResNet& VGG & ResNet \\
\toprule
Transfer PGD %
& 3.3\%& 0.0\% & 0.0\%& 0.0\%\\
Transfer MI-FGSM %
& 1.3\%& 0.0\% & 0.0\% & 0.0\%\\
PGD w/ GE & 35.1\%& 15.5\%& 22.7\% & 52.4\%\\
ZO-AdaMM %
& 45.8\% & 6.0\% & 41.8\% & 53.5\% \\
\texttt{Sparse-RS} + SH %
&63.9\%  & 13.8\% & 48.8\% & \textbf{65.5\%}\\
\texttt{Sparse-RS} + SA %
& \textbf{72.9\% $\pm$ 3.6}& 29.6\% $\pm$ 5.6 & 43.0\% $\pm$ 1.2 & 63.6\% $\pm$ 2.1\\
\texttt{Patch-RS} / \texttt{Frame-RS}& 70.8\% $\pm$ 1.3 & \textbf{30.4\% $\pm$ 8.0} & \textbf{55.5\% $\pm$ 0.3} & 65.3\% $\pm$ 2.3\\
\bottomrule
\end{tabular} %
\caption{Average success rate of targeted  universal attacks over 10 target classes on VGG and ResNet. We use patches of size $50\times 50$ and frames of width $6$ pixels and repeat the attacks based on random search with 3 random seeds.}
\label{tab:univ_attacks}\end{table*}

\begin{figure*}[t] \centering 
    \setlength{\tabcolsep}{1pt}
    \begin{tabular}{ccc|ccc}
        \multicolumn{3}{c|}{\textbf{Targeted universal patches}} & \multicolumn{3}{c}{\textbf{Targeted universal frames}} \\
        \scriptsize Sulphur butterfly $\rightarrow$ Rottweiler & 
        \scriptsize Persian cat $\rightarrow$ Slug & 
        \scriptsize Starfish $\rightarrow$ Polecat & 
        \scriptsize African elephant $\rightarrow$ Slug & 
        \scriptsize Projector $\rightarrow$ Colobus (monkey) & 
        \scriptsize Vine snake $\rightarrow$ Butcher shop \\
        \includegraphics[clip, trim=1mm 1mm 1mm 1mm, width=0.34\columnwidth]{universal_targeted_patches-img=1-true=sulphur_butterfly_orig=sulphur_butterfly_final=Rottweiler_target=Rottweiler.pdf} &
        \includegraphics[clip, trim=1mm 1mm 1mm 1mm, width=0.34\columnwidth]{universal_targeted_patches-img=19-true=Persian_cat_orig=tiger_cat_final=slug_target=slug.pdf} &
        \includegraphics[clip, trim=1mm 1mm 1mm 1mm, width=0.34\columnwidth]{universal_targeted_patches-img=24-true=starfish_orig=starfish_final=polecat_target=polecat.pdf} &
        \includegraphics[clip, trim=35mm 13mm 32mm 14mm, width=0.34\columnwidth]{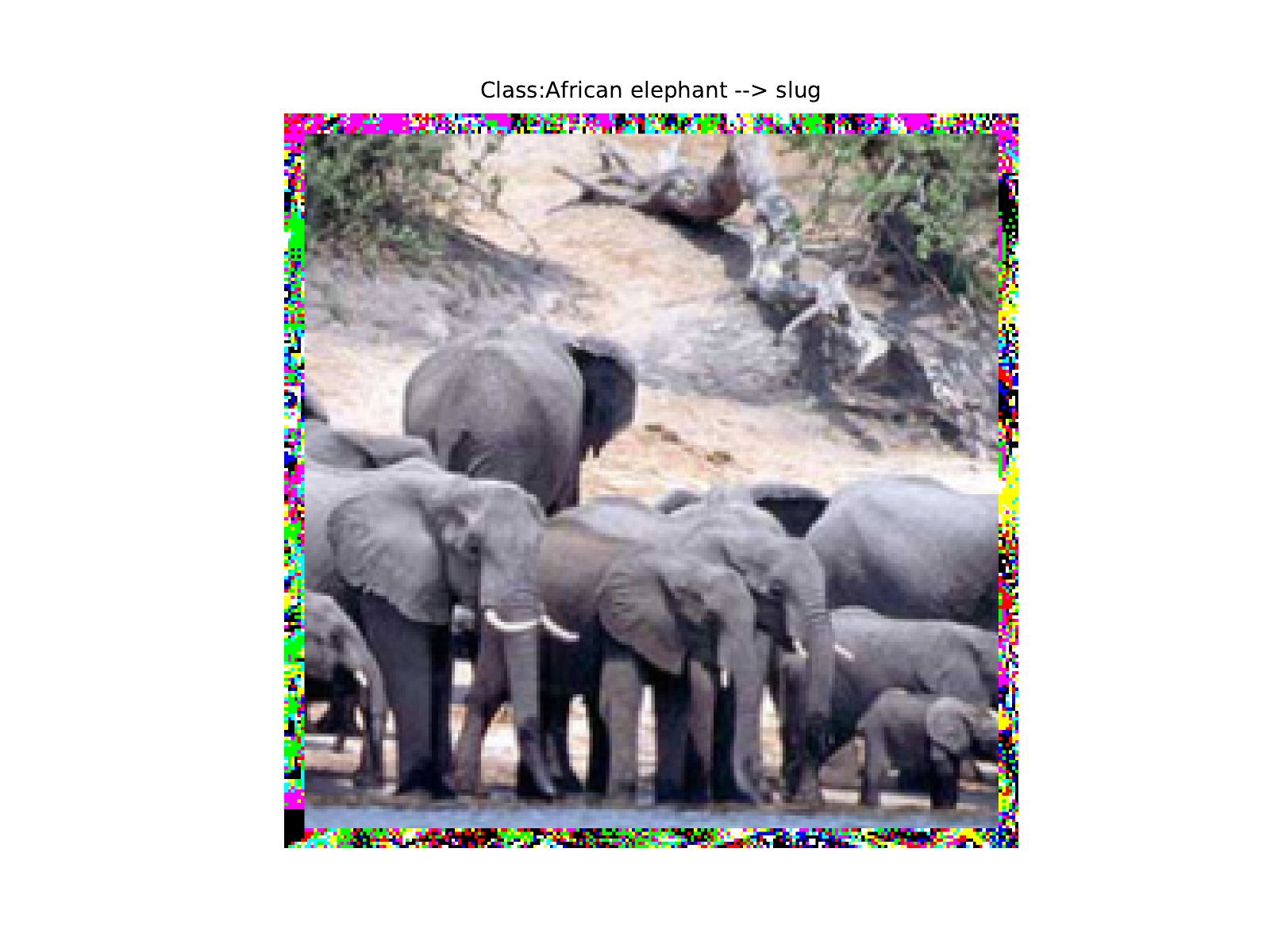} &
        \includegraphics[clip, trim=35mm 13mm 32mm 14mm, width=0.34\columnwidth]{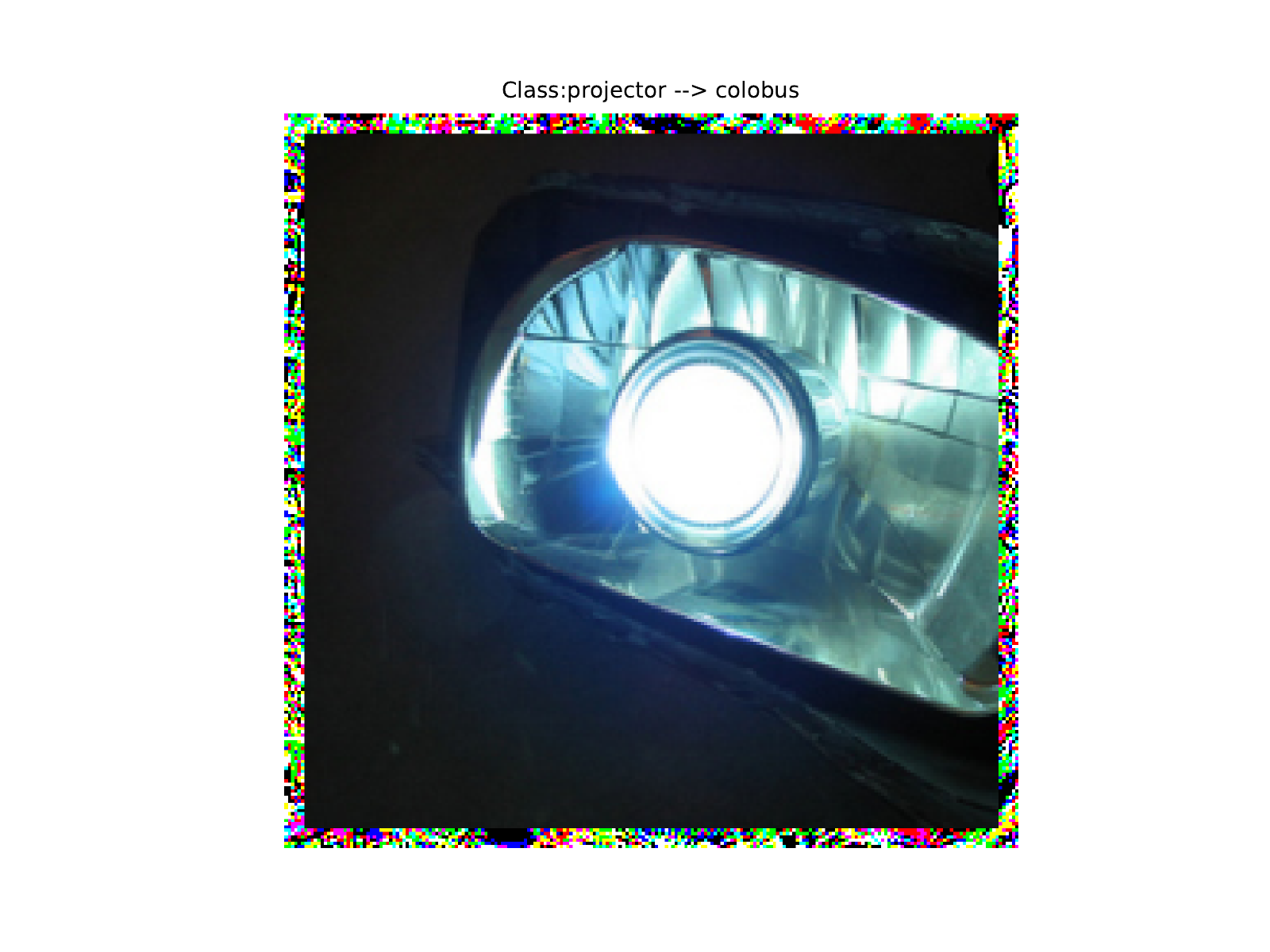} &
        \includegraphics[clip, trim=35mm 13mm 32mm 14mm, width=0.34\columnwidth]{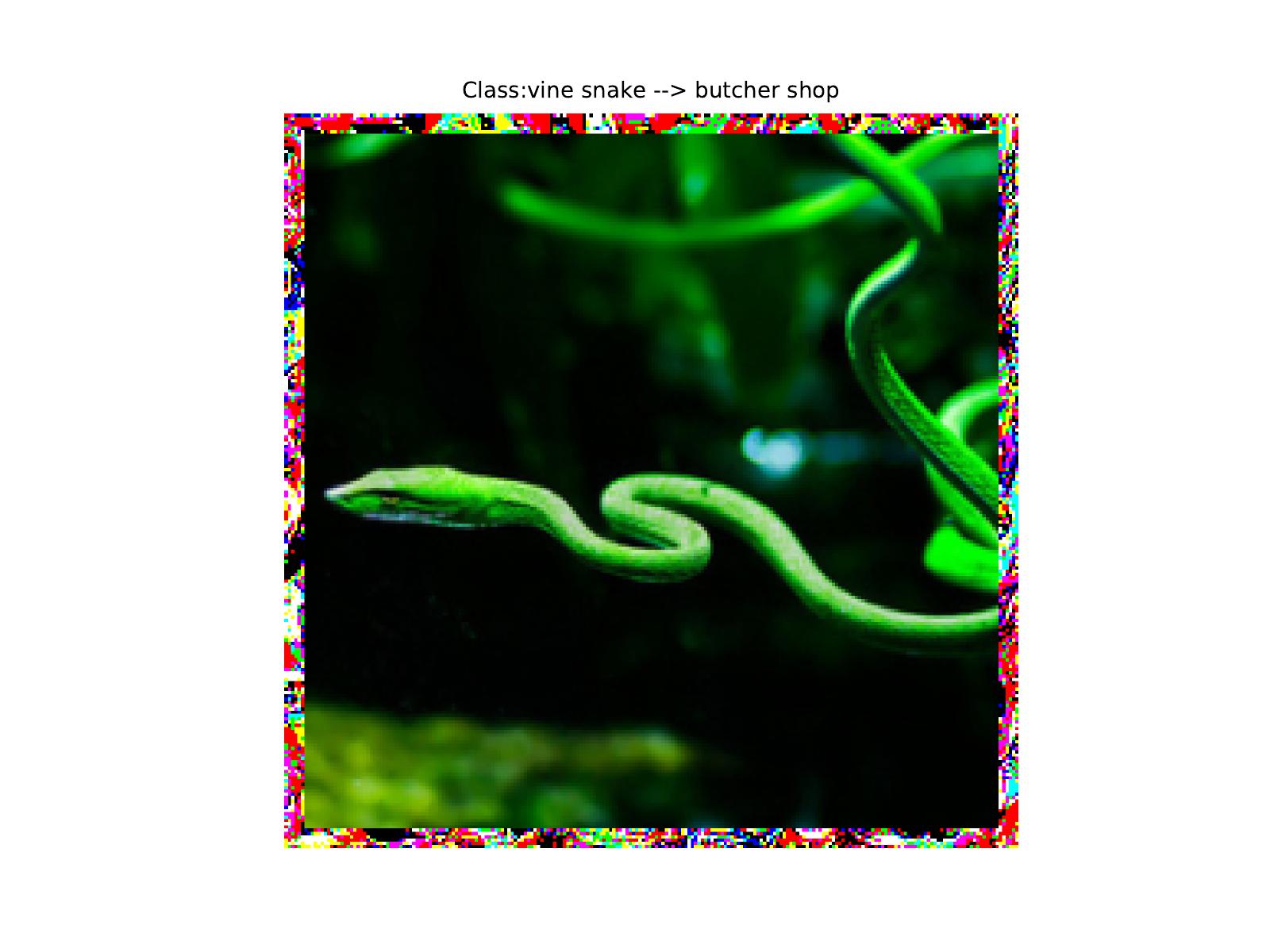} \\
        \scriptsize Echidna $\rightarrow$ Rottweiler & 
        \scriptsize Geyser $\rightarrow$ Slug & 
        \scriptsize Electric guitar $\rightarrow$ Polecat & 
        \scriptsize Ballpoint $\rightarrow$ Slug & 
        \scriptsize Grey whale $\rightarrow$ Colobus (monkey) & 
        \scriptsize Beacon $\rightarrow$ Butcher shop \\
        \includegraphics[clip, trim=1mm 1mm 1mm 1mm, width=0.34\columnwidth]{universal_targeted_patches-img=42-true=echidna_orig=echidna_final=Rottweiler_target=Rottweiler.pdf} &
        \includegraphics[clip, trim=1mm 1mm 1mm 1mm, width=0.34\columnwidth]{universal_targeted_patches-img=10-true=geyser_orig=geyser_final=slug_target=slug.pdf} &
        \includegraphics[clip, trim=1mm 1mm 1mm 1mm, width=0.34\columnwidth]{universal_targeted_patches-img=25-true=electric_guitar_orig=electric_guitar_final=polecat_target=polecat.pdf} &
        \includegraphics[clip, trim=35mm 13mm 32mm 14mm, width=0.34\columnwidth]{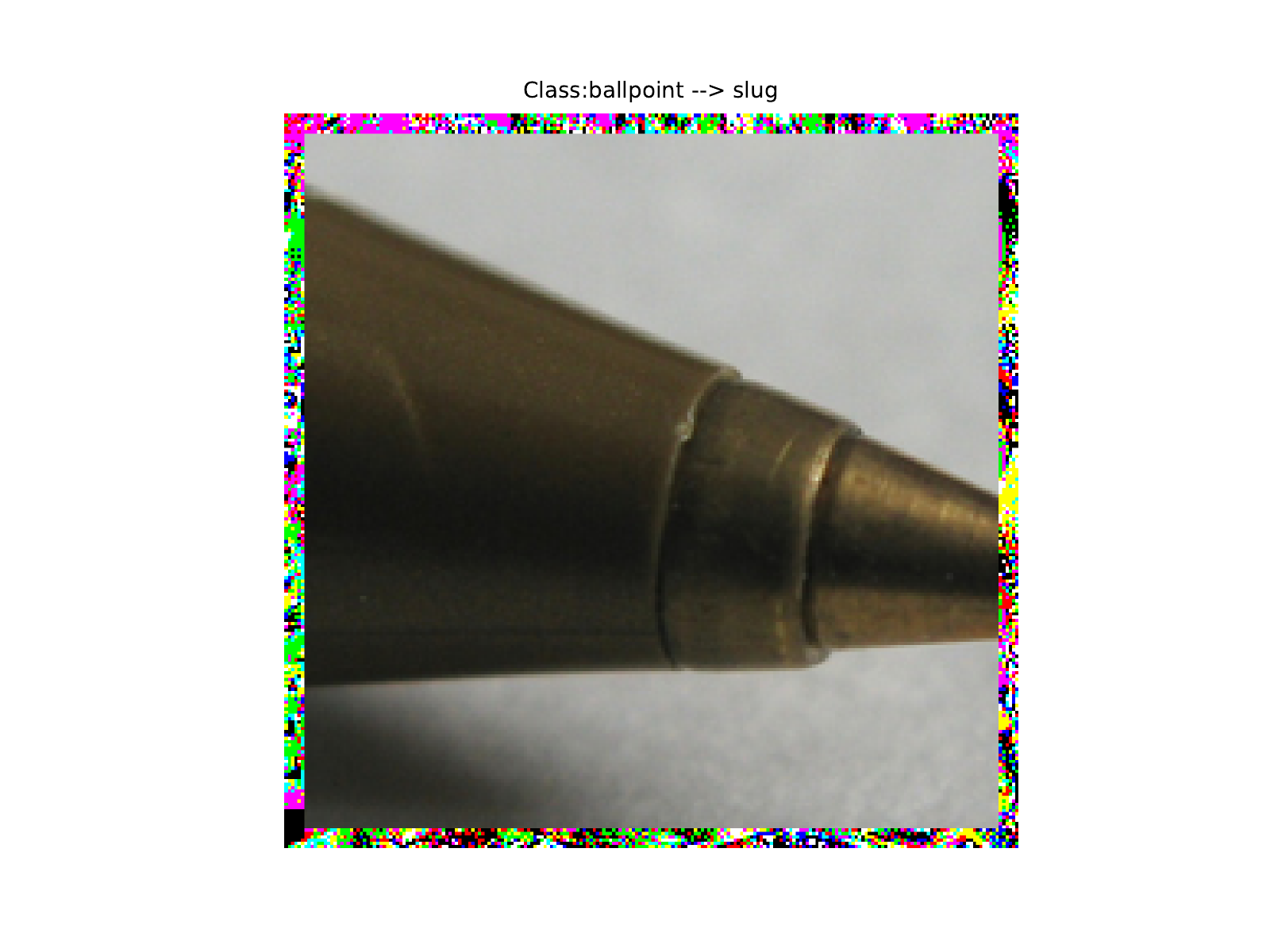} &
        \includegraphics[clip, trim=35mm 13mm 32mm 14mm, width=0.34\columnwidth]{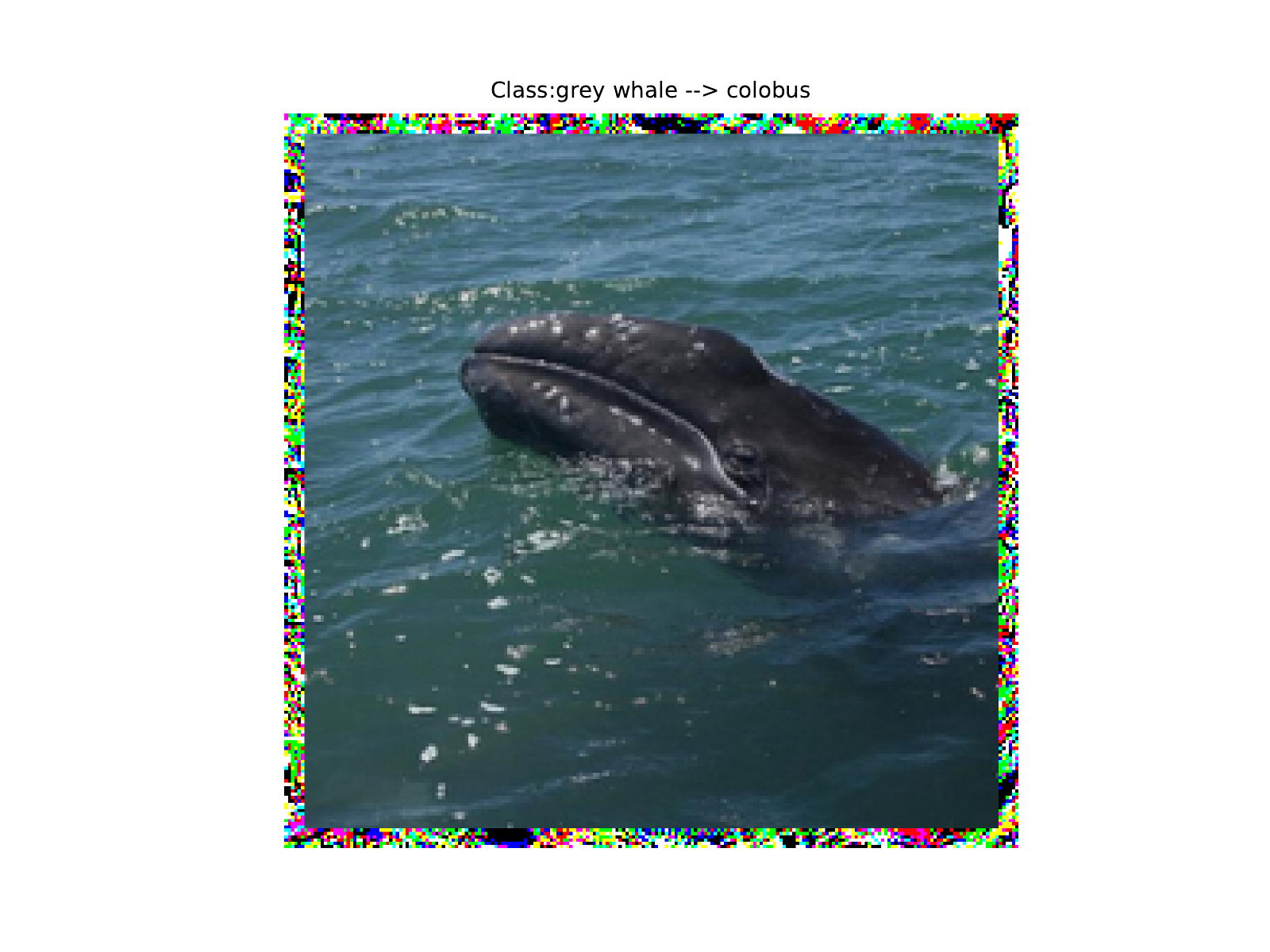} &
        \includegraphics[clip, trim=35mm 13mm 32mm 14mm, width=0.34\columnwidth]{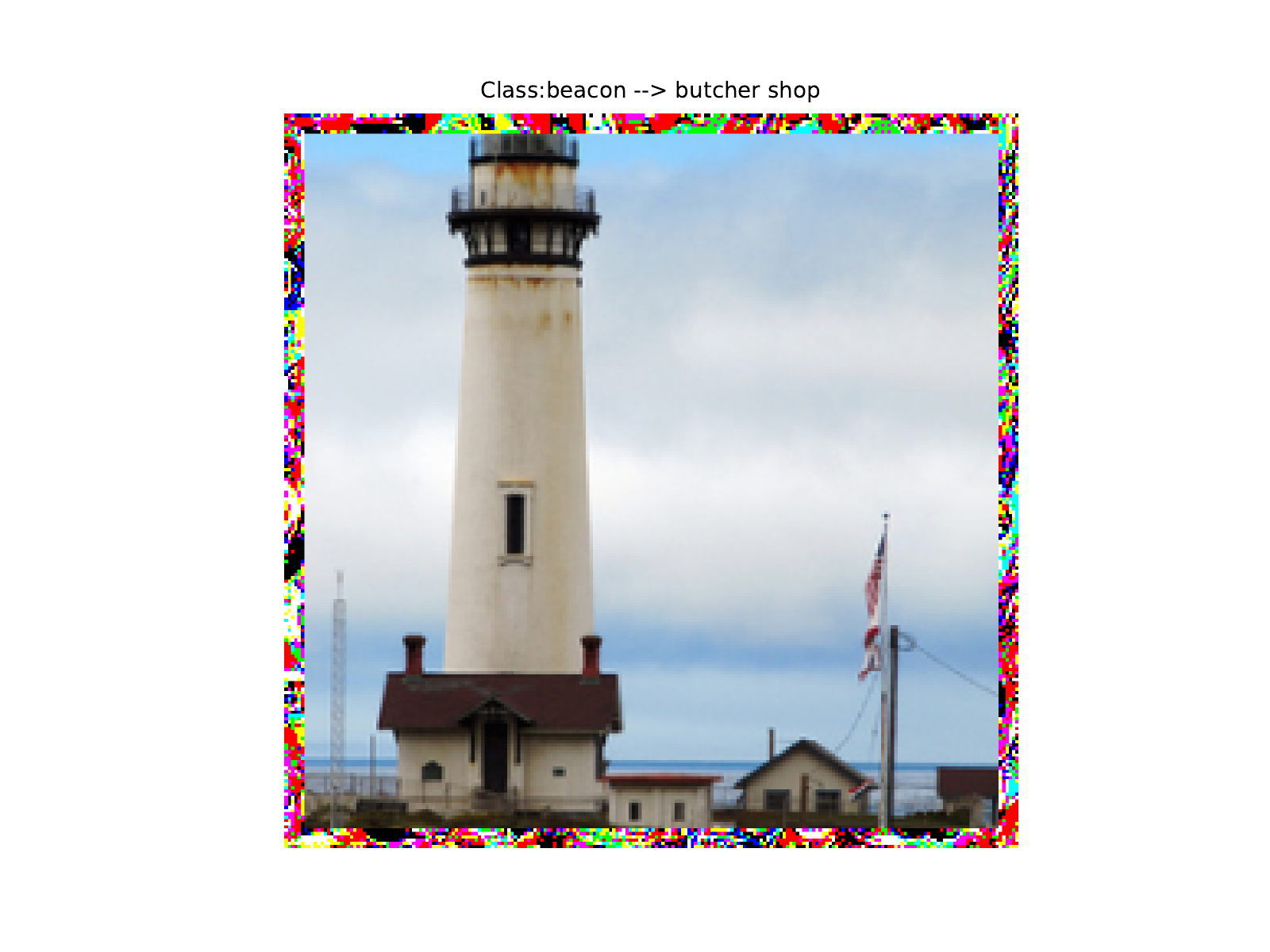}
    \end{tabular}
    \caption{Targeted universal patches and frames generated by our %
    \texttt{Patch-RS} and \texttt{Frame-RS}.
    We show two images in each column with the same universal patch/frame that is able to change the decision of the classifier into the desired target class.} \label{fig:univ_patches_frames_visualizations}
\end{figure*}

\subsection{Additional results}
To evaluate the success of a targeted  universal patch, we apply to each of 500 images the patch at 100 random locations and compute the percentage of the perturbed images which are classified in the target class. For targeted universal  frames, we compute the success rate applying the perturbation to 5,000 images (unseen during training).
The results in Table~\ref{tab:universal} in Sec.~\ref{sec:universal} are computed as mentioned above for 10 randomly chosen target classes and then averaged (the same set of target classes is used for patches and frames). For the random search based attacks, we repeat the attacks for 3 random seeds. %
We report the complete results for both threat models in Table~\ref{tab:univ_attacks}. Visualizations of the perturbations found by our scheme are shown in Fig.~\ref{fig:univ_patches_frames_visualizations}. Finally, we report the details of the success rate on each class in Table~\ref{tab:det_classes_patches} for adversarial patches and Table~\ref{tab:det_classes_frames} for adversarial frames.

\begin{table*}[p]
\centering \small \setlength{\tabcolsep}{4pt}
\begin{tabular}{c |c c c|c c c}
\multirow{2}{*}{\textit{class}} & \multicolumn{3}{c}{VGG} & \multicolumn{3}{c}{ResNet}\\
&\rsa{} + SH & \rsa{} + SA & \texttt{Patch-RS} &\rsa{} + SH & \rsa{} + SA & \texttt{Patch-RS}\\
\toprule
Rottweiler & 61.4\% & 77.1\% $\pm$ 14.6 & 83.4\% $\pm$ 3.9 & 0.2\% & 42.6\% $\pm$ 26.6 & 44.4\% $\pm$ 29.3 \\ 
Lakeland terrier & 65.4\% & 86.0\% $\pm$ 2.7 & 76.1\% $\pm$ 9.7 & 0.5\% & 0.6\% $\pm$ 0.2 & 12.2\% $\pm$ 14.8 \\ 
polecat & 46.7\% & 29.5\% $\pm$ 20.6 & 47.1\% $\pm$ 32.3 & 0.1\% & 27.3\% $\pm$ 15.5 & 15.0\% $\pm$ 9.7 \\ 
waffle iron & 86.3\% & 82.3\% $\pm$ 3.2 & 74.7\% $\pm$ 3.7 & 17.4\% & 16.2\% $\pm$ 3.9 & 18.6\% $\pm$ 4.9 \\ 
whippet & 37.7\% & 73.6\% $\pm$ 1.6 & 52.7\% $\pm$ 6.0 & 0.1\% & 36.9\% $\pm$ 24.0 & 35.4\% $\pm$ 24.0 \\ 
digital clock & 97.8\% & 97.6\% $\pm$ 0.2 & 97.0\% $\pm$ 0.5 & 90.5\% & 90.4\% $\pm$ 3.1 & 92.5\% $\pm$ 1.3 \\ 
colobus & 69.1\% & 68.9\% $\pm$ 5.2 & 80.1\% $\pm$ 4.7 & 0.0\% & 0.0\% $\pm$ 0.0 & 0.0\% $\pm$ 0.0 \\ 
slug & 61.0\% & 55.8\% $\pm$ 6.9 & 68.1\% $\pm$ 4.8 & 27.9\% & 43.7\% $\pm$ 1.7 & 44.2\% $\pm$ 10.7 \\ 
butcher shop & 42.4\% & 67.4\% $\pm$ 2.1 & 44.1\% $\pm$ 31.1 & 1.8\% & 2.4\% $\pm$ 0.6 & 1.0\% $\pm$ 0.5 \\ 
bluetick & 71.3\% & 90.5\% $\pm$ 1.1 & 85.0\% $\pm$ 2.4 & 0.0\% & 35.8\% $\pm$ 18.5 & 40.9\% $\pm$ 19.5\\
\bottomrule
\end{tabular} \vspace{0mm} \caption{Success rate of targeted universal patches for each of the 10 randomly chosen target classes.}
\label{tab:det_classes_patches}
\end{table*}

\begin{table*}[p]
\centering \small \setlength{\tabcolsep}{4pt}
\begin{tabular}{c |c c c|c c c}
\multirow{2}{*}{\textit{class}} & \multicolumn{3}{c}{VGG} & \multicolumn{3}{c}{ResNet}\\
&\rsa{} + SH & \rsa{} + SA & \texttt{Frame-RS} &\rsa{} + SH & \rsa{} + SA & \texttt{Frame-RS}\\
\toprule
Rottweiler & 
39.4\% & 34.9\% $\pm$ 7.5 & 61.8\% $\pm$ 8.5 & 79.9\% &77.6\% $\pm$ 4.5 & 81.6\% $\pm$ 8.2 \\ 
Lakeland terrier & 60.3\% & 45.6\% $\pm$ 9.7 & 69.9\% $\pm$ 8.3 & 81.9\% & 77.7\% $\pm$ 6.5 & 88.4\% $\pm$ 1.7 \\ 
polecat & 13.8\% & 31.2\% $\pm$ 1.5 & 33.5\% $\pm$ 2.8 & 17.7\% & 15.1\% $\pm$ 6.0 & 11.2\% $\pm$ 9.1 \\ 
waffle iron & 36.0\% & 35.5\% $\pm$ 3.9 & 49.4\% $\pm$ 6.8 & 81.2\% & 65.2\% $\pm$ 11.4 & 70.5\% $\pm$ 8.0 \\ 
whippet & 63.6\% & 22.1\% $\pm$ 2.4 & 44.5\% $\pm$ 11.1 & 51.0\% & 63.4\% $\pm$ 9.0 & 51.5\% $\pm$ 2.5 \\ 
digital clock & 95.3\% & 96.0\% $\pm$ 0.2 & 96.5\% $\pm$ 0.7 & 99.4\% & 99.6\% $\pm$ 0.2 & 99.8\% $\pm$ 0.1 \\ 
colobus & 3.5\% & 3.9\% $\pm$ 2.1 & 2.5\% $\pm$ 0.7 & 16.8\% & 0.5\% $\pm$ 0.1 & 4.1\% $\pm$ 2.9 \\ 
slug & 51.1\% & 45.1\% $\pm$ 4.7 & 50.7\% $\pm$ 3.5 & 73.8\% & 68.1\% $\pm$ 6.4 & 72.5\% $\pm$ 6.3 \\ 
butcher shop & 78.9\% & 67.9\% $\pm$ 6.3 & 83.5\% $\pm$ 3.3 & 92.2\% & 88.9\% $\pm$ 2.5 & 91.6\% $\pm$ 3.9 \\ 
bluetick & 46.2\% & 48.0\% $\pm$ 3.9 & 65.1\% $\pm$ 7.4 & 61.3\% & 79.2\% $\pm$ 1.9 & 82.0\% $\pm$ 4.2\\
\bottomrule
\end{tabular} \vspace{0mm} \caption{Success rate of targeted  universal frames for each of the 10 randomly chosen target classes.}
\label{tab:det_classes_frames}
\end{table*}

\section{Theoretical analysis of \texttt{$l_0$-RS}}\label{sec:app_theory}
In this section we prove the following proposition which shows that  \texttt{$l_0$-RS} requires a sublinear number of queries (improving upon  the linear number of queries needed by naive gradient estimation approaches).
\begin{customprop}{\ref{prop:l0_rs}}
    The expected number $t_k$ of queries needed for \texttt{$l_0$-RS} with $\iter{\alpha}{i} = \nicefrac{1}{k}$ to find a set of $k$ weights out of the smallest $m$ weights of a linear model is:  %
    \begin{align*}
        \Exp \left[ t_k \right] = (d-k)k\sum_{i=0}^{k-1} \frac{1}{(k-i)(m-i)}  < (d-k)k \frac{\ln(k) + 2}{m-k}. 
    \end{align*}
\end{customprop}
\begin{proof}
    According to the \texttt{$l_0$-RS} algorithm, we have $d$ features grouped in a set $U$ and the goal is to find a set $M\subset U$ containing $k$ elements among the $m$ smallest elements of $U$.
    Since $\alpha^{(i)} = \nicefrac{1}{k}$, at every iteration we pick one element $p\in M$ to remove from $M$ and one element $q\in U\setminus M$ to add to $M$. This results in a binary vector $z_{new} \in \{0, 1\}^d$ indicating which are the features in $M$. Then we query the black-box linear model to determine whether the loss at a new point $z_{new}$ improves compared to the point on the previous iteration $z_{current}$, i.e. $L(z_{new}) < L(z_{current})$, which gives us information whether $q < p$.
    
    If the current set $M$ contains $i$ elements which belong to the smallest $m$, the probability of increasing it to $i+1$ elements with the next pick is equal to 
    \begin{align*}
    \P[i \rightarrow i+1] =& \P[p \notin \text{smallest }m, q \in \text{smallest }m]\\ =& \frac{k-i}{k} \cdot \frac{m - i}{d-k}.
    \end{align*}
    Then the expected number of iterations for the step $i \rightarrow i+1$ is equal to
    \[ \Exp[t_{i+1} - t_i] = \frac{(d-k)k}{(k-i)(m-i)}\] 
    since all the draws are independent. Assuming that we begin with none of the smallest $m$ elements in $M$, the expected number of iterations $t_k$ needed to find a set of $k$ weights out of the smallest $m$ weights is given by 
    \begin{equation} \label{eq:expectation_exact} \begin{split}
        \Exp[t_k] =& \sum_{i=0}^{k-1} \Exp[t_{i+1} - t_i] = \sum_{i=0}^{k-1} \frac{(d-k)k}{(k-i)(m-i)}\\ =& (d-k)k \sum_{i=0}^{k-1} \frac{1}{(k-i)(m-i)}.         
    \end{split} \end{equation}
    Now assuming that $m > k$, we can write the summation as:
    \begin{equation}\begin{split}
        \sum_{i=0}^{k-1} \frac{1}{(k-i)(m-i)} &= \sum_{j=1}^{k} \frac{1}{j (j + m-k)}\\ &= \frac{1}{m-k} \sum_{j=1}^{k} \left( \frac{1}{j} - \frac{1}{j + m-k} \right)  \\
                                              &= \frac{1}{m-k} \left( H_k - H_m + H_{m-k} \right),  \label{eq:prop_summation}
    \end{split} 
    \end{equation}
    where $H_k$ is the $k$-th harmonic number.
    Using the fact that $\ln(k) < H_k \leq \ln(k) + 1$, we have
    \begin{align} \nonumber
         H_k - H_m + H_{m-k}  <  \ln(k) - \ln(m) + \ln(m-k) + 2 %
    \end{align}
    and \[\ln(k) - \ln(m) + \ln(m-k) + 2 < \ln(k) + 2.\]
    If we combine this result with Eq.~\eqref{eq:expectation_exact} and Eq.~\eqref{eq:prop_summation}, we obtain the desired upper bound on $\Exp \left[ t_k \right]$: 
    \[ \Exp \left[ t_k \right] < (d-k)k \frac{\ln(k) + 2}{m-k}. \]
\end{proof}

\begin{remark} We further remark that for $m = k$ we get that
\[
\Exp[t_k] = (d-k)k \sum_{j=1}^{k} \frac{1}{j^2} < \frac{\pi^2}{6} (d-k)k, 
\]
however we are interested in the setting when the gap $m-k$ is large enough so that $\Exp \left[ t_k \right]$ becomes sublinear.
\end{remark}
The main conclusion from Proposition~\ref{prop:l0_rs} is that $\Exp \left[ t_k \right]$ becomes sublinear for large enough gap $m-k$ which we further illustrate in Fig.~\ref{fig:theory_exps} where we plot %
the expected number of queries needed by  \texttt{$l_0$-RS}  for $d = 150,528$ and $k = 150$ which is equivalent to our ImageNet experiments  with $50$ pixels perturbed.
\begin{figure}[t]
    \centering \small
    \includegraphics[width=0.85\columnwidth]{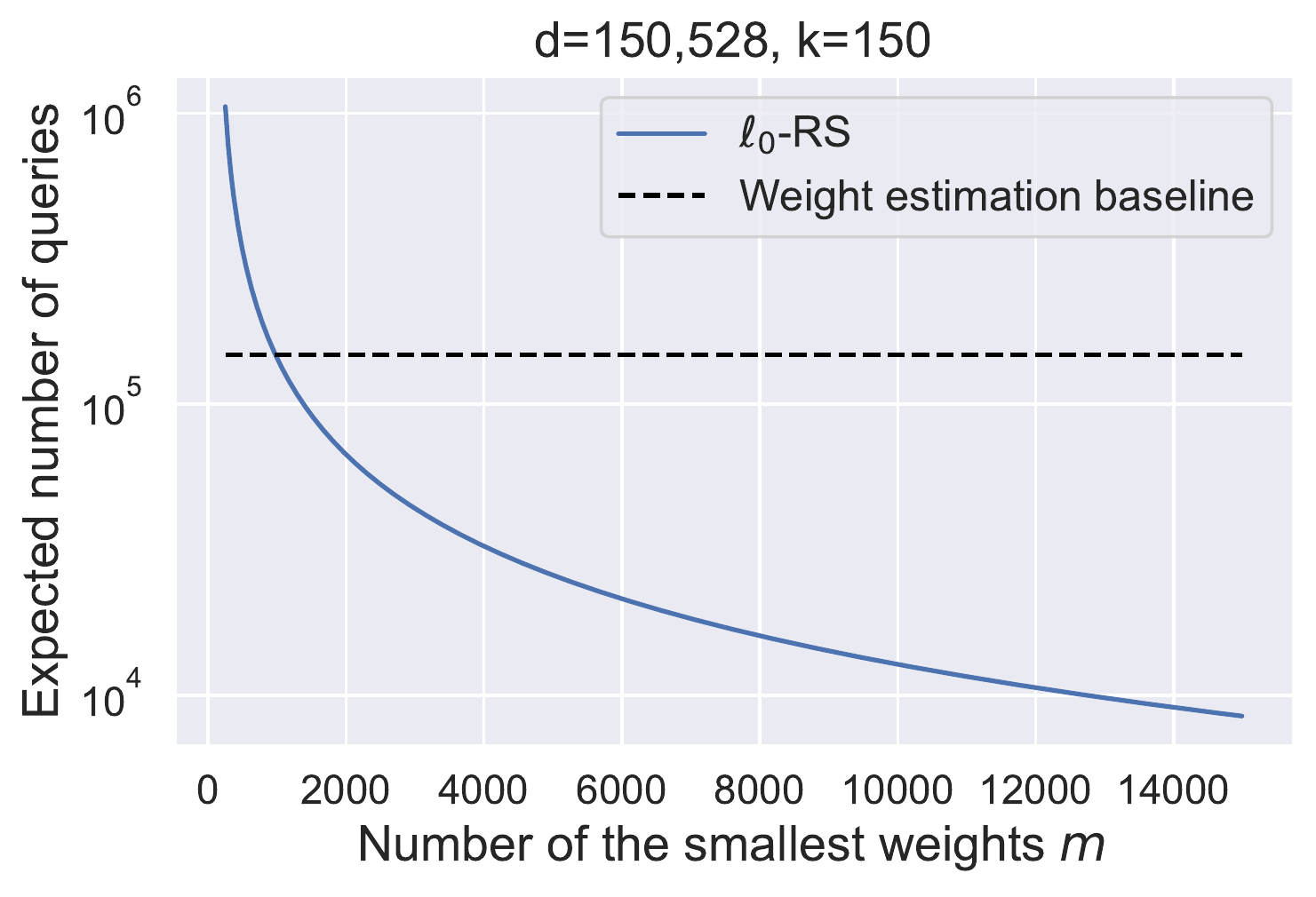}
    \caption{
        \small Comparison of the query efficiency of \texttt{$l_0$-RS} to that of the naive weight estimation baseline which requires $O(d)$ queries to estimate the gradient.
    }
    \label{fig:theory_exps}
\end{figure}

\section{Ablation studies} \label{sec:app_ablation_st}
We here present a series of ablation studies to better understand the robustness of our algorithms to different random seeds, how its performance changes depending on $\alpha_\text{init}$, and to justify the algorithmic choice of the piecewise decaying schedule for $\iter{\alpha}{i}$. 
We focus the ablation study only on \texttt{$l_0$-RS} and \texttt{Patch-RS} methods since for other threat models the setting is less challenging as there is no need to optimize the location of the perturbation. %

\subsection{$l_0$-bounded attacks}
\label{app:subsec_l0_ablation}
\textbf{Different random seeds.} First, we study how the performance of \texttt{$l_0$-RS} in the pixel space varies when using different random seeds, which influence the stochastic component inherent in random search. In Table~\ref{tab:abl_seed} we report mean and standard deviation over 10 runs (with different seeds) of success rate, average and median queries of \texttt{$l_0$-RS} on VGG and ResNet with sparsity levels $k\in\{50, 100, 150\}$ (the same setup of Sec.~\ref{sec:l0_imagenet}, with the additional $k=100$). One can observe that the success rate is very stable in all the cases and the statistics of query consumption consistent across different runs.

\begin{table*}[t]\centering \small
    \begin{tabular}{c c| c| c c |c c} \multirow{2}{*}{\textit{model}}& \multirow{2}{*}{$k$} &\multirow{2}{*}{\textit{success rate (\%)}} & \multicolumn{2}{c|}{\textit{successful points}} & \multicolumn{2}{c}{\textit{all points}}\\
    & & &\textit{avg. queries}&\textit{med. queries}& \textit{avg. queries}&\textit{med. queries}\\
    \toprule
    \multirow{3}{*}{VGG} & $50$ & $97.4 \pm 0.42$ & $497 \pm 28$ & $80 \pm 4$ & $745 \pm 43$ & $86 \pm 5$\\& $100$ & $99.8 \pm 0.14$ & $320 \pm 20$ & $42 \pm 3$ & $335 \pm 17$ & $43 \pm 3$\\& $150$ & $100.0 \pm 0.00$ & $193 \pm 16$ & $26 \pm 2$ & $193 \pm 16$ & $26 \pm 2$ \\
    \midrule
    \multirow{3}{*}{ResNet} & $50$ & $94.6 \pm 0.70$ & $686 \pm 47$ & $135 \pm 6$ & $1187 \pm 48$ & $166 \pm 11$\\& $100$ & $99.7 \pm 0.15$ & $544 \pm 36$ & $73 \pm 5$ & $571 \pm 26$ & $74 \pm 5$\\& $150$ & $100.0 \pm 0.00$ & $365 \pm 19$ & $49 \pm 3$ & $365 \pm 19$ & $49 \pm 3$ \\
    \bottomrule
    \end{tabular}%
    \caption{Mean and standard deviation of the success rate and query efficiency of untargeted \texttt{$l_0$-RS} repeated with 10 different random seeds. The success rate and query efficiency are very stable over random seeds.}\label{tab:abl_seed}
\end{table*}

\textbf{Different values of $\alpha_\text{init}$.} Then we analyze the behaviour of our algorithm with different values of $\alpha_\text{init}$, since it is the only free hyperparameter of \rsa{}. Let us recall that it is used to regulate how much $M'$ and $\Delta'$ differ from $M$ and $\Delta$ respectively in steps 6-7 of Alg.~\ref{alg:rs_general} at each iteration.
Fig.~\ref{fig:abl_alpha_init} shows the success rate and query usage (computed on the successful samples) of our untargeted $l_0$-bounded attack on VGG at the usual three sparsity levels $k$ for runs with $\alpha_\text{init}\in \{0.01, 0.05, 0.1, 0.2, 0.3, 0.4, 0.6, 0.8, 1\}$ (for our experiments in Sec.~\ref{sec:l0_imagenet} we use $\alpha_\text{init}=0.3$). We observe that the success rate is similar (close to $100\%$) for all the values, with a slight degradation for the largest ones. In order to minimize the queries of the classifier, the runs with $\alpha_\text{init}$ between $0.1$ and $0.4$ are comparably good, with small differences in the tradeoff between average and median number of queries.

\textbf{Constant vs decaying $\iter{\alpha}{i}$.} In order to demonstrate the role of decaying the difference between the candidate updates $M'$ and $\Delta'$ and the current iterates $M$ and $\Delta$ over iterations  (see steps 6-7 of Alg.~\ref{alg:rs_general}) to achieve good performance, we run our attack with constant $\iter{\alpha}{i}$ schedule instead of the piecewise constant schedule with decreasing values.
We fix $\iter{\alpha}{i}=c \in [0, 1]$ for every $i$
so that for the whole algorithm $M'$ and $M$ differ in $\max\{c \cdot k, 1\}$ elements. In Fig.~\ref{fig:abl_constant_alpha} we report the results achieved by $c\in\{0, 0.05, 0.15, 0.3, 0.5, 1\}$ on VGG at $k\in\{50, 100, 150\}$, together with the baseline (black dashed line) of the standard version of \texttt{$l_0$-RS}. One can observe that small constant values $c$ for $\iter{\alpha}{i}=c$ achieve a high success rate but suffer in query efficiency,
in particular computed regarding the median and for larger $k$, while the highest values of $c$ lead to significantly worse success rate (note that average and median queries are computed only on the successful points) than the baseline. These results highlight how important it is to have an initial exploration phase, with a larger step size, and at a later stage a more local optimization.

\begin{figure*}[t] \centering \small
\setlength{\tabcolsep}{1pt}
\begin{tabular}{c c} \begin{tabular}{c} $k=50$ \\[22mm]
$k=100$\\[22mm] $k=150$
\end{tabular} & \includegraphics[align=c, width=1.8\columnwidth]{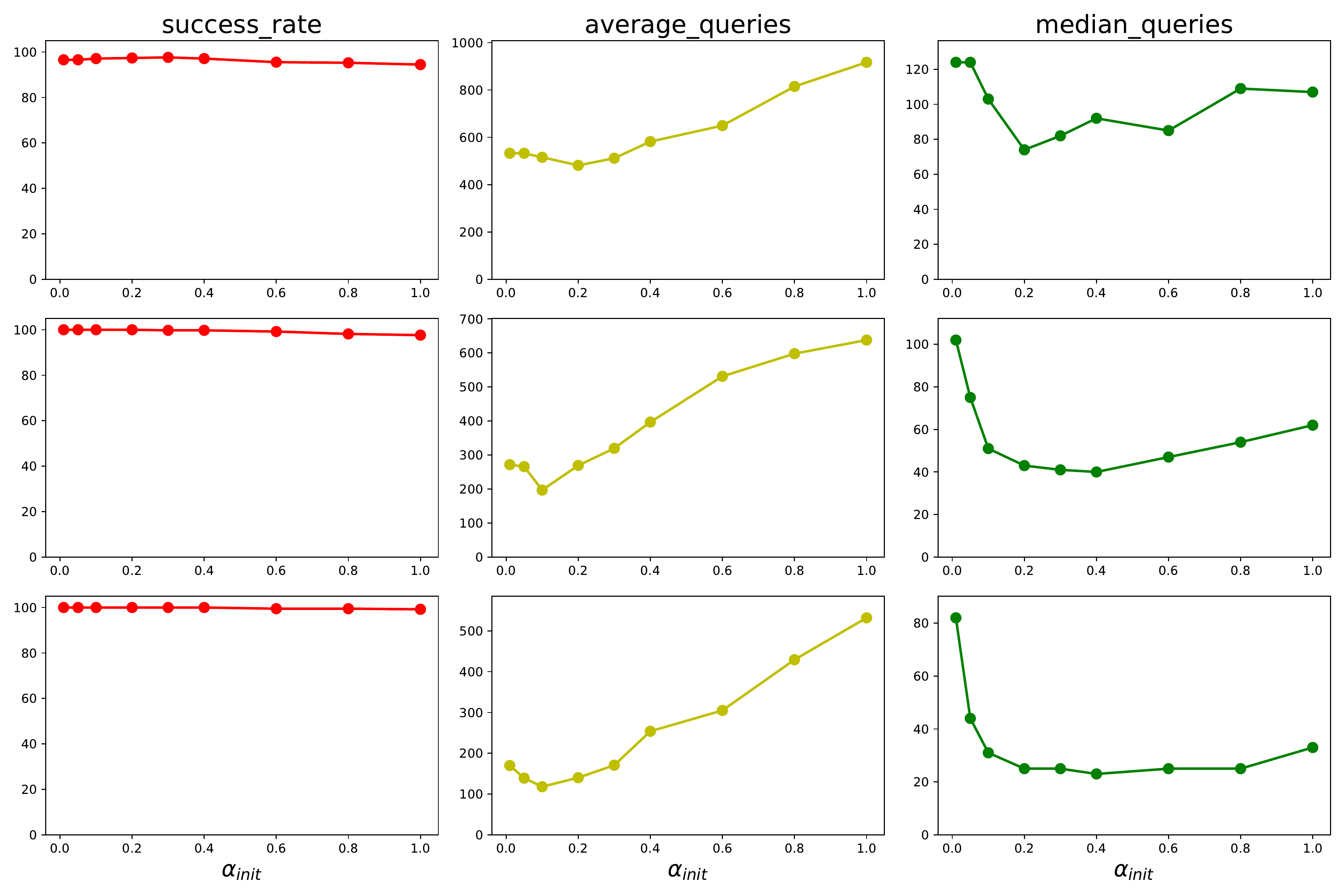}
\end{tabular}
\caption{Ablation study on the influence of $\alpha_\text{init}$, the hyperparameter which regulates the size of the updates at each iteration. We show success rate (first column), average (second) and median queries (third) achieved by \texttt{$l_0$-RS} on VGG at sparsity levels $k=\{50, 100, 150\}$. Considering jointly the three statistics values in $[0.1, 0.4]$ are preferable for this threat model and schedule.} \label{fig:abl_alpha_init}\end{figure*}

\begin{figure*}[t] \centering \small
\setlength{\tabcolsep}{1pt}
\begin{tabular}{c c} \begin{tabular}{c} $k=50$ \\[22mm]
$k=100$\\[22mm] $k=150$
\end{tabular} & \includegraphics[align=c, width=1.8\columnwidth]{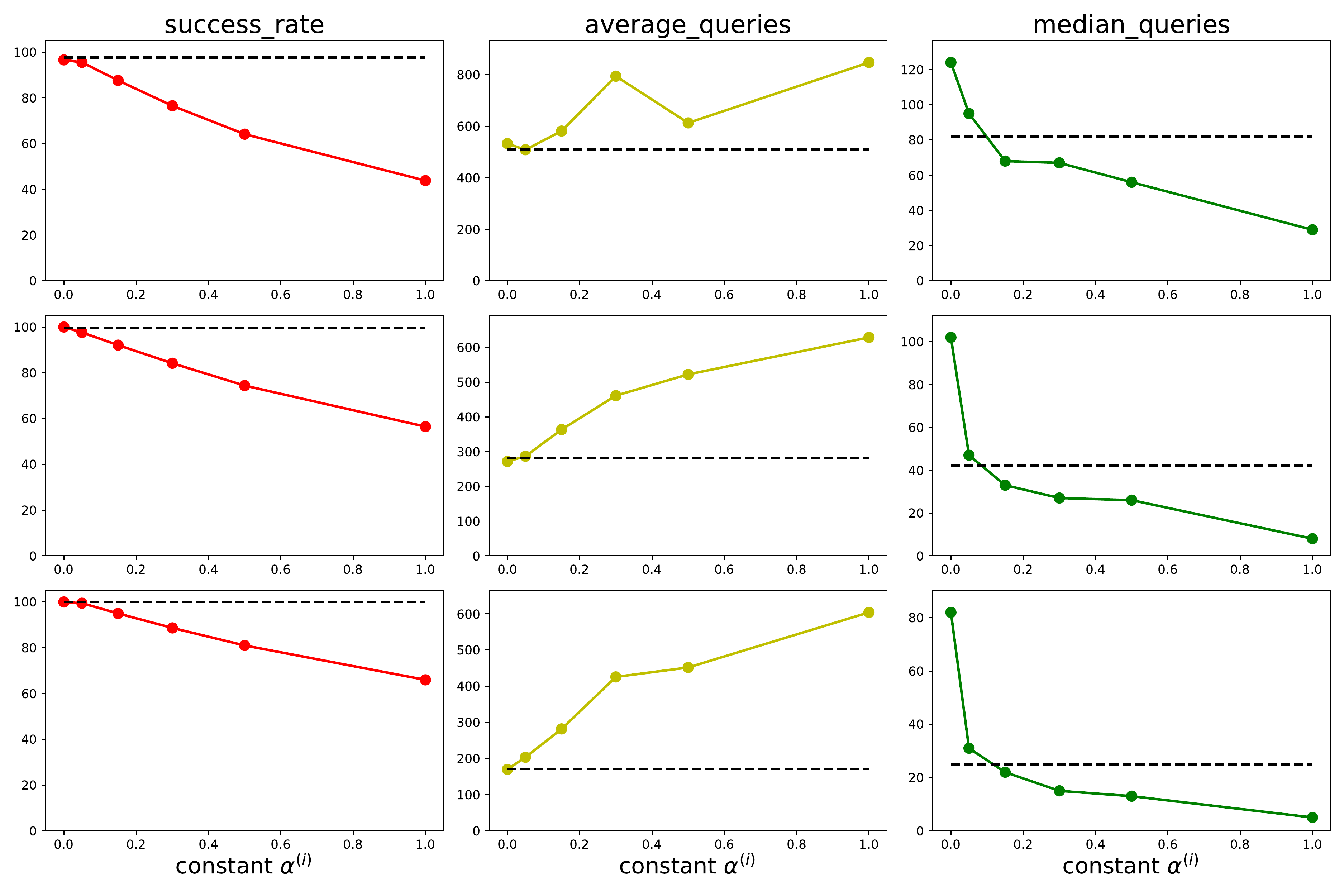}
\end{tabular}
\caption{Performance of untargeted \texttt{$l_0$-RS} on VGG when using a constant schedule for $\iter{\alpha}{i}$, that is the size of $|A|=|B|=c \cdot k$ (see Sec.~\ref{sec:l0_bounded}) at every iteration, or equivalently $\iter{\alpha}{i}= c$ for every $i=1, \ldots, N$. The dashed line is the reference of the results achieved with the piecewise constant schedule to decay $\iter{\alpha}{i}$. While constantly small updates lead to good success rate, the median number of queries used increases notably, especially with larger sparsity levels $k$.} \label{fig:abl_constant_alpha}
\end{figure*}

\subsection{Image-specific adversarial patches}\label{sec:app_abl_patches}
We here show the individual effect of our algorithmic choices to craft image-specific adversarial patches.

\textbf{Initialization and single-channel updates.}
\begin{table*}[t]
    \centering \small
    \setlength{\tabcolsep}{5pt}
    \begin{tabular}{c c c | c c  c |c  c c } %
    \multirow{2}{*}{\textit{attack}} &  %
    \multirow{2}{*}{\textit{init.}} & \textit{single}&
    \multicolumn{3}{c}{VGG} & \multicolumn{3}{c}{ResNet}\\
    & & \textit{channel} &\textit{succ. rate} & \textit{mean queries} & \textit{med. queries} & \textit{succ. rate} & \textit{mean queries} & \textit{med. queries}\\
    \toprule
    \texttt{Sparse-RS} + SH %
    & - & - &82.6\% & 2479 & 514  & 75.3\% & 3290 & 549\\
    \texttt{Sparse-RS} + SA %
    & stripes& no& %
    85.6\% $\pm$ 1.1 & 2367 $\pm$ 83 & 533 $\pm$ 40  & 78.5\% $\pm$ 1.0 & 2877 $\pm$ 64 & 458 $\pm$ 43\\
    \rowcolor{lightgrey}
    & stripes & yes & %
    85.7\% $\pm$ 1.4 & 2359 $\pm$ 89 & 533 $\pm$ 40  & 79.1\% $\pm$ 1.1 & 2861 $\pm$ 58 & 458 $\pm$ 43\\
    \rowcolor{lightgrey}
    & squares & no & %
    87.1\% $\pm$ 0.9 & 2174 $\pm$ 50 & 429 $\pm$ 22  & 78.9\% $\pm$ 1.3 & 2820 $\pm$ 91 & 438 $\pm$ 68\\
    \rowcolor{lightgrey}
    \textbf{\texttt{Patch-RS}} & \textbf{squares} & \textbf{yes} & %
    87.8\% $\pm$ 0.7 & 2160 $\pm$ 44 & 429 $\pm$ 22  & 79.5\% $\pm$ 1.4 & 2808 $\pm$ 89 & 438 $\pm$ 68\\
    \bottomrule \end{tabular}
    \caption{Ablation study on the effect of initialization and single-channel updates for untargeted image-specific adversarial  patches (size $20\times 20$ pixels) within our proposed framework. We highlight in grey our modified schemes that lead to a higher success rate and query efficiency. Note that the single-channel updates come later in the optimization process when 50\% success rate is reached, and thus they do not influence the median. We also report the results of \texttt{Sparse-RS} + SH for reference.} \label{tab:abl_init_cu_patches}
    \end{table*}
The original Square Attack \cite{ACFH2019square} for $l_\infty$- and $l_2$-bounded perturbations applies as initialization vertical stripes on the target image, and it does not consider updates of a single channel (these can be randomly sampled but are not explicitly performed). Table~\ref{tab:abl_init_cu_patches} shows the success rate and query efficiency (computed on all points) for untargeted adversarial attacks with patches of size $20\times 20$ on VGG and ResNet, averaged over 5 random seeds for methods based on random search. Both the initialization with squares and the single-channel updates already individually improve the performance of \texttt{Sparse-RS} + SA and, when combined, allow \texttt{Patch-RS} to outperform the competitors (see also Sec.~\ref{sec:image_specific_patches}).

\textbf{Ratio location to patch updates.}
\begin{table*}[t]
    \centering \small
    \setlength{\tabcolsep}{6pt}
    \begin{tabular}{c c | c c  c |c  c c } %
    \textit{ratio}& \multirow{2}{*}{\textit{attack}} &  %
    \multicolumn{3}{c}{VGG} & \multicolumn{3}{c}{ResNet}\\
    \textit{location : patch} & &\textit{succ. rate} & \textit{mean queries} & \textit{med. queries} & \textit{succ. rate} & \textit{mean queries} & \textit{med. queries}\\
\toprule
\multirow{3}{*}{9:1} & \rsa{} + SH %
& 74.9\% & 3897 & 1331  & 68.0\% & 4038 & 1121\\
&\rsa{} + SA %
& %
71.4\% $\pm$ 0.9 & 3927 $\pm$ 104 & 1361 $\pm$ 224  & 68.7\% $\pm$ 1.0 & 3931 $\pm$ 76 & 789 $\pm$ 65\\
&\texttt{Patch-RS} & %
76.8\% $\pm$ 0.8 & 3414 $\pm$ 48 & 779 $\pm$ 69  & 72.0\% $\pm$ 0.8 & 3655 $\pm$ 57 & 731 $\pm$ 97\\
\midrule
\multirow{3}{*}{4:1} & \rsa{} + SH %
& 77.1\% & 3332 & 781  & 72.9\% & 3639 & 701\\
&\rsa{} + SA %
& %
77.7\% $\pm$ 1.2 & 3236 $\pm$ 106 & 735 $\pm$ 41  & 73.6\% $\pm$ 0.6 & 3408 $\pm$ 35 & 548 $\pm$ 65\\

&\texttt{Patch-RS} & %
82.4\% $\pm$ 0.8 & 2816 $\pm$ 28 & 485 $\pm$ 74  & 75.3\% $\pm$ 0.5 & 3248 $\pm$ 51 & 443 $\pm$ 41\\

\midrule
\multirow{3}{*}{1:1} & \texttt{Sparse-RS} + SH %
& 82.9\% & 2718 & 500  & 75.3\% & 3168 & 451\\
&\texttt{Sparse-RS} + SA %
& %
84.6\% $\pm$ 1.2 & 2485 $\pm$ 92 & 504 $\pm$ 20  & 78.2\% $\pm$ 0.7 & 2943 $\pm$ 142 & 389 $\pm$ 46\\
&\texttt{Patch-RS} & %
86.8\% $\pm$ 0.8 & 2301 $\pm$ 61 & 352 $\pm$ 28  & 79.6\% $\pm$ 1.1 & 2764 $\pm$ 86 & 316 $\pm$ 10\\
\midrule
\multirow{3}{*}{\textbf{1:4}} & \rsa{} + SH %
& 82.6\% & 2479 & 514  & 75.3\% & 3290 & 549\\
&\texttt{Sparse-RS} + SA %
& %
85.6\% $\pm$ 1.1 & 2367 $\pm$ 83 & 533 $\pm$ 40  & 78.5\% $\pm$ 1.0 & 2877 $\pm$ 64 & 458 $\pm$ 43\\
& \texttt{Patch-RS}& %
87.8\% $\pm$ 0.7 & 2160 $\pm$ 44 & 429 $\pm$ 22  & 79.5\% $\pm$ 1.4 & 2808 $\pm$ 89 & 438 $\pm$ 68\\
\midrule
\multirow{3}{*}{1:9} & \rsa{} + SH %
& 80.7\% & 2719 & 564  & 72.2\% & 3451 & 843\\
&\rsa{} + SA %
& %
85.9\% $\pm$ 0.8 & 2419 $\pm$ 60 & 542 $\pm$ 34  & 77.6\% $\pm$ 0.8 & 3000 $\pm$ 92 & 617 $\pm$ 41\\
& \texttt{Patch-RS}& %
87.5\% $\pm$ 1.6 & 2261 $\pm$ 44 & 542 $\pm$ 39  & 78.4\% $\pm$ 0.6 & 2961 $\pm$ 34 & 533 $\pm$ 54\\
\bottomrule
\end{tabular} \vspace{0mm} \caption{Performance of different attacks for untargeted image-specific adversarial patches of size $20\times 20$ varying the ratio between updates of the location and of the patch. We use 1:4 ratio in our framework which is the best choice for \texttt{Sparse-RS} + SA and \texttt{Patch-RS} (for \texttt{Sparse-RS} + SH 1:1 would work equally well). Note that our novel attack \texttt{Patch-RS} outperforms the competitors consistently in terms of the success rate and query efficiency across all choices of the ratio of the location and patch updates. For random search based methods, we repeat the attack with 5 random seeds.} \label{tab:abl_ratio_location_patch}
\end{table*}
We study here the effect of different ratios between the number of location and patch updates in our framework for image-specific untargeted patches (size $20\times 20$). In Table~\ref{tab:abl_ratio_location_patch} we report the performance \texttt{Sparse-RS} + SH, \texttt{Sparse-RS} + SA and our final method \texttt{Patch-RS} with various schemes: alternating an update of the location to one of the patch (1:1), doing 4 or 9 location updates for one patch update (4:1, 9:1) and vice versa (1:4, 1:9). Considering the results on both networks, the scheme with 1:4 ratio yields for every attack the highest success rate or very close. In query consumption, the schemes 1:1 and 1:4 are similar, with some trade-off between average and median. Moreover, we observe that the ranking of the three methods is preserved for all ratios of location and patch updates we tested, with \texttt{Patch-RS} outperforming the others. Also, its effectiveness is less affected by suboptimal ratios, showing the stability of our algorithm.

\end{document}